\documentclass[10pt,journal,compsoc]{IEEEtran}

\ifCLASSOPTIONcompsoc
  \usepackage[nocompress]{cite}
\else
  \usepackage{cite}
\fi

\ifCLASSOPTIONcompsoc
    \usepackage[caption=false, font=normalsize, labelfont=sf, textfont=sf]{subfig}
\else
\usepackage[caption=false, font=footnotesize]{subfig}
\fi

\usepackage{booktabs} 
\usepackage{algorithm}
\usepackage{makecell}
\usepackage{algorithmic}
\usepackage{multirow}
\usepackage{amsfonts}
\usepackage{amsmath}
\usepackage{float}
\usepackage{bbm}
\usepackage{verbatim}
\usepackage{amsthm}
\usepackage[super]{nth}
\usepackage{graphicx}
\usepackage{caption}
\usepackage{color, colortbl}
\ifCLASSOPTIONcompsoc
  \usepackage[caption=false,font=normalsize,labelfont=sf,textfont=sf]{subfig}
\else
  \usepackage[caption=false,font=footnotesize]{subfig}
\fi

\theoremstyle{definition}
\newtheorem{definition}{Definition}

\definecolor{Gray}{gray}{0.9}

\begin{document}
\title{Disentangling Structured Components: Towards Adaptive, Interpretable and Scalable Time Series Forecasting}

\author{Jinliang~Deng,
        Xiusi~Chen,
        Renhe~Jiang,
        Du~Yin,
        Yi~Yang,
        Xuan~Song,~and
        Ivor~W.~Tsang,~\IEEEmembership{Fellow,~IEEE}\
\IEEEcompsocitemizethanks{\IEEEcompsocthanksitem J. Deng  is with Australian Artificial Intelligence Institute, University of Technology Sydney, Sydney, Australia. J. Deng is also affiliated with Department of Computer Science and Engineering, Southern University of Science and Technology, Shenzhen, China.
E-mail: jinliang.deng@student.uts.edu.au.
\IEEEcompsocthanksitem X. Chen is with University of California, Los Angeles, USA. Email: xchen@cs.ucla.edu.
\IEEEcompsocthanksitem R. Jiang is with Center for Spatial Information Science, University of Tokyo, Tokyo, Japan. Email: jiangrh@csis.u-tokyo.ac.jp.
\IEEEcompsocthanksitem D. Yin is with the Department of Computer Science and Engineering, University of New South Wales, Sydney, Australia. Email: yind7@outlook.com.
\IEEEcompsocthanksitem Y. Yang is with Tencent WXG, Guangzhou, China. Email: paulyyyang@tencent.com.
\IEEEcompsocthanksitem X. Song is with SUSTech-UTokyo Joint Research Center on Super Smart City, Department of Computer Science and Engineering, Southern University of Science and Technology (SUSTech), Shenzhen, China. Email: songx@sustech.edu.cn.
\IEEEcompsocthanksitem Ivor W. Tsang is with the Center for Frontier AI Research, Agency for Science, Technology and Research (A*STAR), Singapore. Email: ivor.tsang@gmail.com.
}
\thanks{Xuan~Song and Ivor~W.~Tsang are corresponding authors.}}

\IEEEtitleabstractindextext{%
\begin{abstract}
Multivariate time-series (MTS) forecasting is a paramount and fundamental problem in many real-world applications. The core issue in MTS forecasting is how to effectively model complex spatial-temporal patterns. In this paper, we develop a adaptive, interpretable and scalable forecasting framework, which seeks to individually model each component of the spatial-temporal patterns. We name this framework SCNN, as an acronym of \underline{S}tructured \underline{C}omponent-based \underline{N}eural \underline{N}etwork. SCNN works with a pre-defined generative process of MTS, which arithmetically characterizes the latent structure of the spatial-temporal patterns. In line with its reverse process, SCNN decouples MTS data into structured and heterogeneous components and then respectively extrapolates the evolution of these components, the dynamics of which are more traceable and predictable than the original MTS. Extensive experiments are conducted to demonstrate that SCNN can achieve superior performance over state-of-the-art models on three real-world datasets\footnote{Code available at: https://github.com/KDDtest/SCNN.git}. Additionally, we examine SCNN with different configurations and perform in-depth analyses of the properties of SCNN.
\end{abstract}

}

\maketitle

%

%

\maketitle

\section{Introduction}

Multivariate time series (MTS) forecasting is a fundamental problem in the machine learning field \cite{oreshkin2019n, zhang2017deep} since a wide array of promising applications can be conceptualized as MTS forecasting problems. Examples include predicting activities and events~\cite{jiang2019deepurbanevent}, nowcasting precipitation~\cite{lam2022graphcast}, forecasting traffic~\cite{zhang2017deep}, and estimating pedestrian and vehicle trajectories~\cite{li2022graph}. The primary challenge in MTS forecasting is to effectively capture spatial-temporal patterns from MTS data. Spatial characteristics arise from external factors such as regional population, functionality, and geographical location. Temporal characteristics are influenced by factors like the time of day, day of the week, and weather conditions.

\begin{figure}[tb]
\centering
\subfloat[]{
\includegraphics[width=0.49\linewidth]{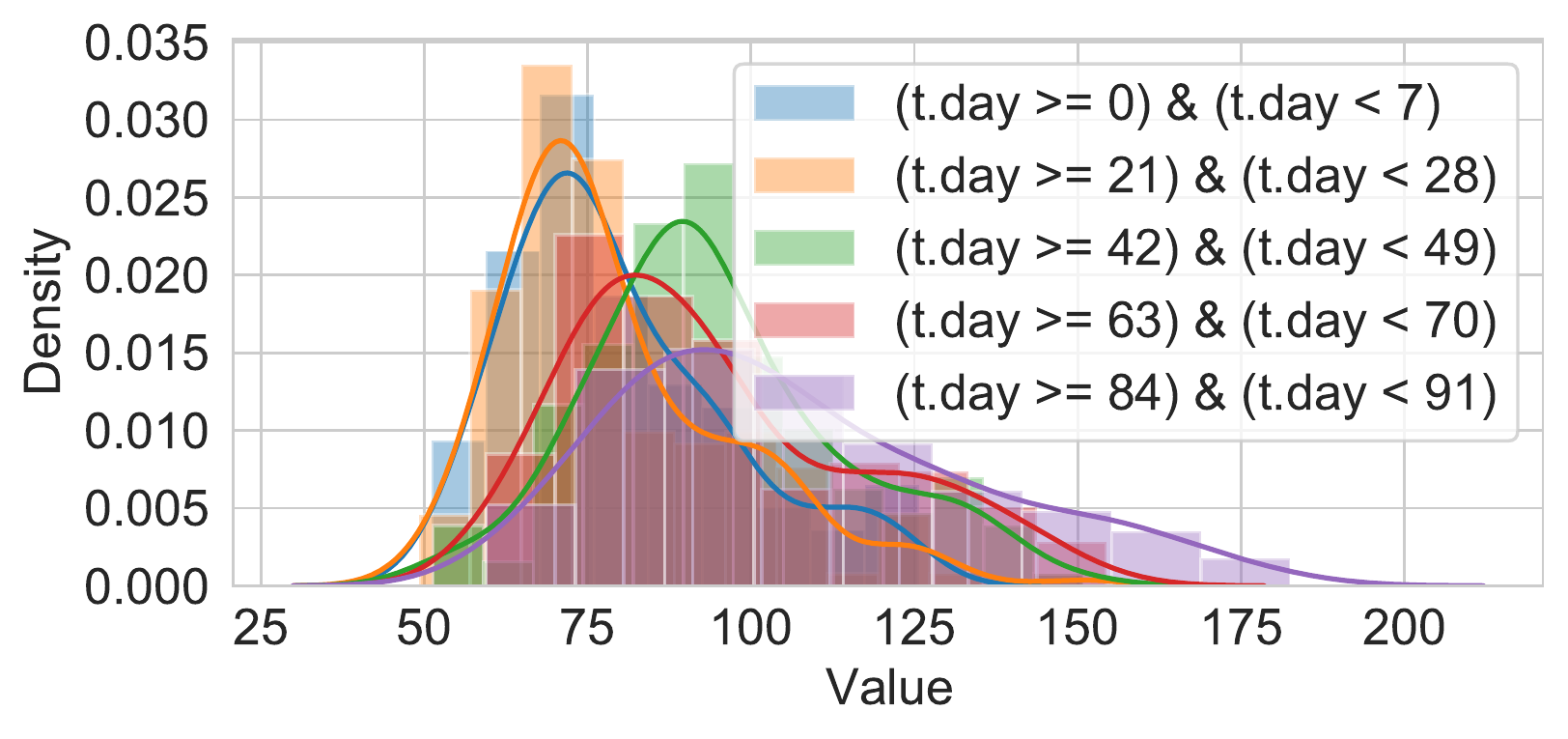}
\label{fig:dist_shift}
}
\subfloat[]{
\includegraphics[width=0.49\linewidth]{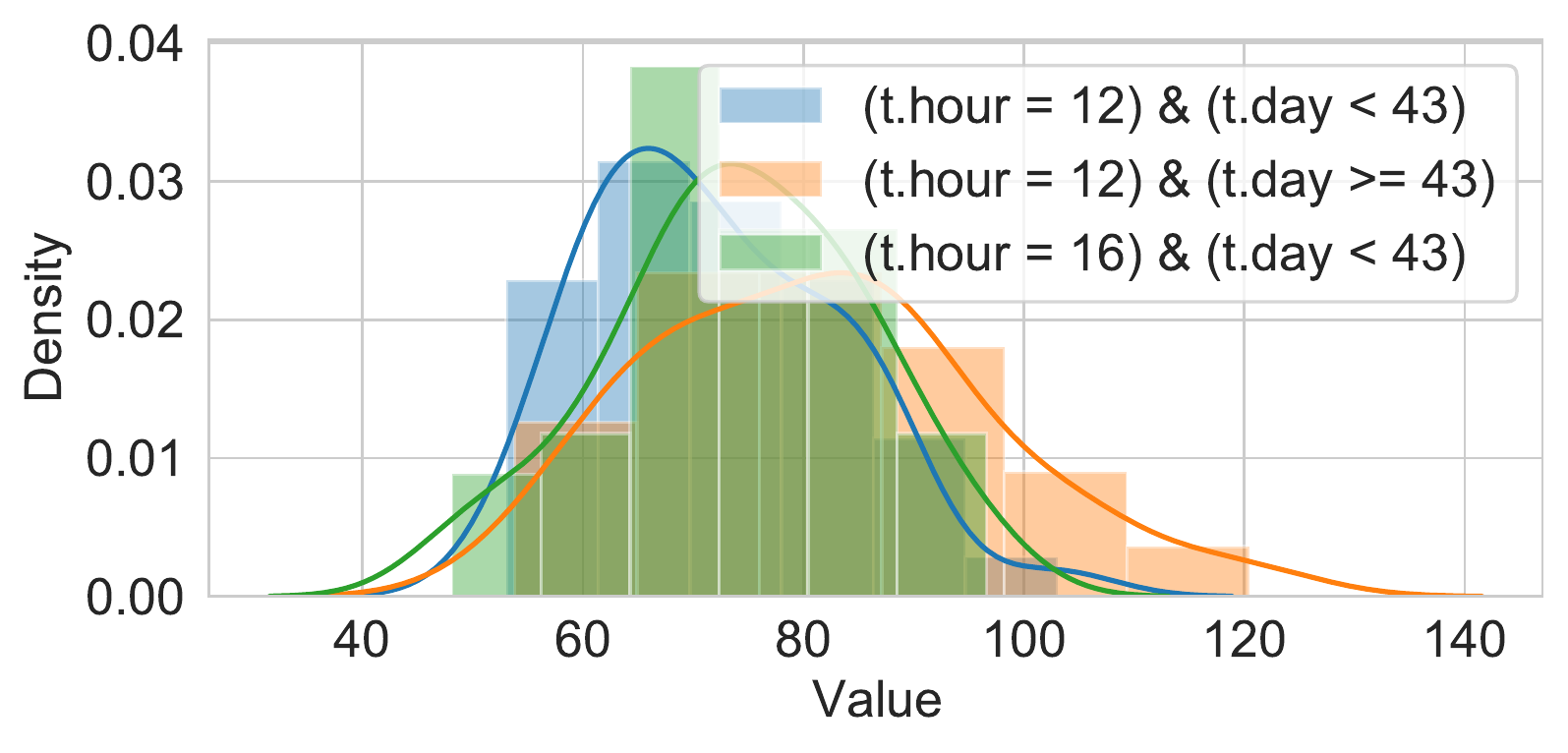}
\label{fig:con_dist_shift}
}

\subfloat[]{
\includegraphics[width=0.49\linewidth]{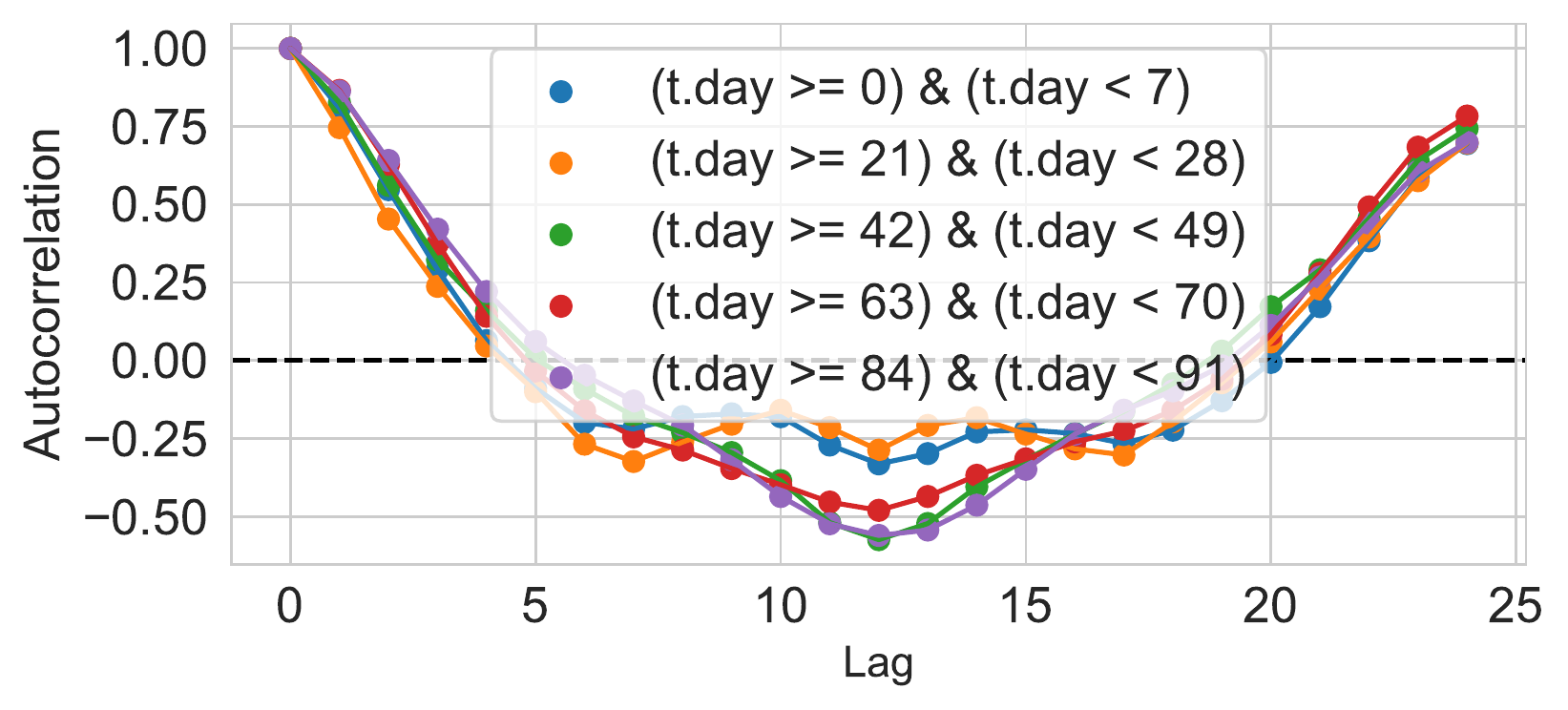}
\label{fig:acf_shift}
}
\subfloat[]{
\includegraphics[width=0.49\linewidth]{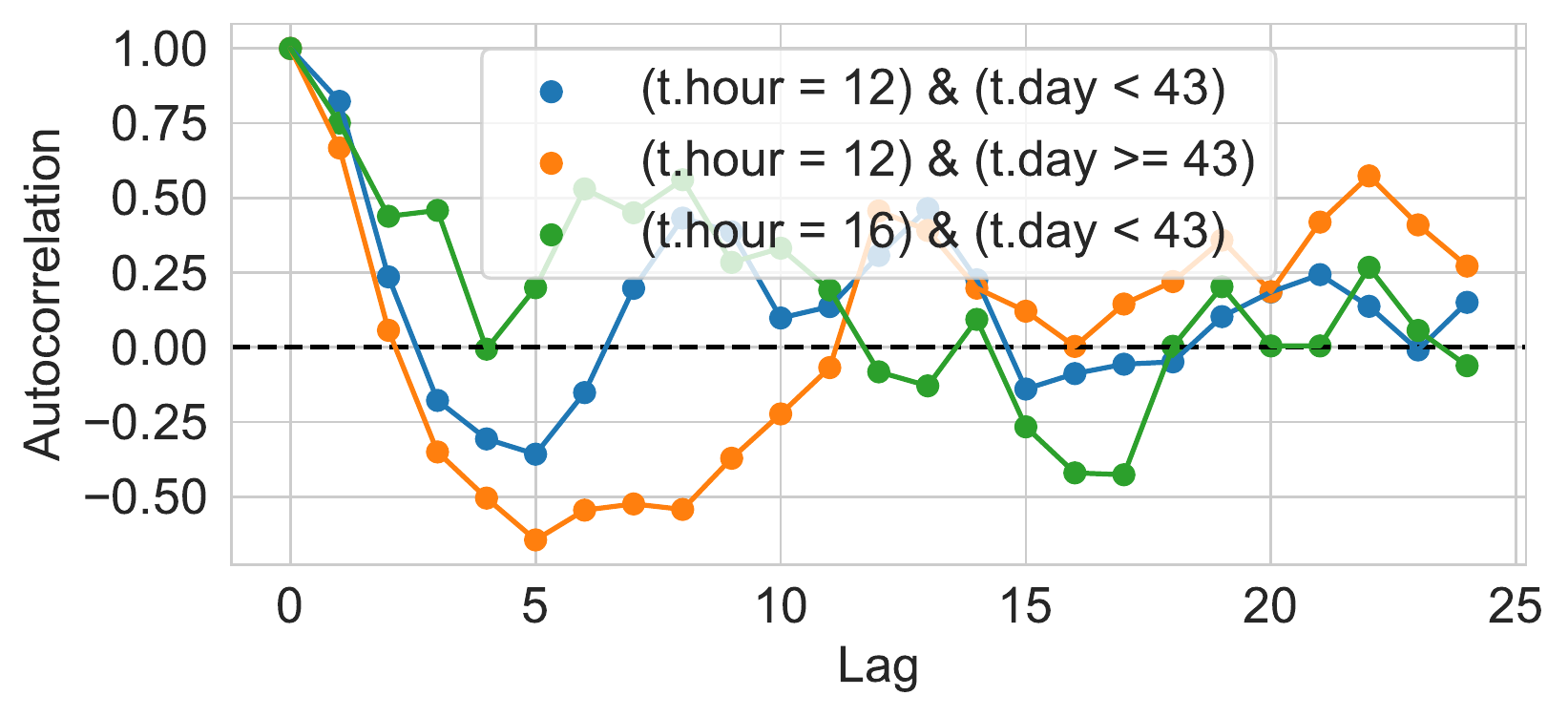}
\label{fig:con_acf_shift}
}
\caption{(a) $P(Y_t|t.\text{day})$; (b) $P(Y_t|t.\text{day}, t.\text{hour})$; (c) $Corr(Y_t, Y_{t-i}|t.\text{day})$; (d) $Corr(Y_t, Y_{t-i}|t.\text{day}, t.\text{hour})$. These visualizations emphasize that both data distribution and auto-correlation exhibit complex, heterogeneous shifts correlated with factors like time span and hour of the day.}
\label{fig:shifts}
\end{figure}

Traditional methods assume that the time series to be modeled is stationary \cite{shumway2000time}. However, real-world multivariate time series are often non-stationary, containing diverse and heterogeneous structured patterns such as multiple-resolution continuity and seasonality. These patterns significantly complicate the dynamics of time series, leading to various forms of distribution shifts, as illustrated in Fig. \ref{fig:dist_shift} and Fig. \ref{fig:con_dist_shift}. These shifts occur constantly and irregularly across hours and days, influenced by long-term continuity and seasonality. Additionally, as shown in Fig. \ref{fig:acf_shift} and Fig. \ref{fig:con_acf_shift}, not only does the data distribution change over time, but the auto-correlation also varies. This variation in auto-correlation, which has received little attention in literature, suggests that the relationships between historical observations and future targets are also unstable, making prediction more challenging.

To address non-stationary time series, modern methods employ deep neural networks like Transformers, temporal convolution networks (TCNs), and recurrent neural networks (RNNs), which do not rely on the assumption of stationarity. However, their effectiveness is limited to handling in-distribution (ID) non-stationary patterns. For example, with sine and cosine functions, their non-stationary patterns recur over time, allowing their dynamics to be captured accurately by deep learning models. However, for out-of-distribution (OOD) non-stationary patterns, the performance of these models often degrades significantly. Thus, adaptability and generalization under complex distribution shifts remain underexplored in current deep spatial-temporal models \cite{lai2018modeling, bai2020adaptive, wu2019graph, wu2020connecting, zhang2021traffic}. Additionally, these methods render the prediction process a black box, lacking interpretability. They also require extensive parameters and operations, leading to prohibitively expensive computations.

Time series decomposition \cite{shumway2000time}, which separates time series into trend, seasonal, and residual components, has recently emerged as a promising approach to enhance adaptability to OOD non-stationary patterns and improve interpretability of deep learning models \cite{wu2021autoformer, wang2022learning, deng2021st, liu2022non, woo2021cost}. Even simple linear models \cite{Zeng2022AreTE} have shown the ability to outperform various deep learning models \cite{wu2021autoformer, zhou2021informer, zhou2022fedformer} when using this approach.

Despite these advancements, current studies still have limitations. First, they focus mainly on long-term and seasonal components, capturing only coarse-grained trends while neglecting short-term or volatile components crucial for detailed deviations. Second, the segregated processing of different components without information exchange inhibits the extraction of high-order and non-linear interactions among them. Third, employing static model parameters is sub-optimal for OOD patterns behaving dynamic auto-correlation, given that the optimal parameter solution should correlate with the real-time evaluation of auto-correlation. Due to these shortcomings, previous decomposition-driven methods still rely on large-scale MLPs or Transformers to enhance model expressiveness, resulting in reduction in scalability and interpretability \cite{nie2023a, liu2022non, zhang2023crossformer}.

\begin{figure}
    \centering
    \includegraphics[width=0.9\linewidth]{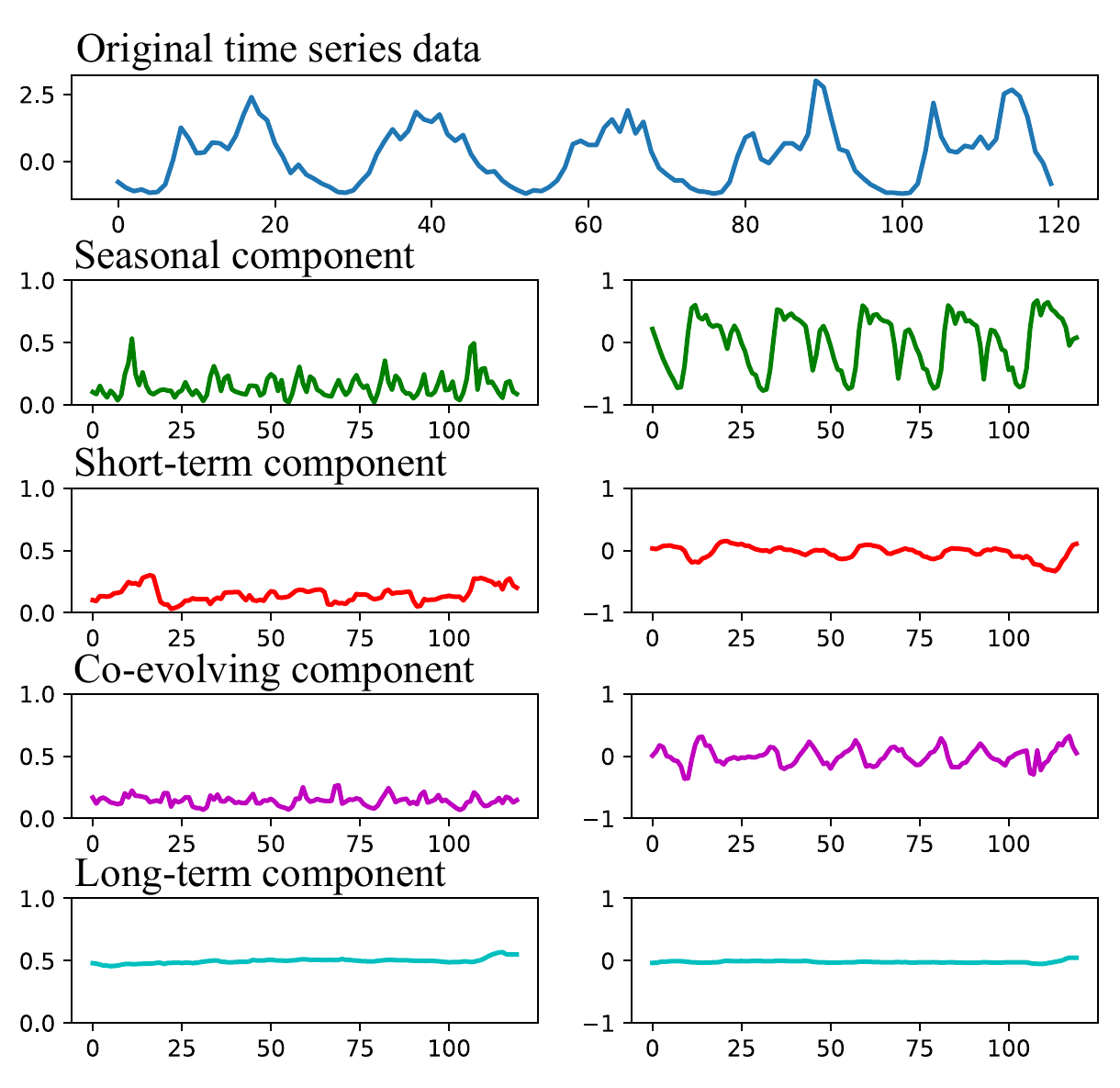}
    \caption{Structured components extracted by SCNN from BikeNYC time series data. The underlying structure of TS might be far more complicated than just trend (long-term) and seasonal components.}
    \label{fig:example_data}
\end{figure}

In response to the limitations identified above, our study introduces a structured component-based neural network (SCNN) for MTS forecasting. First, SCNN employs a divide-and-conquer strategy, strategically disentangling time series data into multiple structured components, as shown in Fig. \ref{fig:example_data}, extending beyond long-term and seasonal components. These components exhibit heterogeneous dynamics, suitable for simulation with simple, specially-designed models. This approach significantly enhances the model's ability to handle heterogeneous distribution shifts while improving the transparency of its internal mechanisms. Second, unlike previous methods, where decomposition and recomposition are applied only at the input and output stages, respectively, we integrate these operations into the design of the neural modules comprising SCNN. Deep and iterative decoupling of components allows for incorporating a wide range of high-order interactions among them, thereby enhancing the model's expressiveness. Third, to address auto-correlation shifts, each neural module features a bifurcated structure, enabling dynamic and adaptive model parameter updates: one branch adjusts model parameters based on real-time data, akin to a small hyper-network \cite{ha2017hypernetworks}, while the other processes hidden features with the adjusted parameters. Finally, to improve SCNN's generalization ability, we introduce auxiliary structural regularization alongside the standard regression loss. This encourages the model to focus more on structured components less prone to corruption. The components utilized in SCNN enable an adaptive, interpretable, scalable, yet powerful neural architecture for time series forecasting.

We summarize our contributions as follows:
\begin{itemize}
\item We introduce the Structured Component Neural Network (SCNN) for multivariate time series forecasting, marking the first completely decomposition-based neural architecture.

\item We propose a novel structural regularization method to explicitly shape the structure of the representation space learned from SCNN.

\item We conduct extensive experiments on three public datasets to validate the effectiveness of SCNN, and observe general improvement over competing methods.

\item Empirical and analytical evidence demonstrates the SCNN's superior performance in handling distribution shifts and anomalies, while maintaining computational efficiency.
\end{itemize}

\section{Related Work}
The time series forecasting community has undergone rapid development since the flourishing of deep learning models \cite{jiang2021dl}. The vast majority of works inherit from a small group of canonical operations, consisting of the attention operator, the convolution operator and the recurrent operator. In particular, the derivatives of the attention operator include spatial attention \cite{fang2019gstnet, zheng2020gman, liang2018geoman}, temporal attention \cite{zhou2018predicting, zheng2020gman, liu2023spatio} and sparse attention (to improve computational efficiency) \cite{wu2021autoformer, zhou2021informer, li2019enhancing}; the convolution operator is developed to spatial convolution \cite{yu2018spatio, li2018diffusion, guo2021hierarchical}, temporal convolution \cite{wu2019graph, wu2020connecting}, spatial-temporal convolution \cite{8684259, yang2021space} and adaptive convolution (where the parameters of the convolution operator can adapt to external conditions) \cite{pan2019urban}; the recurrent operator stimulates the development of gated recurrent units (GRU) \cite{zhao2019t}, long short-term memory (LSTM) \cite{yao2019revisiting, yao2018deep} and adaptive RNN \cite{wang2019deep, salinas2020deepar, rangapuram2018deep, pan2019urban}. 

To further supplement the operations above, various tricks are created. For example, to handle cases where spatial or temporal relationships are incomplete, several studies \cite{li2021spatial, wu2019graph, wu2020connecting, zhang2021traffic, han2021dynamic, liu2022multivariate, pmlr-v162-lan22a, 10.1145/3534678.3539274, 10.1145/3534678.3539396, jiang2022spatio} make use of an adaptive graph learning module to recover the relationships from data adaptively. To incorporate domain knowledge, such as periodicity, into modeling, several studies \cite{geng2019spatiotemporal, chai2018bike, zonoozi2018periodic, chen2019gated} have devised ad-hoc network architecture with handcrafted connections between neural units; another line of research \cite{zheng2020gman, deng2021pulse} represents knowledge with a group of learnable vectors, and feeds them into the model accompanied by MTS data. Furthermore, \cite{cao2020spectral, pmlr-v162-zhou22g} used Fourier transform to decompose original MTS data into a group of orthogonal signals; \cite{yao2019learning} resorted to memory networks to enable the long-term memory of historical observations; \cite{fang2021spatial} exploited a graph ordinary differential equation (ODE) to address the over-smoothing problem of graph convolution networks; \cite{pan2021autostg, li2020autost} took advantage of neural architecture search algorithms to search for the optimal connections between different kinds of neural blocks; and \cite{lin2021ssdnet} integrated a transformer with a state space model to provide probabilistic and interpretable forecasts. The study by \cite{pan2023magicscaler} innovatively integrates multi-scale attention mechanisms, renowned for their efficacy in identifying complex, multi-scale features, with stochastic process regression, known for its ability to quantify prediction uncertainty. This synergistic combination facilitates highly accurate demand forecasting while providing quantified uncertainty levels, marking a significant advancement in the field.

Recently, an emerging line of approaches capitalize on the decomposition techniques to enhance the effectiveness and interpretability of time series forecasting models. \cite{wang2022learning, woo2021cost} disentangled trend and seasonal components from TS data in latent space via a series of auxiliary objectives; \cite{liu2022non} integrated a decomposition module into the transformer framework to approach the non-stationary issue; \cite{deng2021st, deng2021multiview} proposed spatial and temporal normalization to decompose MTS data from the spatial and temporal view, respectively. The novelty of our work is that we are the first to devise a completely decomposition-based neural architecture where the components are estimated in an attentive way to allow for data-driven adaptation. Our model achieves remarkable results compared to the state-of-the-arts based on TCNs, Transformer or RNNs.

\section{Preliminaries}
\newtheorem{assumption}{Premise}

In this section, we introduce the definitions and the assumption. All frequently used notations are reported in Table \ref{tab:notation}.

\begin{table}[thb]
\small
\caption{Notations} \label{tab:notation}
\begin{tabular}{l|l}
\hline
Notation    & Description   \\
\hline
$N, L$ & Number of variables / network layers. \\
$T_{\text{in}}, T_{\text{out}}$ & Number of input steps / output steps. \\
$Y \in \mathbb{R}^{N \times T}$ & Multivariate time series. \\
$Y^{\text{in}}_{n, t} \in \mathbb{R}$ & Observation of $n^\text{th}$ variable at time $t$. \\
$\hat{Y}^{\text{out}}_{n, t+i} \in \mathbb{R}$ & \begin{tabular}{@{}l@{}}Mean prediction of the $n^\text{th}$ variable for\\ the $i^\text{th}$ forecast horizon at time $t$\end{tabular}. \\
$\hat{\sigma}^{\text{out}}_{n, t+i} \in \mathbb{R}$ & \begin{tabular}{@{}l@{}}Standard deviation prediction of the\\ $n^\text{th}$ variable for the $i^\text{th}$ forecast horizon\\ at time $t$.\end{tabular}. \\
lt, se, st, ce & 
\begin{tabular}{@{}l@{}}Abbreviations for 4 types of structured\\ components: long-term, seasonal,\\ short-term,  co-evolving.\end{tabular}\\
$\mu_{n, t}^{*}, \sigma_{n, t}^{*}  \in \mathbb{R}^{d_z}$ & Historical structured component. \\
$\hat{\mu}_{n, t+i}^{*}, \hat{\sigma}_{n, t+i}^{*}  \in \mathbb{R}^{d_z}$ & Extrapolation of the structured component. \\
$H_{n, t} \in \mathbb{R}^{8d_z}$ & \begin{tabular}{@{}l@{}}Concatenation of historical structured\\ components of 4 types.\end{tabular} \\
$\hat{H}_{n, t+i} \in \mathbb{R}^{8d_z}$ & \begin{tabular}{@{}l@{}}Concatenation of extrapolated \\structured components of 4 types. \end{tabular}\\
$Z_{n, t}^{(l)} \in \mathbb{R}^{d_z}$ & \begin{tabular}{@{}l@{}}Historical residual representation at\\ the $l^\text{th}$ layer in the decoupling block.\end{tabular} \\
$\hat{Z}_{n, t+i}^{(l)} \in \mathbb{R}^{d_z}$ & \begin{tabular}{@{}l@{}}Extrapolation of the residual \\representation at the $l^\text{th}$ layer.\end{tabular} \\
$Z_{n, t} \in \mathbb{R}^{4d_z}$ & \begin{tabular}{@{}l@{}}Concatenation of historical residual \\representations at 4 layers.\end{tabular} \\
$\hat{Z}_{n, t+i} \in \mathbb{R}^{4d_z}$ & \begin{tabular}{@{}l@{}}Concatenation of extrapolated \\ residual representations at 4 layers.\end{tabular} \\ 
$S_{n, t} \in \mathbb{R}^{d_z}$ & Historical state. \\
$\hat{S}_{n, t+i} \in \mathbb{R}^{d_z}$ & Extrapolation of the state. \\ 
\hline
\end{tabular}
\end{table}

\begin{definition}[Multivariate time series forecasting] Multivariate time series is formally defined as a collection of random variables $\{Y_{n, t}\}_{n \in N, t \in T}$, where $n$ denotes the index on the spatial domain and $t$ denotes the index on the temporal domain. Time series forecasting is formulated as the following conditional distribution:
\[
P(Y_{:, t+1:t+T_{\text{out}}} | Y_{:, t-T_{\text{in}}+1:t}) = \prod_{i=1}^{T_{\text{out}}} P(Y_{:, t+i} | Y_{:, t-T_{\text{in}}+1:t}).
\]
\end{definition}

Our study delves into a specific category of time series that can be represented as a superposition of various elementary signals. These include the \underline{l}ong-\underline{t}erm (lt) component, the \underline{se}asonal (se) component, the \underline{s}hort-\underline{t}erm (st) component, the \underline{c}o-\underline{e}volving (ce) component, and the residual component. Each component offers a distinct perspective on the underlying dynamic system of the time series, enriching the information content of the series.

\begin{definition}[Generative Process for Multivariate Time Series]
We postulate that the time series is generated through the following process:
\begin{align}
    Z^{(3)}_{n, t} &= \sigma^\text{ce}_{n, t}  R_{n, t} + \mu^\text{ce}_{n, t}, \\
    Z^{(2)}_{n, t} &= \sigma^\text{st}_{n, t}  Z^{(3)}_{n, t} + \mu^\text{st}_{n, t}, \\
    Z^{(1)}_{n, t} &=  \sigma^\text{se}_{n, t}  Z^{(2)}_{n, t} + \mu^\text{se}_{n, t}, \\
    Z^{(0)}_{n, t} &= \sigma^\text{lt}_{n, t}  Z^{(1)}_{n, t}+ \mu^\text{lt}_{n, t},
\end{align}
where $R_{n, t}$ denotes the residual component; $Z^{(0)}_{n, t}$ represents the original data, and $Z^{(i)}_{n, t}$ ($i \in \{1, 2, 3\}$) signifies the intermediate representation at the $i^\text{th}$ level. Each structured component is defined by a multiplicative (scaling) factor $\sigma^*_t$ and an additive factor $\mu^*_t$, with $* \in \{\text{ce}, \text{st}, \text{se}, \text{lt}\}$.
\end{definition}

To illustrate this generative process intuitively, we consider the analysis of traffic density data. In this scenario, different components capture distinct aspects of traffic dynamics. The long-term component reflects overarching trends in traffic patterns, such as increases due to urban development or population growth. The seasonal component represents cyclical changes, like the rush hour peaks or reduced flow during off-peak times. The short-term component captures immediate, transient effects caused by events like road work or weather changes. The co-evolving component quantifies the simultaneous impact of sudden events on multiple traffic series, such as a traffic accident affecting adjacent roads. Finally, the residual component accounts for random effects, including unpredictable elements like sensor errors.

It is crucial to understand that these classifications in traffic data analysis are dynamic. For example, a sudden traffic increase at a junction might initially be considered an anomaly (residual component) but could evolve into a short-term pattern if it persists due to a temporary detour. If this change becomes permanent, it would then shift to the long-term component. This fluidity highlights the need for adaptable and dynamic analytical methods in traffic data analysis.

Each component in this framework exhibits both multiplicative and additive effects, reflecting the intricate nature of traffic dynamics. The multiplicative effect is vital for understanding proportional changes in traffic volume, such as varying impacts of percentage increases during peak or off-peak hours. The additive effect, on the other hand, represents uniform changes, such as the consistent impact of road constructions or new traffic signals, irrespective of current traffic levels. Incorporating both effects into each component ensures a thorough understanding of traffic dynamics, as different scenarios may necessitate focusing on either proportional (multiplicative) or absolute (additive) changes.

\section{Structured Component-based Neural Network}

\begin{figure*}[t]
    \centering
    \includegraphics[width=\linewidth]{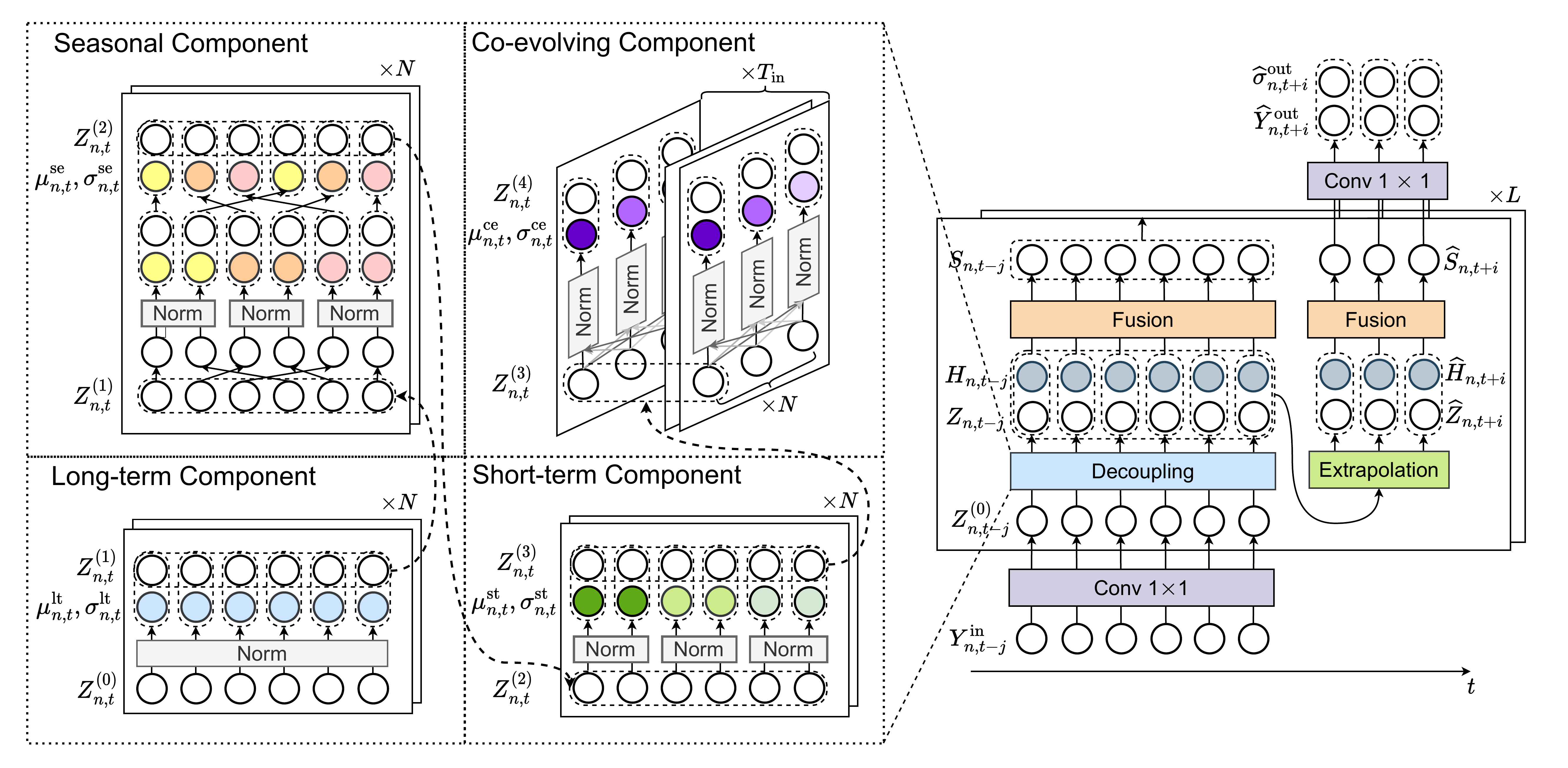}
    \caption{A schematic diagram of SCNN.}
    \label{fig:architecture}
\end{figure*}

Figure \ref{fig:architecture} illustrates an overview of our model architecture. SCNN is composed of four major parts, namely \textit{component decoupling}, \textit{component extrapolation}, \textit{component fusion} and \textit{structural regularization}. We will introduce each part in the following sections.

\subsection{Component Decoupling}

This section introduces how to estimate a specific structured component, and decouple this component from the residuals by applying a normalization operator. This process is presented in the left part of Fig. \ref{fig:architecture}.

\subsubsection{Long-Term Component}
\label{sec:lt}
The long-term component aims to be the characterization of the long-term patterns of the time series data, such as increases due to urban development or population growth, as mentioned in the previous section. To avoid ambiguity, we refer to the pattern as the distribution of the aggregated samples without considering the chronological order among them; the long-term pattern refers to the data distribution over an extended period that should cover multiple seasons. By aggregating the samples collected from multiple seasons, we can eliminate the short-term impact that will affect only a handful of time steps, and acquire the estimation of the long-term component with less bias. 

We create a sliding window of size $\Delta$ to dynamically select the set of samples over time. Then, the location (mean) and scale (standard deviation) of the samples are computed and jointly taken as the measurement of the long-term component. Finally, we transform the representation by subtracting the location from it and dividing the difference by the scale, in order to unify the long-term components for different samples. The formula takes the following form:
\begin{align}
    \mu^{\text{lt}}_{n, t} &= \frac{1}{\Delta}\sum_{i=0}^{\Delta-1} Z^{\text{(0)}}_{n, t-i}\;, \\
    (\sigma^\text{lt}_{n, t})^2 &= \frac{1}{\Delta}\sum_{i=0}^{\Delta-1} (Z^\text{(0)}_{n, t-i})^2 - (\mu^\text{lt}_{n, t})^2 + \epsilon, \\
    Z_{n, t}^{\text{(1)}} &= \frac{Z^\text{(0)}_{n, t} - \mu^\text{lt}_{n, t}}{\sigma^\text{lt}_{n, t}},
\end{align}
where $\mu^\text{lt}_{n, t}$ and $\sigma^\text{lt}_{n, t}$ are the location and the scale respectively; $Z_{n, t}^{\text{(1)}}$ notates the intermediate representation derived by the 1$^\text{st}$ normalization layer, which will be passed to the following normalization layers.


\subsubsection{Seasonal Component}
The seasonal component aims to characterize the seasonal patterns of the time series data, such as the peak flow during rush hours. Our study makes a mild assumption that the cycle length is invariant over time. For those applications with time-varying cycle lengths, we can resort to the Fast Fourier Transform (FFT) to automate the identification of cycle length, which is compatible with our framework and is applied in a bunch of methods like Autoformer \cite{wu2021autoformer}.

Disentanglement of the seasonal component resembles the long-term component, except that we apply a dilated window whose dilation factor is set to the cycle length. Let $\tau$ denote the window size, and $m$ denote the dilation factor. The normalization then proceeds as follows:
\begin{align}
    \mu^\text{se}_{n, t} &= \frac{1}{\tau}\sum_{i=0}^{\tau-1} Z^{\text{(1)}}_{n, t-i * m}\;, \label{eq:se}\\
    (\sigma^\text{se}_{n, t})^2 &= \frac{1}{\tau}\sum_{i=0}^{\tau-1} (Z^{\text{(1)}}_{n, t-i * m})^2 - (\mu^\text{se}_{n, t})^2 + \epsilon, \label{eq:epsilon} \\
    Z_{n, t}^{\text{(2)}} &= \frac{Z^{\text{(1)}}_{n, t} - \mu^\text{se}_{n, t}}{\sigma^\text{se}_{n, t}},
\end{align}
where $Z_{n, t}^{\text{(2)}}$ represents the intermediate representation derived by the 2$^\text{nd}$ normalization layer, which will be passed to the following normalization layers. In this way, the resulting $\mu^\text{se}_{n, t}$ and $\sigma^\text{se}_{n, t}$ will exhibit only seasonal patterns without interference by any temporary or short-term impacts. 

\subsubsection{Short-Term Component}
The short-term component captures the irregular and short-term effects, which cannot be explained by either the long-term component or the seasonal component, such as the influence of weather change or road work. In contrast to the long-term normalization, the window size here needs to be set to a small number, notated by $\delta$, such that the short-term effect will not be smoothed out. Likewise, the formula takes the following form:
\begin{align}
    \mu^\text{st}_{n, t} &= \frac{1}{\delta}\sum_{i=0}^{\delta-1} Z^{\text{(2)}}_{n, t-i}\;, \label{eq:lt}\\
    (\sigma^\text{st}_{n, t})^2 &= \frac{1}{\delta}\sum_{i=0}^{\delta-1} (Z^{\text{(2)}}_{n, t-i})^2 - (\mu^\text{st}_{n, t})^2 + \epsilon, \\
    Z_{n, t}^{\text{(3)}} &= \frac{Z^{\text{(2)}}_{n, t} - \mu^\text{st}_{n, t}}{\sigma^\text{st}_{n, t}},
\end{align}
where $Z_{n, t}^{\text{(3)}}$ stands for the intermediate representation derived by the 3$^\text{rd}$ normalization layer, which will be passed to the last normalization layer. The downside of the short-term component is that it cannot timely detect a short-term change in data, demonstrating response latency. Also, it is insensitive to changes that only endure for a limited number (e.g., two or three) of time steps. To mitigate this issue, we can make use of the contemporary measurements of the co-evolving time series. 

\subsubsection{Co-evolving Component}
The co-evolving component, derived from the spatial correlations between time series, is advantageous for capturing instant changes in time series, which distinguishes it from the above three components. A co-evolving behavior shared across multiple time series indicates that these time series are generated from the same process. Then, we can get an estimator of this process by aggregating multiple samples drawn from it.

A key problem to be solved here is identifying which time series share the same co-evolving component. Technically, this amounts to measuring correlations between different time series. This measurement can be done either by hard-coding the correlation matrix with prior knowledge or by parameterizing and learning it. Our study adopts the latter practice, which allows for more flexibility, since many datasets do not present prior knowledge about the relationship between time series. We assign an individual attention score to every pair of time series, and then normalize the attention scores associated with the same time series via softmax to ensure that all attention scores are summed up to 1. Formally, let $\alpha_{n, n'}$ and $a_{n, n'}$ respectively denote the unnormalized and normalized attention scores between the $n^{\text{th}}$ 
 and $n'^{\text{th}}$ variable. The formula is written as follows:
\begin{align}
    a_{n, n'} &= \frac{\exp(\alpha_{n, n'})}{\sum_{j=1}^N\exp(\alpha_{n, j})}, \\
    \mu^\text{ce}_{n, t} &= \sum_{n'=1}^{N} a_{n, n'} Z^{\text{(3)}}_{n', t}\;, \\
    (\sigma^\text{ce}_{n, t})^2 &= \sum_{n'=1}^{N} a_{n, n'} (Z^{\text{(3)}}_{n', t})^2 - (\mu^\text{ce}_{n, t})^2 + \epsilon, \\
    R_{n, t} &= \frac{Z^{\text{(3)}}_{n, t} - \mu^\text{ce}_{n, t}}{\sigma^\text{ce}_{n, t}},
\end{align}
where $R_{n, t}$ denotes the residuals that cannot be modeled by any of our proposed components. This computation can be further modified to improve the scalability via the adjacency matrix learning module proposed in \cite{wu2019graph}.

The decoupled components and residual representations are sequentially concatenated to form a wide vector:
\begin{align*}
    Z_{n, t} =& [Z^\text{(1)}_{n, t}\;,  Z^\text{(2)}_{n, t}\;, Z^\text{(3)}_{n, t}\;, Z^\text{(4)}_{n, t}], \\
    H_{n, t} =& [\mu^\text{lt}_{n, t}\;, \sigma^\text{lt}_{n, t}\;, \mu^\text{se}_{n, t}\;, \sigma^\text{se}_{n, t}\;, \\
    &\mu^\text{st}_{n, t}\;, \sigma^\text{st}_{n, t}\;, \mu^\text{ce}_{n, t}\;, \sigma^\text{ce}_{n, t}].
\end{align*}

\subsection{Component Extrapolation}

\begin{figure}[t]
    \centering
    \includegraphics[width=0.75\linewidth]{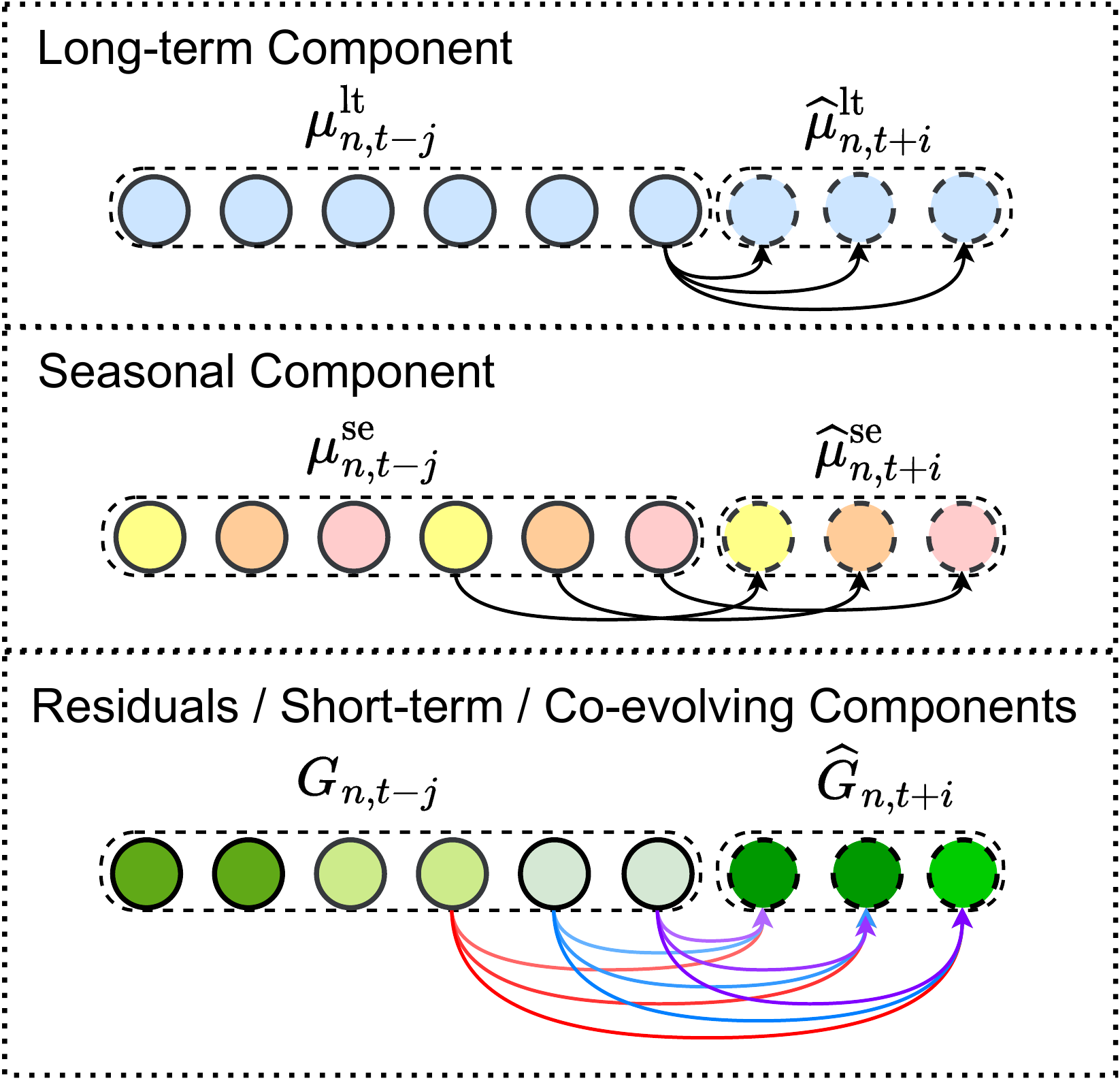}
    \caption{Component Extrapolation}
    \label{fig:extrapolate}
\end{figure}

We simulate the evolution of each component with a customized and basic model, given the heterogeneity of their dynamics. This allows for the explainability of the features being accounted for by the model and the provision of insights into the capacity of the forecasting model. With the acquired understanding of the features and the model capacity, practitioners can detect the anomaly points where the model may not present reliable results, and adopt specific measures to handle the anomalies. The components exhibit different dynamics with varying degrees of predictability, motivating us to create separate models to mimic the prospective development of their dynamics. The models are visualized in Fig. \ref{fig:extrapolate}.

\subsubsection{Regular Components}
For a short period of time in the future, the long-term component and the seasonal component change in a relatively regular behavior, so we can directly specify the law for extrapolation without introducing extra parameters

Addressing long-term component, we trivially reuse the (estimated) state of the long-term component at the current time point for the extrapolation of each future time point. 
\begin{align}
    \hat{\mu}^\text{lt}_{n, t+i} = \mu^\text{lt}_{n, t}\;, \; \hat{\sigma}^\text{lt}_{n, t+i} = \sigma^\text{lt}_{n, t}\;. \label{eq:lt_extra}
\end{align}

For the seasonal component, we also conduct replication but from the time point at the same phase as the target time point in the previous season, following its seasonal nature:
\begin{align}
    \hat{\mu}^\text{se}_{n, t+i} = \mu^\text{se}_{n, t - m + i}\;, \; \hat{\sigma}^\text{se}_{n, t+i} = \sigma^\text{se}_{n, t - m + i}\;. \label{eq:se_extra}
\end{align}

\subsubsection{Irregular Components}
The short-term component, the co-evolving component, and the residual representations vary with greater stochasticity and thereby less regular than the above two components due to their irregularity. Since the dynamics are now much more complicated, we opt to parameterize the dynamical model to allow for more flexibility than specifying a fixed heuristic law. For each of these three types of representations, we employ an auto-regressive model, predicting the representation for the $i^\text{th}$ forecast horizon based on the past $\delta$ representations. For the sake of brevity, we present the extrapolation processes of the short-term and co-evolving components together with the residuals in a single figure, given that they share the same model form:
\begin{align}
    \hat{G}_{n, t+i} = \sum_{j=0}^{\delta-1}\hat{W}_{ji} G_{n, t-j} + b_i, \label{eq:short_term_extra}
\end{align}
where $ G \in \{Z^{(l)}_{n, t+i}, \mu^\text{st}_{n, t+i}, \sigma^\text{st}_{n, t+i}, \mu^\text{ce}_{n, t+i},  \sigma^\text{ce}_{n, t+i}\}$; $\hat{W}_{ji}$, a parameter matrix of size $d_z \times d_z$, quantifies the contribution from $G_{n, t-j}$ to $\hat{G}_{n, t+i}$; $b_i$ is the bias term. $\hat{W}_{ji}$ and $b_i$ are subject to training.

We concatenate the extrapolated components, denoted as $\hat{H}_{n, t+i}$, and the residuals, $\hat{Z}_{n, t+i}$. We then model their interactions, parameterized by two learnable matrices, $\hat{W}^{(1)}$ and $\hat{W}^{(2)}$, both belonging to $\mathbb{R}^{d_z \times 12d_z}$, as follows:
\begin{align}
    \hat{S}_{n, t+i} =& \left(\hat{W}^{(1)} [\hat{Z}_{n, t+i}, \hat{H}_{n, t+i}] \right) \nonumber 
    \\ &\otimes \left(\hat{W}^{(2)} [\hat{Z}_{n, t+i}, \hat{H}_{n, t+i}] \right),
\end{align}

So far, we construct a projection from the past to the future, consisting of statistically meaningful operations.

\subsection{Component Fusion}

\begin{figure}[t]
    \centering
    \includegraphics[width=0.6\linewidth]{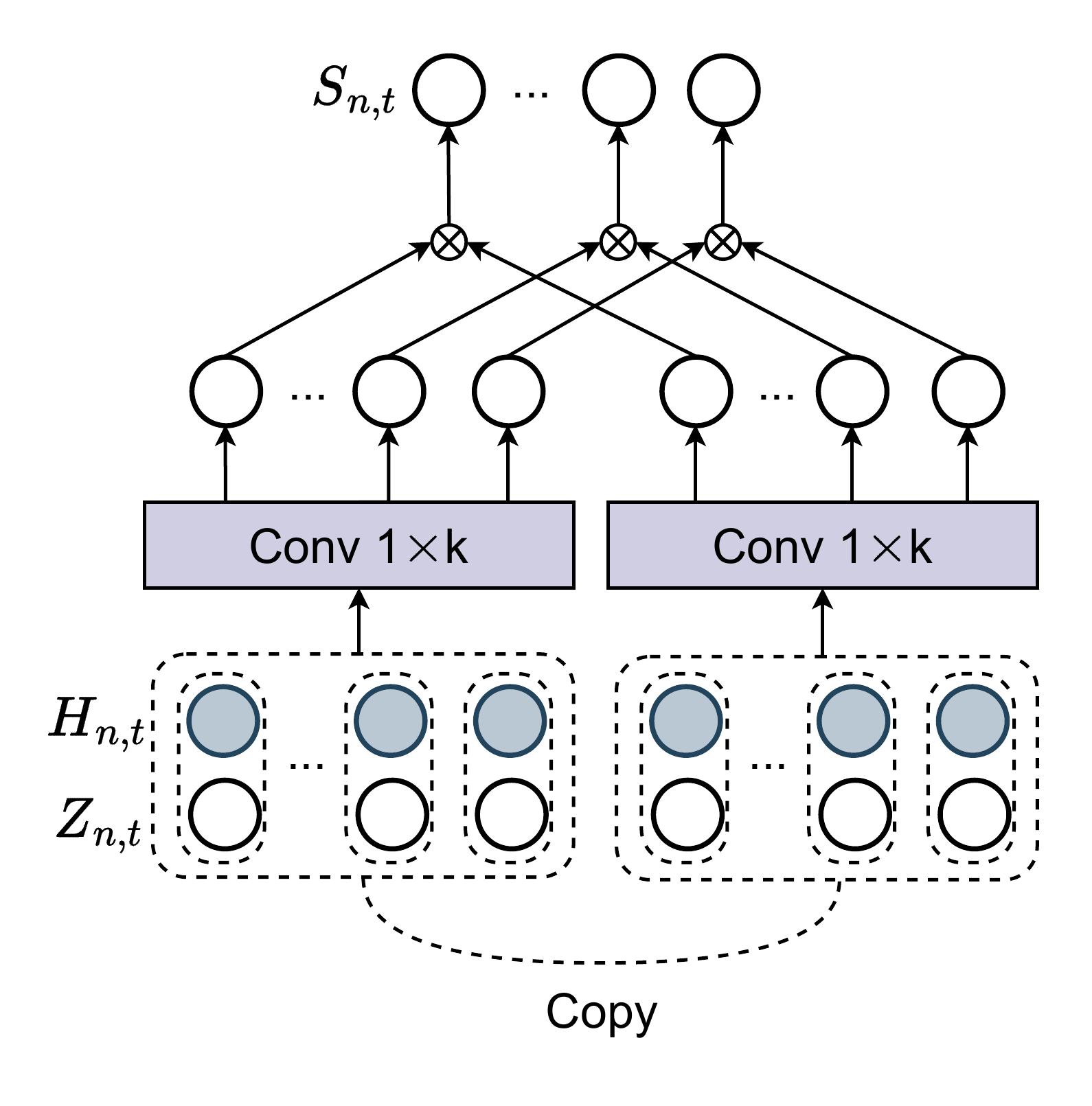}
    \caption{Component Fusion}
    \label{fig:fusion}
\end{figure}
As illustrated in Fig. \ref{fig:shifts}, there is a notable divergence in both the data distribution and the auto-correlation, observed both intra-days and inter-days. While the auto-correlation holds significance comparable to the data distribution, it has been relatively overlooked in research and discussions. At its core, the model aims to discern the auto-correlations between forward and backward observations. Consequently, these correlations are intrinsically embedded within the model parameters. Recognizing and adapting to the subtle shifts in auto-correlations can enhance forecasting accuracy.

To equip the model with the capability to discern when and how these auto-correlations evolve, structured components prove beneficial. A closer examination of Fig. \ref{fig:dist_shift} versus Fig. \ref{fig:acf_shift} and Fig. \ref{fig:con_dist_shift} versus Fig. \ref{fig:con_acf_shift} reveals a correlation between shifts in auto-correlations and shifts in data distributions. This observation implies that structured components can also serve as indicators of auto-correlations. Therefore, these components serve a dual purpose in forecasting: they capture both data distribution patterns and temporal correlations. To fully harness the capabilities of structured components, we introduce a neural module bifurcated into two branches: one dedicated to feature learning and the other to parameter learning. The outputs from these branches are then amalgamated using an element-wise multiplication operation. For the sake of simplicity, each branch employs a convolution operator, though this can be augmented with more intricate operations, such as MLP. This computational process is graphically represented in Fig. \ref{fig:fusion}, and is formally written as:
\begin{align}
    S_{n, t} =& \left(\sum_{j=0}^{k-1} W^{(1)}_{j}[Z_{n, t-j}, H_{n, t-j}]\right) \nonumber\\
    &\otimes \left(\sum_{j=0}^{k-1} W^{(2)}_{j}[Z_{n, t-j}, H_{n, t-j}] \right),
\end{align}
where $k$ is the kernel size of the convolution operator and $W^{(1)}_j, W^{(2)}_j \in \mathbb{R}^{d_z \times 12d_z}$ are learnable matrices. $S_{n, t}$ can be passed to another component estimation block as $Z^{\text{(0)}}_{n, t}$ to produce richer compositions of the structural components.

\subsection{Structural Regularization}

\begin{figure}[t]
    \centering
    \includegraphics[width=\linewidth]{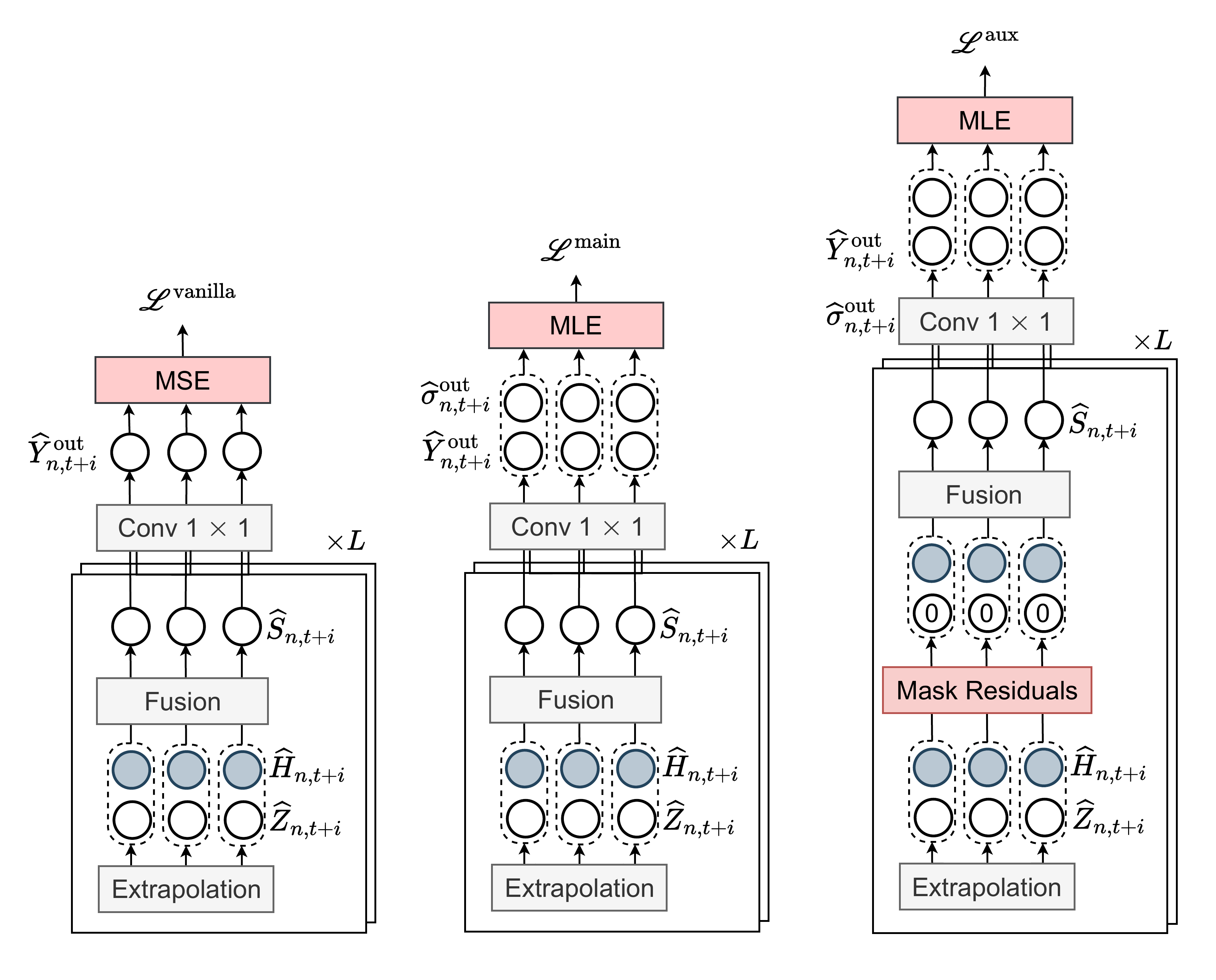}
    \caption{Structural Regularization. The term $\mathcal{L}^{\text{vanilla}}$ denotes the standard MSE loss function. On the other hand, $\mathcal{L}^{\text{main}}$ and $\mathcal{L}^{\text{aux}}$ are specifically designed to enforce regularization within the feature space, thereby ensuring a more structured representation of the data. These two loss functions work together to optimize the model's performance.}
    \label{fig:aux_framework}
\end{figure}

Conventionally, the objective function for time series forecasting aims to minimize the mean squared errors (MSE) or mean absolute errors (MAE) between the predictions and the ground truth observations. The assumption inherent to this objective is that all the variables share the same variance of 1. However, this does not enable the learned representations to be organized in a desired structure, where variables can see different degrees of variance at different times due to the time-varying scaling effects prescribed by the generative structure of time series. Instead, we opt to optimize the maximum likelihood estimate (MLE) \cite{salinas2020deepar}, which allows SCNN to improve the shaping of the structure of the representation space. In addition, an auxiliary objective function is designed to improve the nuances in feature space at the component level. We graphically contrast the two designed objective functions against the vanilla MSE loss Fig. \ref{fig:aux_framework}

We apply linear transformations to the representations output from the component extrapolation module, producing the location (i.e. mean) $\hat{Y}^{\text{out}}_{n, t+i}$ and the scale (i.e. standard deviation) $\hat{\sigma}^{\text{out}}_{n, t+i}$, where $\hat{\sigma}^{\text{out}}_{n, t+i}$ further goes through a SoftPlus function to enable itself to be non-negative. The MLE loss is written as:
\begin{align*}
\mathcal{L}^{\text{main}} &= \sum_{n=1}^N\sum_{i=1}^{T_{\text{out}}}(\log(\text{SoftPlus}(\hat{\sigma}^{\text{out}}_{n, t+i})) + \frac{(Y_{n, t+i} - \hat{Y}^{\text{out}}_{n, t+i})^2}{2(\text{SoftPlus}(\hat{\sigma}^{\text{out}}_{n, t+i}))^2}).
\end{align*}
The first term in the above loss function encourages the scaling factor to be small, and the second term penalizes the deviation between the extrapolated data and the ground truth data weighted by the inverse of the scaling factor.

Solely leveraging the above objective to learn the forecasting dynamics does not ensure robust estimation of the structured components with their contribution to the projection. The intuition is that since the residual components, especially at the bottom levels, still contain a part of the structural information, they will take a certain amount of attributions that are supposed to belong to the structured components as learning the corresponding weights for the components. Attributing improper importance to the residual components incurs considerable degradation in the model performance, once the time series data is contaminated with random noise that heavily impacts the high-frequency signal.

To approach this issue, the basic idea is to accentuate the structured components that suffer less from corruption with an additional regularizer. This regularizer works to prompt the model to achieve a reasonable forecast using purely the structured components without the need for residual components. In particular, in the forward process of a training iteration, SCNN forks another branch after the component extrapolation module. This branch starts by zero-masking all the residual components, passing only structured components through the following operations. Finally, it yields an auxiliary pair of forecast coefficients $\hat{Y}^{\text{aux}}_{n, t+i}$ and $\hat{\sigma}^{\text{aux}}_{n, t+i}$, which are also being tailored by MLE. 

 The ultimate objective to be optimized is an aggregation of all the above objective functions in a weighted fashion:
\begin{align}
    \mathcal{L} = \alpha \mathcal{L}^{\text{aux}} + \mathcal{L}^{\text{main}},
\end{align}
where $\alpha$ is the hyper-parameter that controls the importance of the corresponding objective. We use the Adam optimizer \cite{kingma2014adam} to optimize this target. 

\subsection{Discussion}

\setlength{\parindent}{10pt}

\subsubsection{Expressiveness Analysis} In modeling spatial-temporal correlations, SCNN processes data through normalization layers implemented in four distinct ways. In contrast, Transformers utilize attention layers, while MLPs depend on fully-connected layers. Essentially, a normalization layer—specifically, the averaging operator—represents a constrained form of attention and fully-connected layers. It does so by assigning equal attention scores (or weights) of $\frac{1}{l}$ to each data point within a window, where $l$ is the window size. This implies that SCNN assumes an equal contribution from every data point in the window to the component being extracted. Despite its seemingly limited expressiveness compared to fully-connected and attention layers, normalization shows a competitive, and at times superior, capacity compared to SOTA baselines in time series forecasting. The effectiveness of the normalization layer in this context is attributed to the semantically constrained nature of time series data. As indicated by \cite{zhou2021informer}, the normalized attention scores produced by the attention layer in time series data often display a sparse and regular pattern. This observation suggests that there is no need to assign distinct weights to each position in the sequence. Our research marks the first empirical demonstration that by extracting long-term, seasonal, short-term, and co-evolving components, a model can effectively capture the major spatial-temporal correlations in time series data. This approach goes beyond what has been achieved by SOTA baselines, encompassing a more comprehensive range of spatial-temporal correlations than previously explored.

\begin{figure}[t]
    \centering
    \includegraphics[width=.85\linewidth]{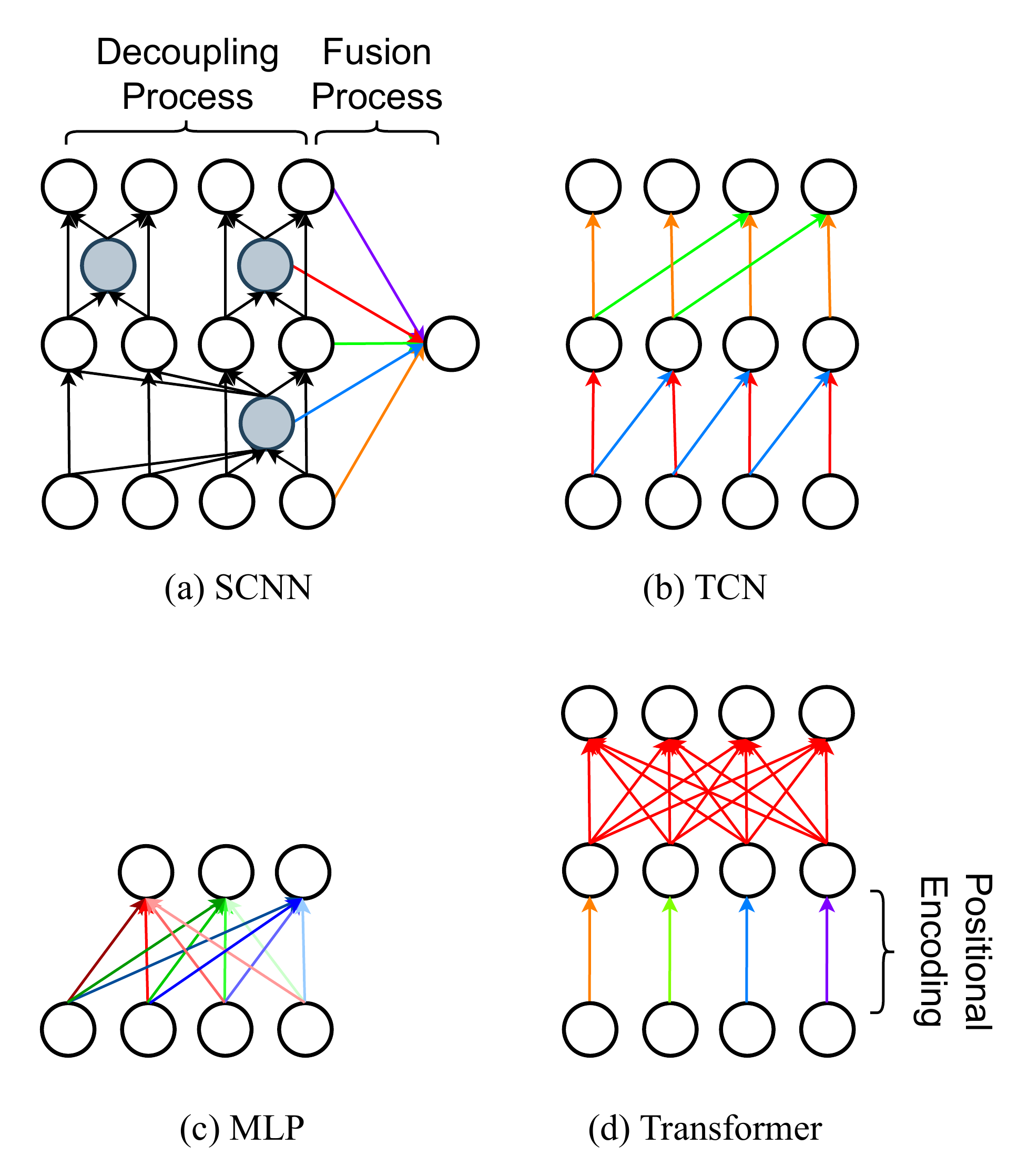}
    \caption{Data and computational flow. Each edge symbolizes an atomic operation involving a single variable situated at the tail of the edge. If an operation is parameterized, the corresponding edge is color-coded.}
    \label{fig:complexity}
\end{figure}
\subsubsection{Complexity Analysis} \label{sec:complexity} We conduct an analysis of two types of complexity associated with our model: first, the parameter complexity, which refers to the number of parameters involved in the model; and second, the computational complexity. We draw a comparison between the complexity of the SCNN and three prominent frameworks, namely the Transformer, the TCN, and the MLP.

Figure \ref{fig:complexity} provides a visual representation of the data and computational flow associated with these four frameworks. Within these diagrams, each edge symbolizes an atomic operation involving a single variable situated at the tail of the edge. If an operation is parameterized, the corresponding edge is color-coded. Edges sharing the same color denote operations utilizing the same set of learnable parameters. Within the SCNN framework, the decoupling process is carried out without parameterization, thus these edges are illustrated in black. The structured components that emerge from this process are subsequently integrated, employing component-dependent parameters.

Let's denote the number of components crafted within our model as $m$. The number of parameters within SCNN scales in proportion to the number of components inherent in the time series, which is $\mathcal{O}(m)$. This contrasts with the majority of SOTA models, where the parameter count scales with the length of the input sequence. To illustrate, TCN or WaveNet-based models necessitate at least $\mathcal{O}(\log T)$ parameters to process a sequence of length $T$; MLP or Linear Regression (LR)-based models require $\mathcal{O}(T)$ parameters; and Transformer-based models also demand $\mathcal{O}(T)$ parameters to attain SOTA performance, as demonstrated in \cite{zhang2023crossformer}. Our approach aligns with the principle that the complexity of the underlying dynamical system dictates the requisite number of parameters, regardless of the input sequence length.

Regarding the computational complexity relative to sequence length, SCNN attains a complexity of $\mathcal{O}(Tm)$. This stands in contrast to alternative methods such as the MLP, which achieves a complexity of $\mathcal{O}(Th)$, with $h$ representing the number of units in the hidden layer, which is typically large. The Transformer model yields a complexity of $\mathcal{O}(T^2)$, while the TCN model reaches a complexity of $\mathcal{O}(T \log T)$. Therefore, in terms of computational complexity with respect to sequence length, the SCNN proves to be the most efficient model, particularly when the structured component is estimated in a moving average manner. This observation underscores the advantage of SCNN in scenarios where computational efficiency and scalability are critical considerations.

Notably, we can further reduce the complexity of an inference step to $\mathcal{O}(m)$ by approximating the structured component using a moving average approach. A significant feature of SCNN is its statistically interpretable operations, which augment its scalability when applied to online testing. During the online testing phase, each model is tasked with processing each sample sequentially as new observations arrive, contrasting with the parallel processing of multiple samples during the offline training phase. SOTA methods typically tackle this scenario by dynamically selecting the preceding $T_\text{in}$ consecutive observations as input, consistent with the training input format. In contrast, SCNN uniquely requires only the current observation and previously estimated components as input, thereby eliminating a significant amount of redundant computations involved for manipulating the historical observations. The required computation only involves dynamically updating the structured components with the available observations through an exponential moving average.

\section{Evaluation}

In this section, we conduct extensive experiments on three common datasets to validate the effectiveness of SCNN from various aspects.

\subsection{Experiment Setting}
\subsubsection{Datasets}
To evaluate the performance of our model, we conduct experiments on three popular spatial-temporal forecasting datasets, namely BikeNYC\footnote{https://ride.citibikenyc.com/system-data}, PeMSD7\footnote{https://pems.dot.ca.gov/} and Electricity\footnote{https://archive.ics.uci.edu/ml/datasets/ElectricityLoadDiagrams
 20112014}. The statistics and the experiment settings regarding the three datasets are reported in Table \ref{tab:dataset}. Long-term time series forecasting (LTSF) is an emerging application that focuses on making predictions for an extensively long period, e.g. hundreds of horizons, into the future, where the ability with long-term forecasting of the model can be revealed. To holistically benchmark SCNN, we also evaluate it on 7 popular real-world LTSF tasks, including Weather, Traffic, ELC and 4 ETT datasets (ETTh1, ETTh2, ETTm1, ETTm2)\footnote{https://github.com/thuml/Time-Series-Library}. We adopt the same data pre-processing strategy as most of the current works \cite{wu2019graph, wu2020connecting}, where the TS data of each variable is individually standardized.

\begin{table*}[tb]
\caption{Statistics of spatial-temporal datasets.}
\label{tab:dataset}
\begin{tabular}{l|ccc|ccccccc}
\hline
Tasks           & \multicolumn{3}{c|}{Spatial-temporal Forecasting} & \multicolumn{7}{c}{Long-term Time Series Forecasting}                                                \\ \hline
Datasets        & Electricity      & PeMSD7          & BikeNYC      & ELC          & Traffic      & ETTh1       & ETTh2       & ETTm1        & ETTm2        & Weather      \\ \hline
Sample rate     & 1 hour           & 30 minutes      & 1 hour       & 1 hour       & 1 hour       & 1 hour      & 1 hour      & 15 minutes   & 15 minutes   & 10 minutes   \\
\# Variate      & 336              & 228             & 128          & 321          & 862          & 7           & 7           & 7            & 7            & 21           \\
Training size   & 1848             & 1632            & 3912         & $\sim$18,000 & $\sim$12,000 & $\sim$8,000 & $\sim$8,000 & $\sim$34,000 & $\sim$34,000 & $\sim$36,000 \\
Validation size & 168              & 240             & 240          & $\sim$2,500  & $\sim$1,600  & $\sim$2,700 & $\sim$2,700 & $\sim$11,000 & $\sim$11,000 & $\sim$5,000  \\
Testing size    & 168              & 240             & 240          & $\sim$5,000  & $\sim$3,300  & $\sim$2,700 & $\sim$2,700 & $\sim$11,000 & $\sim$11,000 & $\sim$10,000 \\
Input length    & 144              & 288             & 144          & 168          & 168          & 168         & 168         & 384          & 384          & 432          \\
Output length   & \multicolumn{3}{c|}{3}                            & \multicolumn{7}{c}{3, 24, 96, 192}                                                                   \\ \hline
\end{tabular}
\end{table*}
 
 \begin{table*}[!ht]
\small
\centering
\caption{Performance on the BikeNYC dataset}
\label{tab:bike}
\begin{tabular}{l|c|c|c|c|c|c|c|c|c}
\hline
\multirow{2}{*}{Model}       &  \multicolumn{3}{c|}{MAPE (\%)}      & \multicolumn{3}{c|}{MAE}  & \multicolumn{3}{c}{RMSE}  \\ \cline{2-10}
 & Horizon 1  & Horizon 2 & Horizon 3 & Horizon 1 & Horizon 2 &  Horizon 3 &  Horizon 1 & Horizon 2  & Horizon 3  \\ \hline
LSTNet        & 21.2& 22.3& 23.8 & 2.71& 2.91& 3.15 & 5.80& 6.34& 6.97  \\
StemGNN       & 19.0& 20.8& 22.5 & 2.50& 2.74& 2.93 & 5.25& 6.09& 6.62  \\ 
AGCRN         & 17.4& 18.8& 20.5 & 2.28& 2.50& 2.68 & 4.74& 5.50& 5.97  \\
GW & 18.2& 19.5& 20.9 & 2.35& 2.57& 2.75 & 4.83& 5.56& 6.06  \\
MTGNN         & 18.0& 19.5& 20.9 & 2.35& 2.57& 2.73 & 4.87& 5.69& 6.18  \\
SCINet         & 17.9& 19.8& 21.4 & 2.38& 2.68& 2.94 & 4.88& 5.78& 6.60  \\
STG-NCDE         & 18.7& 20.6& 22.2 & 2.40& 2.67& 2.90 & 5.04& 5.86& 6.56  \\
GTS         & 20.6& 23.6& 26.7 & 2.38& 2.58& 2.74 & 4.85& 5.53& 6.01  \\
ST-Norm       & \underline{17.3}& \underline{18.6}& \underline{19.9} & \underline{2.26} & \underline{2.46} & \underline{2.62} & \underline{4.66} & \underline{5.38} & \underline{5.84}  \\
SCNN     & \textbf{16.5} & \textbf{17.3}& \textbf{18.4} & \textbf{2.13} & \textbf{2.27}& \textbf{2.40} & \textbf{4.44} & \textbf{5.02} & \textbf{5.42}  \\
 \hline
 Imp & +4.6\% & +6.9\%& +7.5\% & +5.7\% & +7.7\% & +8.3\% & +4.7\% & +6.6\% & +7.1\%  \\
\hline
\end{tabular}
\end{table*}

\begin{table*}[!ht]
\small
\centering
\caption{Performance on the PeMSD7 dataset}
\label{tab:pems}
\begin{tabular}{l|c|c|c|c|c|c|c|c|c}
\hline
\multirow{2}{*}{Model}       &  \multicolumn{3}{c|}{MAPE (\%)}      & \multicolumn{3}{c|}{MAE}  & \multicolumn{3}{c}{RMSE}  \\ \cline{2-10}
 & Horizon 1  & Horizon 2 & Horizon 3 & Horizon 1 & Horizon 2 &  Horizon 3 &  Horizon 1 & Horizon 2  & Horizon 3  \\ \hline
LSTNet        & 7.48& 7.77& 8.19          & 3.58& 3.71& 3.90         & 6.24& 6.40& 6.64         \\

StemGNN       & 5.50& 7.33& 8.09          & 2.65& 3.49& 3.84         & 4.55& 5.99& 6.53         \\
AGCRN         & 4.97& 6.49& 7.21          & 2.35& 3.02 & \underline{3.34}         & 4.29& 5.57& 6.10         \\
GW & 5.02& 6.56& 7.10          & 2.39& 3.10& 3.35         & 4.28& \underline{5.51} & \underline{5.94}         \\
MTGNN         & 5.32& 6.71& 7.31          & 2.57& 3.15& 3.44         & 4.36& 5.56& 6.01         \\
SCINet         & 5.16& 6.72& 7.23 & 2.47 & 3.18& 3.45 & 4.31& 5.60& 6.05  \\
STG-NCDE         & 4.94& 6.63& 7.58 & 2.32 & 3.06& 3.47 & 4.42& 5.91& 6.70  \\
GTS         & 5.35& 6.97& 7.70 & 2.53& 3.26& 3.58 & 4.42& 5.74& 6.30  \\
ST-Norm       & \underline{4.76}& \underline{6.27}& \underline{7.03}          & \underline{2.27}& \underline{2.98} & 3.36         & \underline{4.21}& 5.54& 6.07         \\
SCNN     &         \textbf{4.47} & \textbf{5.92} & \textbf{6.50}                  &           \textbf{2.10} & \textbf{2.75} & \textbf{2.99}               &       \textbf{4.06}& \textbf{5.29} & \textbf{5.76}                   \\\hline
Imp & +6\% & +5.5\% & +7.5\% & +7.4\% & +7.7\% & +10\% & +3.5\% & +3.9\% & +3.7\%  \\
\hline
\end{tabular}
\end{table*}

\begin{table*}[!ht]
\small
\centering
\caption{Performance on the Electricity dataset}
\label{tab:elec}
\begin{tabular}{l|c|c|c|c|c|c|c|c|c}
\hline
\multirow{2}{*}{Model}       &  \multicolumn{3}{c|}{MAPE (\%)}      & \multicolumn{3}{c|}{MAE}  & \multicolumn{3}{c}{RMSE}  \\ \cline{2-10}
 & Horizon 1  & Horizon 2 & Horizon 3 & Horizon 1 & Horizon 2 &  Horizon 3 &  Horizon 1 & Horizon 2  & Horizon 3  \\ \hline
LSTNet   & 22.4& 23.0& 24.8          & 31.1& 31.8& 33.8         & 61.2& 62.6& 66.8         \\
StemGNN  & 10.8& 13.7& 15.7          & 15.5& 19.6& 22.3         & 34.3& 43.9& 49.7         \\
AGCRN    & 11.4& 15.6& 18.0          & 17.3& 23.0& 26.4         & 38.9& 51.2& 57.9         \\
GW       & 11.3& 15.6& 17.3          & 16.3& 22.0& 24.3         & 32.5& 43.6& 48.7         \\
MTGNN    & 10.2& 13.9& 16.0          & 14.4& 19.4& 22.2         & \underline{29.8} & \underline{40.3} & \underline{46.5}         \\
SCINet         & 10.3 & 13.7 & 16.2 & 14.7 & 20.2 & 23.6 & 33.2 & 44.0& 51.7  \\
STG-NCDE         & 10.9& 14.2& 16.0 & 16.2& 21.1& 23.7 & 36.3& 47.7& 52.9  \\
GTS         & \underline{10.0} & 14.2 & 17.1 & \underline{14.1} & \underline{19.0} & \underline{22.1} & 31.6& 42.5 & 48.2  \\
ST-Norm  & 10.2 & \underline{13.2} & \underline{15.3}          & 15.2& 19.8& 22.8         & 32.3& 42.9& 50.2         \\
SCNN  &   \textbf{7.69} & \textbf{10.5} & \textbf{12.2} &         \textbf{11.1} & \textbf{15.0} & \textbf{17.3}               &                \textbf{23.9} & \textbf{32.9} & \textbf{38.4} \\ \hline
Imp & +23.1\% & +20.4\% & +20.2\% & 21.9\% & +20.9\% & +21.7\% & +19.7\% & +18.3\% & +17.4\%  \\
\hline
\end{tabular}
\end{table*}

\begin{table*}[tb]
\centering
\caption{Performance on LTSF datasets.}
\label{tab:LTSF}
\begin{tabular}{lc|cc|cc|cc|cc|cc|cc|cc|cc}
\hline
\multicolumn{2}{l|}{Models}                             & \multicolumn{2}{c|}{\begin{tabular}[c]{@{}c@{}}SCNN\\ (Ours)\end{tabular}} & \multicolumn{2}{c|}{\begin{tabular}[c]{@{}c@{}}iTransformer\\ (2023)\end{tabular}} & \multicolumn{2}{c|}{\begin{tabular}[c]{@{}c@{}}PatchTST\\ (2023)\end{tabular}} & \multicolumn{2}{c|}{\begin{tabular}[c]{@{}c@{}}TimesNet\\ (2023)\end{tabular}} & \multicolumn{2}{c|}{\begin{tabular}[c]{@{}c@{}}Crossformer\\ (2023)\end{tabular}} & \multicolumn{2}{c|}{\begin{tabular}[c]{@{}c@{}}DLinear\\ (2023)\end{tabular}} & \multicolumn{2}{c|}{\begin{tabular}[c]{@{}c@{}}Triformer\\ (2022)\end{tabular}} & \multicolumn{2}{c}{\begin{tabular}[c]{@{}c@{}}Autoformer\\ (2021)\end{tabular}} \\ \hline
\multicolumn{2}{l|}{Metric}                             & \multicolumn{1}{l}{MSE}             & \multicolumn{1}{l|}{MAE}             & \multicolumn{1}{l}{MSE}                 & \multicolumn{1}{l|}{MAE}                 & \multicolumn{1}{l}{MSE}               & \multicolumn{1}{l|}{MAE}               & \multicolumn{1}{l}{MSE}               & \multicolumn{1}{l|}{MAE}               & \multicolumn{1}{l}{MSE}                 & \multicolumn{1}{l|}{MAE}                & \multicolumn{1}{l}{MSE}               & \multicolumn{1}{l|}{MAE}              & MSE                                    & MAE                                    & \multicolumn{1}{l}{MSE}                & \multicolumn{1}{l}{MAE}                \\ \hline
\multicolumn{1}{l|}{\multirow{3}{*}{\rotatebox[origin=c]{90}{ELC}}} & 3   & \underline{0.059}                               & \underline{0.152}                                & \underline{0.059}                                   & \underline{0.152}                                    & 0.063                                 & 0.160                                  & 0.119                                 & 0.232                                  & \textbf{0.058}                                   & \textbf{0.151}                                   & 0.077                                 & 0.175                                 & 0.075                                  & 0.176                                  & 0.147                                  & 0.273                                  \\
\multicolumn{1}{l|}{}                             & 24  & \underline{0.096}                               & \underline{0.192}                                & \textbf{0.094}                                   & \textbf{0.189}                                    & 0.100                                 & 0.197                                  & 0.135                                 & 0.245                                  & 0.098                                   & 0.195                                   & 0.122                                 & 0.221                                 & 0.108                                  & 0.208                                  & 0.168                                  & 0.286                                  \\
\multicolumn{1}{l|}{}                             & 96  & 0.145                               & 0.238                                & 0.133                                   & 0.229                                    & 0.136                                 & 0.230                                  & 0.169                                 & 0.272                                  & 0.136                                   & 0.238                                   & 0.154                                 & 0.248                                 & 0.144                                  & 0.241                                  & 0.186                                  & 0.301                                  \\
\multicolumn{1}{l|}{}                             & 192 & 0.160                               & 0.252                                & \underline{0.157}                                   & \underline{0.251}                                    & \textbf{0.153}                                 & \textbf{0.243}                                  & 0.191                                 & 0.288                                  & 0.158                                   & 0.255                                   & 0.168                                 & 0.260                                 & 0.163                                  & 0.259                                  & 0.218                                  & 0.328                                  \\ \hline
\multicolumn{1}{l|}{\multirow{3}{*}{\rotatebox[origin=c]{90}{Traffic}}}     & 3   & \textbf{0.246}                               & \textbf{0.194}                                & \underline{0.250}                                   & 0.197                                    & 0.252                                 & \underline{0.195}                                  & 0.510                                 & 0.283                                  & 0.289                                   & 0.210                                   & 0.331                                 & 0.255                                 & 0.320                                  & 0.221                                  & 0.524                                  & 0.344                                  \\
\multicolumn{1}{l|}{}                             & 24  & \textbf{0.316}                               & \underline{0.234}                                & \textbf{0.316}                                   & \underline{0.234}                                    & \underline{0.323}                                 & \textbf{0.229}                                  & 0.531                                 & 0.293                                  & 0.335                                   & 0.231                                   & 0.402                                 & 0.281                                 & 0.383                                  & 0.251                                  & 0.548                                  & 0.335                                  \\
\multicolumn{1}{l|}{}                             & 96  & 0.386                               & 0.271                                & \underline{0.375}                                   & \underline{0.261}                                    & \textbf{0.371}                                 & \textbf{0.251}                                  & 0.602                                 & 0.319                                  & 0.392                                   & 0.272                                  & 0.452                                 & 0.302                                 & 0.438                                  & 0.273                                  & 0.623                                  & 0.350                                  \\
\multicolumn{1}{l|}{}                             & 192 & 0.416                               & 0.280                                & \underline{0.396}                                   & \underline{0.268}                                    & \textbf{0.394}                                 & \textbf{0.260}                                  & 0.615                                 & 0.321                                  & 0.423                                       & 0.269                                       & 0.465                                 & 0.304                                 & 0.482                                  & 0.297                                  & 0.669                                  & 0.410                                  \\ \hline
\multicolumn{1}{l|}{\multirow{3}{*}{\rotatebox[origin=c]{90}{ETTh1}}}      & 3   & \textbf{0.146}                               & \textbf{0.242}                                & 0.165                                   & 0.262                                    & \underline{0.148}                                 & 0.248                                  & 0.272                                 & 0.337                                  & 0.142                                   & 0.241                                   & 0.224                                 & 0.310                                 & 0.203                                  & 0.298                                  & 0.299                                  & 0.382                                  \\
\multicolumn{1}{l|}{}                             & 24  & \underline{0.304}                               & \textbf{0.353}                                & 0.320                                   & 0.367                                    & \textbf{0.299}                                 & \underline{0.355}                                  & 0.352                                 & 0.393                                  & 0.318                                   & 0.366                                   & 0.329                                 & 0.372                                 & 0.332                                  & 0.380                                  & 0.442                                  & 0.466                                  \\
\multicolumn{1}{l|}{}                             & 96  & \underline{0.379}                               & \textbf{0.398}                                & 0.388                                   & 0.407                                    & \textbf{0.376}                                 & \underline{0.401}                                  & 0.402                                 & 0.421                                  & 0.381                                   & 0.405                                   & 0.388                                 & 0.404                                 & 0.395                                  & 0.415                                  & 0.456                                  & 0.469                                  \\
\multicolumn{1}{l|}{}                             & 192 & \textbf{0.427}                               & \textbf{0.423}                                & 0.432                                   & 0.432                                    & \underline{0.428}                                 & \underline{0.427}                                  & 0.464                                 & 0.459                                  & 0.433                                   & 0.431                                   & 0.434                                 & 0.428                                 & 0.450                                  & 0.447                                  & 0.505                                  & 0.491                                  \\ \hline
\multicolumn{1}{l|}{\multirow{3}{*}{\rotatebox[origin=c]{90}{ETTh2}}}       & 3   & \textbf{0.079}                               & \underline{0.177}                                & 0.088                                   & 0.193                                    & \underline{0.081}                                 & 0.178                                  & 0.119                                 & 0.232                                  & \textbf{0.079}                                   & \textbf{0.176}                                   & 0.109                                 & 0.213                                 & 0.104                                  & 0.209                                  & 0.203                                  & 0.310                                  \\
\multicolumn{1}{l|}{}                             & 24  & \textbf{0.163}                               & \textbf{0.253}                                & 0.187                                   & 0.278                                    & \underline{0.176}                                 & \underline{0.264}                                  & 0.210                                 & 0.301                                  & 0.180                                   & 0.271                                   & 0.179                                 & 0.266                                 & 0.193                                  & 0.279                                  & 0.318                                  & 0.393                                  \\
\multicolumn{1}{l|}{}                             & 96  & \textbf{0.289}                               & \textbf{0.340}                                & 0.306                                   & 0.356                                    & \underline{0.294}                                 & \underline{0.345}                                  & 0.340                                 & 0.379                                  & 0.328                                   & 0.376                                   & \textbf{0.289}                                 & \textbf{0.340}                                 & 0.305                                  & 0.351                                  & 0.378                                  & 0.417                                  \\
\multicolumn{1}{l|}{}                             & 192 & \textbf{0.356}                               & \textbf{0.388}                                & 0.397                                   & 0.414                                    & 0.365                                 & \underline{0.400}                                  & 0.402                                 & 0.417                                  & 0.396                                   & 0.416                                   & \underline{0.363}                                 & \textbf{0.388}                                 & 0.393                                  & 0.407                                  & 0.437                                  & 0.452                                  \\ \hline
\multicolumn{1}{l|}{\multirow{3}{*}{\rotatebox[origin=c]{90}{ETTm1}}}       & 3   & 0.058                               & \underline{0.151}                                & 0.062                                   & 0.161                                    & \textbf{0.056}                                 & \textbf{0.149}                                  & 0.067                                 & 0.168                                  & \underline{0.057}                                   & \underline{0.151}                                   & 0.062                                 & 0.156                                 & 0.081                                  & 0.185                                  & 0.227                                  & 0.315                                  \\
\multicolumn{1}{l|}{}                             & 24  & \textbf{0.193}                               & \textbf{0.270}                                & 0.215                                   & 0.297                                    & \underline{0.196}                                 & \underline{0.277}                                  & 0.201                                 & 0.282                                  & 0.209                                   & 0.282                                   & 0.213                                 & 0.284                                 & 0.206                                  & 0.288                                  & 0.466                                  & 0.446                                  \\
\multicolumn{1}{l|}{}                             & 96  & \textbf{0.287}                               & \textbf{0.339}                                & 0.313                                   & 0.363                                    & \underline{0.299}                                 & 0.347                                  & 0.324                                 & 0.370                                  & 0.319                                   & 0.355                                   & 0.304                                 & \underline{0.345}                                 & 0.301                                  & 0.356                                  & 0.471                                  & 0.445                                  \\
\multicolumn{1}{l|}{}                             & 192 & \textbf{0.327}                               & \underline{0.366}                                & 0.349                                   & 0.383                                    & 0.351                                 & 0.381                                  & 0.371                                 & 0.399                                  & 0.387                                   & 0.394                                   & \underline{0.337}                                 & \textbf{0.364}                                 & 0.338                                  & 0.373                                  & 0.566                                  & 0.498                                  \\ \hline
\multicolumn{1}{l|}{\multirow{3}{*}{\rotatebox[origin=c]{90}{ETTm2}}}       & 3   & \textbf{0.042}                               & \textbf{0.119}                                & \underline{0.044}                                   & 0.127                                    & \textbf{0.042}                                 & \underline{0.120}                                  & 0.051                                 & 0.143                                  & \textbf{0.042}                                   & \underline{0.120}                                   & \underline{0.044}                                 & 0.125                                 & 0.056                                  & 0.143                                  & 0.120                                  & 0.234                                  \\
\multicolumn{1}{l|}{}                             & 24  & \underline{0.095}                               & \underline{0.192}                                & 0.104                                   & 0.207                                    & \textbf{0.093}                                 & \textbf{0.191}                                  & 0.108                                 & 0.210                                  & 0.098                                   & 0.197                                   & \underline{0.095}                                 & 0.194                                 & 0.102                                  & 0.201                                  & 0.151                                  & 0.262                                  \\
\multicolumn{1}{l|}{}                             & 96  & \textbf{0.163}                               & \textbf{0.250}                                & 0.188                                   & 0.274                                    & \underline{0.169}                                 & 0.261                                  & 0.192                                 & 0.278                                  & 0.177                                   & 0.264                                   & \textbf{0.163}                                 & \underline{0.252}                                 & 0.173                                  & 0.260                                  & 0.231                                  & 0.317                                  \\
\multicolumn{1}{l|}{}                             & 192 & \underline{0.221}                               & \underline{0.292}                                & 0.244                                   & 0.312                                    & 0.231                                 & 0.300                                  & 0.241                                 & 0.315                                  & 0.231                                   & 0.303                                   & \textbf{0.217}                                 & \textbf{0.288}                                 & 0.234                                  & 0.300                                  & 0.348                                  & 0.392                                  \\ \hline
\multicolumn{1}{l|}{\multirow{3}{*}{\rotatebox[origin=c]{90}{Weather}}}     & 3   & \underline{0.046}                               & 0.066                                & \underline{0.046}                                   & \textbf{0.062}                                    & \textbf{0.045}                                 & \underline{0.064}                                  & 0.055                                 & 0.091                                  & \textbf{0.045}                                   & \underline{0.064}                                   & 0.048                                 & 0.074                                 &              0.055                          &       0.076                                 & 0.054                                  & 0.087                                  \\
\multicolumn{1}{l|}{}                             & 24  & \textbf{0.089}                               & \textbf{0.120}                                & 0.097                                   & 0.130                                    & \underline{0.093}                                 & \underline{0.121}                                  & 0.100                                 & 0.142                                  & \underline{0.093}                                   & 0.134                                   & 0.109                                 & 0.209                                 &              0.096                          &        0.132                                & 0.119                                  & 0.167                                  \\
\multicolumn{1}{l|}{}                             & 96  & \textbf{0.142}                               & \textbf{0.192}                                & 0.168                                   & 0.216                                    & 0.163                                 & \underline{0.207}                                  & 0.173                                 & 0.221                                  & \underline{0.155}                                   & 0.212                                   & 0.171                                 & 0.224                                 &             0.153                           &          0.207                              & 0.201                                  & 0.242                                  \\
\multicolumn{1}{l|}{}                             & 192 & \textbf{0.188}                               & \textbf{0.232}                                & 0.213                                   & 0.258                                    & \underline{0.195}                                 & \underline{0.244}                                  & 0.215                                 & 0.265                                  & 0.213                                   & 0.271                                   & 0.214                                 & 0.259                                 &                  0.204                      &           0.253                             & 0.392                                  & 0.436                                  \\ \hline
\multicolumn{2}{c|}{1$^\text{st}$ Count} & \textbf{16} & \textbf{15} & 2 & 2 & \underline{9} & \underline{6} & 0 & 0 & 4 & 2 & 3 & 4 & 0 & 0 & 0 & 0\\ \hline
\end{tabular}
\end{table*}

\subsubsection{Network Setting}
The input length is set to a multiple of the season length, so that sufficient frames governed by approximately the same seasonal and long-term components can be gathered to yield estimation without much deviation. The layer number is set to $4$; The number of hidden channels $d$ is $8$; $\Delta$ is set to the same quantity as the length of the input sequence; $\delta$ is set to 8; the kernel size of the causal convolution $k$ is configured as 2. In the training phase, the batch size is $8$; the weight for the auxiliary objective $\alpha$ is 0.5; the learning rate of the Adam optimizer is $0.0001$. We also test other configurations in the hyper-parameter analysis. 

\noindent\textbf{The Choice of $\epsilon$:} Previous studies \cite{kim2021reversible, liu2022non, deng2021st} let the $\epsilon$ employed in decoupling, e.g., Eq. \ref{eq:epsilon}, to be an infinitesimal value, e.g. 0.00001, for the purpose of avoiding the division-by-zero issue. We find that this trick, however, incurs an unstable optimization process in some cases, resulting in a sub-optimal solution on the parameter space. Imagine a time series that rarely receives non-zero measurements which can be viewed as unpredictable noises. The standard deviation of this time series would be very small, leading its inverse to be exceptionally large. As a result, the noises would be undesirably magnified, driving the model to fit these chaotic patterns without any predictable structure. To alleviate this dilemma, our study sets $\epsilon$ as 1, which, on the one hand, can prevent the explosion of noises and, on the other hand, cannot dominate the original scaling factor. This simple trick is also employed by \cite{salinas2020deepar}, but they only used it to preprocess the time series data.

\subsubsection{Evaluation Metrics} We validate our model by root mean squared error (RMSE), mean absolute error (MAE) and mean absolute percentage error (MAPE). We repeat the experiment ten times for each model on each dataset and report the mean of the results.

\subsection{Baseline Models}
\subsubsection{Spatial-temporal Forecasting Baselines}
We compare SCNN with the following spatial-temporal forecasting models on the 3 spatial-temporal datasets: 

\begin{itemize} 
\item \textbf{LSTNet\cite{lai2018modeling}.} LSTNet uses CNN to extract local features and uses RNN to capture long-term dependencies. It also employs a classical auto-regressive model to address scale-insensitive limitations.
\item \textbf{StemGNN\cite{cao2020spectral}.} StemGNN models spatial and temporal dependencies in the spectral domain. 
\item \textbf{GW\cite{wu2019graph}.} GW proposes an adaptive graph learning module that progressively recovers the spatial correlations during training. In addition, it employs Wavenet to handle correlations in the temporal domain.
\item \textbf{MTGNN\cite{wu2020connecting}.} MTGNN designs a graph learning module that integrates external knowledge like variable attributes to learn uni-directed relations among variables.
\item \textbf{AGCRN\cite{bai2020adaptive}.} AGCRN develops two adaptive modules to build interactions between the variables. In addition, it selects RNN to undertake the job of modeling temporal evolution. 
\item \textbf{SCINet\cite{liu2022SCINet}} SCINet proposes a downsample-convolve-interact architecture which is beneficial for integrating multi-resolution features.
\item \textbf{STG-NCDE\cite{choi2022graph}.} STG-NCDE takes advantage of Neural Controlled Differential Equations (NCDEs) to conduct spatial-temporal processing. It generalizes canonical RNN and CNN to continuous RNN and GCN based on NCDEs.
\item \textbf{GTS\cite{shang2021discrete}.} GTS proposes a structure learning module to learn pairwise relationships between the variables. 
\item \textbf{ST-Norm\cite{deng2021st}.} ST-Norm designs two normalization modules to refine the high-frequency and local components separately from MTS data. 
\end{itemize}

In order to make the comparison fair, all the competing models are fed with the same number of preceding frames as SCNN. We find that this extension of input horizons can bring performance gain to various degrees.

\subsubsection{Long-term Time Series Forecasting Baselines}
We also compare DSCNN with the following LTSF models on the 7 LTSF datasets: 
\begin{itemize}
\item \textbf{Autoformer \cite{wu2021autoformer}.} To counter the problem with point-wise self-attention of neglecting sequence-wise behavior, Autoformer innovates a attention mechanism based on auto-correlation, a measurement of the series-wise similarities between the time series and its lagged copies.
\item \textbf{Triformer \cite{RazvanIJCAI2022}.} Employing variable-specific model parameters, Triformer enables to capture distinct temporal patterns from different variables. Moreover, it features a triangular, multi-layer structure that applies attention mechanism on the patch level to reduce the computational complexity.
\item \textbf{DLinear \cite{Zeng2022AreTE}.} DLinear is an embarrassingly simple one-layer linear model, serving as a basic but reliable and strong benchmark to compete with.
\item \textbf{Crossformer \cite{cao2020spectral}.} Crossformer segments time series into patches, enabling to maintain local semantics of time series. Besides, Crossformer adopts two-stage attention mechanism to respectively capture cross-time and cross-series dependencies. 
\item \textbf{TimesNet \cite{wu2023timesnet}.} TimesNet transforms the 1D time series into a set of 2D tensors based on multiple periods, making the intraperiod- and interperiod-variations to be easily modeled by 2D kernels.
\item \textbf{PatchTST \cite{nie2023a}.} PatchTST segments time series into subseries-level patches which are served as input tokens to Transformer. In addition, instead of mixing the series together, PatchTST processes different series disjointly with shared parameters.
\item \textbf{iTransformer \cite{liu2023itransformer}.} Inverting the conventional roles of MLP and attention mechanism within Transformer, iTransformer applies MLP to the temporal domain, while applying self-attention mechanism to the spatial domain. 
\end{itemize}

For implementing state-of-the-art models (SOTAs), we adhere to the default settings as provided in the Time-Series-Library.

\subsection{Performance Comparison}
\subsubsection{Spatial-temporal Forecasting}
The experiment results on the three spatial-temporal datasets are respectively reported in Table \ref{tab:bike}, Table \ref{tab:pems}, and Table \ref{tab:elec}. It is evident that the performance of SCNN surpasses that of the baseline models by 4\% to 20\%, especially when performing forecasts for multi-step ahead. This is because SCNN can extract the structured components with a well-conditioned deviation. As we know, raw data contains much noise, unavoidably interfering with the quality of the extracted components.  SCNN can effectively deal with this issue according to the central limit theorem. In contrast, all the benchmark models, except ST-Norm, did not explicitly account for the structured components. For example, SCINet, one of the most up-to-date state-of-the-art models, struggled to achieve competitive performance in short-term MTS forecasting, due to its deficiency in adapting to the short-term distribution shift even with the enhancement of RevIN module proposed by \cite{kim2021reversible}. GTS, GW, MTGNN and AGCRN were capable of learning the spatial correlations across the variables to estimate the translating effect of a co-evolving component, but were insusceptible to the changes in its scaling effects over time. ST-Norm could decouple the long-term component and the global component (a reduced form of co-evolving component), but did not introduce the constraint to the structure of feature space.  

\begin{figure}[thb]
    \centering
    \includegraphics[width=0.9\linewidth]{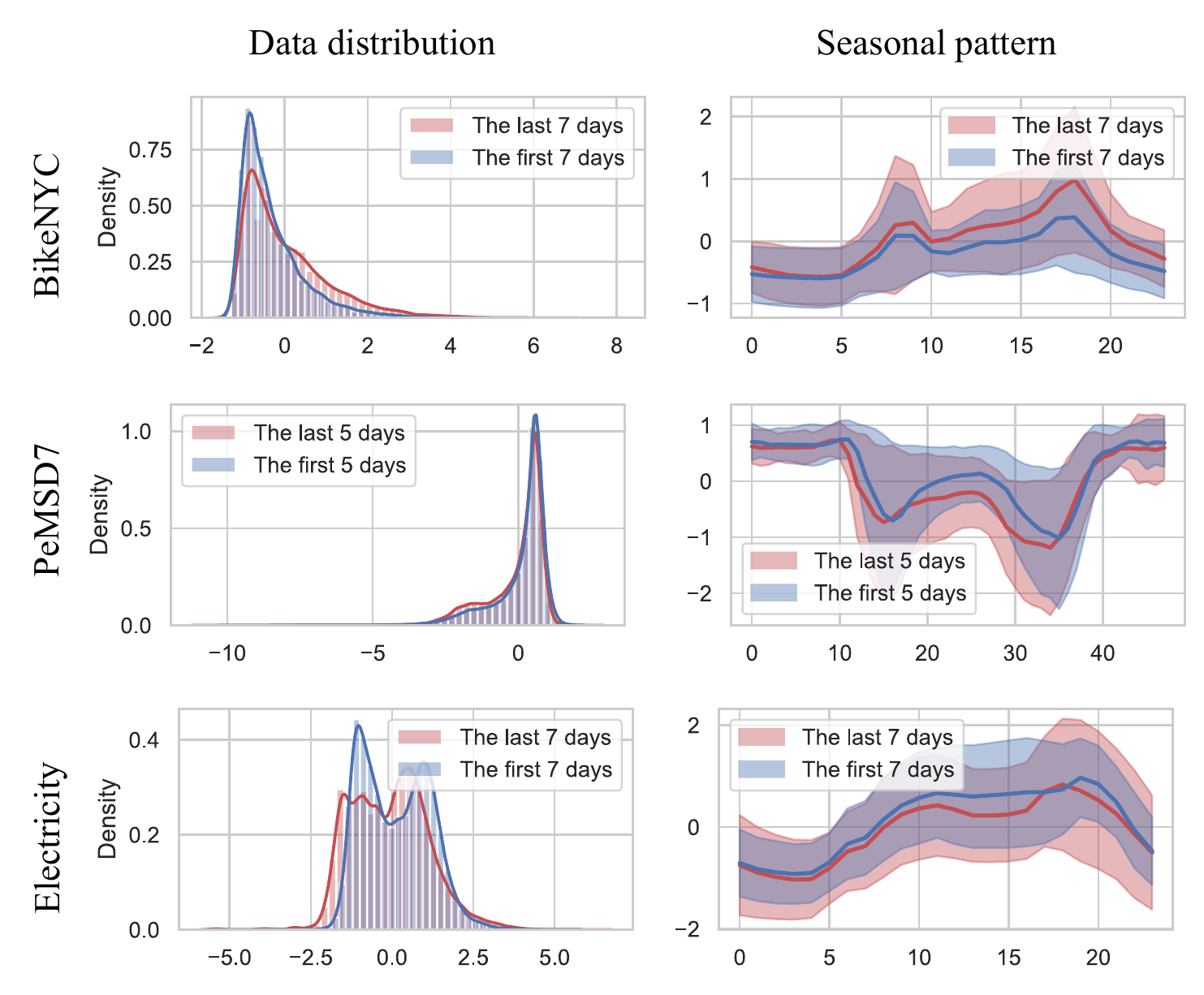}
    \caption{Changes in data patterns as time evolves.}
    \label{fig:data_dist}
\end{figure}

\noindent\textbf{Adaptability to Temporal Shift:} The data patterns for the first and last few days covered by the spatial-temporal datasets are compared in Fig. \ref{fig:data_dist}. The solid line denotes the seasonal mean of MTS; the bind denotes the evolution of the interval between (mean - std, mean + std). It is worth noting that the data patterns for the three datasets, especially the Electricity dataset, show systematic changes from the beginning to the end. As SCNN captures the data patterns on the fly, it can automatically adapt to these statistical changes, which explains that the performance of SCNN, especially when evaluated on Electricity, exceeds that of the other competing methods by a wide margin.

\subsubsection{Long-term Time Series Forecasting}
For LTSF tasks, as reported in Table \ref{tab:LTSF}, SCNN also behaves competitively, compared with recent advancements. Overall, SCNN excels on 31 out of 56 metrics in total; by contrast, PatchTST, the most competent baseline, showcases the best efficacy on only 15 metrics. As a matter of fact, SCNN is capable of achieving the SOTA results on almost all the metrics. The only exceptional cases occurs on ELC and Traffic datasets, when tasked with prediction for the multiple days to come. This sub-optimal performance is attributed to the limitation of SCNN in capturing fine-grained long-range dependencies which are pivotal for these tasks, given that the plain moving average employed by component decoupling, e.g., in Eq. \ref{eq:lt} and Eq. \ref{eq:se}, treats the involved samples as equally important regardless of their temporal positions. In spite of this oversimplification, SCNN showcases remarkable competitiveness in the race with baseline models with complicated designs, e.g, Transformers and MLPs, suggesting the enormous potential of decoupling the heterogeneous structured components in enhancing the forecasts. We leave the optimization of modeling fine-grained long-range dependencies into future exploration.

\begin{table}[tb]
\small
\centering
\caption{Ablation Study}
\label{tab:ablation}
\begin{tabular}{l|c|c|c}
\hline
Models          & BikeNYC & PeMSD7 & Electricity \\ \hline
w/o $\mu^{\text{lt}}$ and  $\sigma^{\text{lt}}$  & 5.12    & \underline{5.04}    & 32.7     \\
w/o $\mu^{\text{se}}$ and  $\sigma^{\text{se}}$        & 5.35    &  5.37    &  37.8     \\
w/o $\mu^{\text{st}}$ and  $\sigma^{\text{st}}$         & 5.11    &  5.08    &  33.7     \\
w/o $\mu^{\text{ce}}$ and  $\sigma^{\text{ce}}$ &  5.56    &    5.17      &  32.5     \\
w/o scaling & \underline{4.98}    &  5.05    &  35.6       \\
w/o adaptive fusion & 5.09    &  5.11   &  33.4       \\
w/o non-negligible $\epsilon$ & 5.50    &  5.12    &  \textbf{30.6}       \\
vanilla MSE loss &  5.22    &  
5.10    &  32.1    \\\hline
SCNN           &  \textbf{4.96}    &  \textbf{5.03}    &  \underline{31.0}     \\ \hline
\end{tabular}
\end{table}

\subsection{Ablation Study}

We design several variants, each of which is without a specific ingredient to be validated. We evaluate these variants on all three datasets and report the overall results on RMSE in Table \ref{tab:ablation}.  It is evident that each component can contribute to the performance of the model, but to different degrees across the three datasets. The co-evolving component is ranked as the most advantageous component in the BikeNYC task. This is because the co-evolving component incorporates the spectrum of effects ranging from long-term to short-term, and can be estimated with reasonable accuracy when the number of co-evolving variables is adequately large, which is the case for the BikeNYC data. The modeling of the long-term component only brings incremental gain to the PeMSD7 task since the training data and the testing data share an identical distribution. The scaling transformation results in significant improvement in the Electricity dataset, owing to its unification of the variables showing great differences in variance. The non-negligible $\epsilon$, as introduced in the last paragraph of Sec. \ref{sec:lt}, is particularly useful for training SCNN on the BikeNYC dataset, as a part of TS in this dataset is very scarce, having only a handful of irregular non-zero measurements. In contrast to the vanilla MSE loss, the structural regularization can shape the structure of the feature space, preventing the overfitting issue and unlocking more power from the structured components.
\begin{figure}[thb]
\centering
\subfloat[]{
\includegraphics[width=0.45\linewidth]{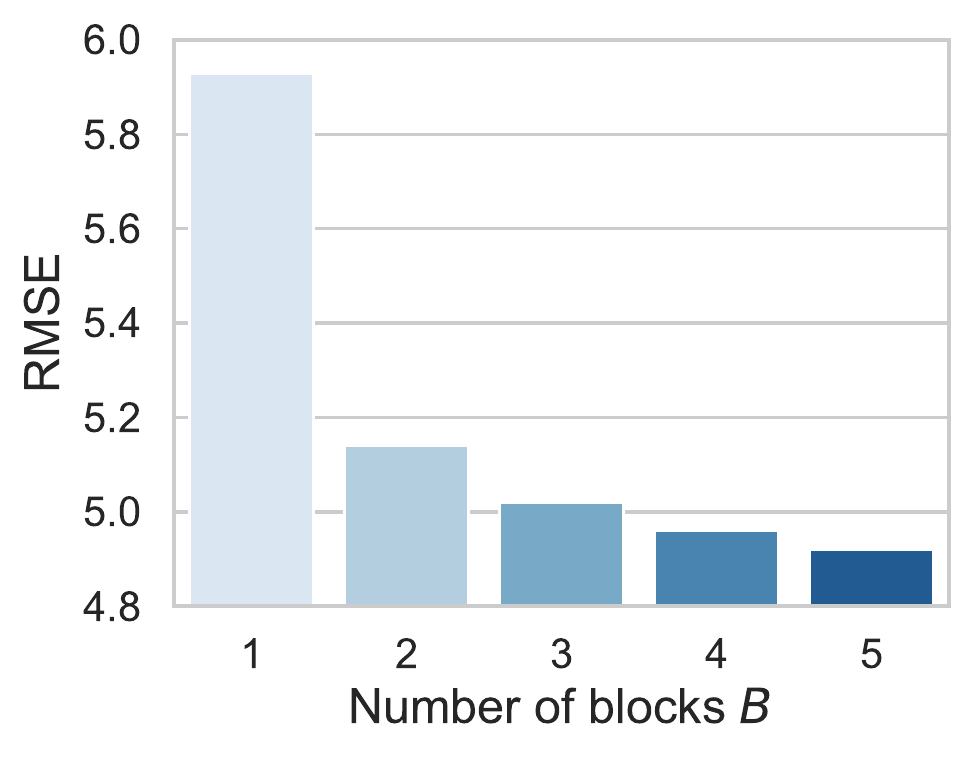}
\label{fig:nlayer}
}
\subfloat[]{
\includegraphics[width=0.45\linewidth]{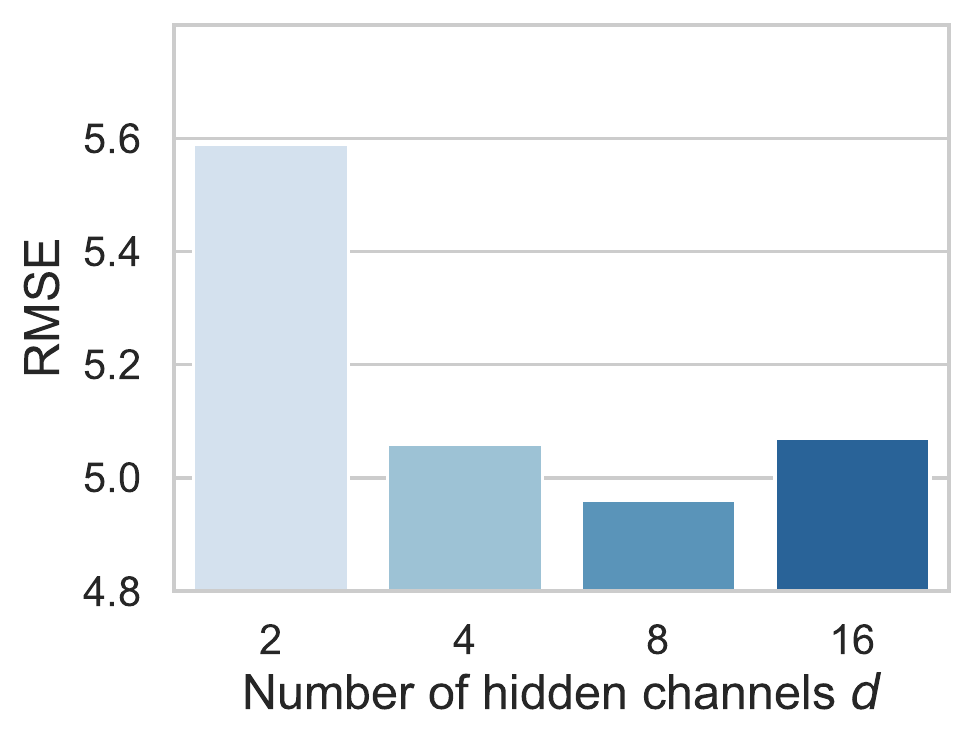}
\label{fig:nchannel}
}

\subfloat[]{
\includegraphics[width=0.45\linewidth]{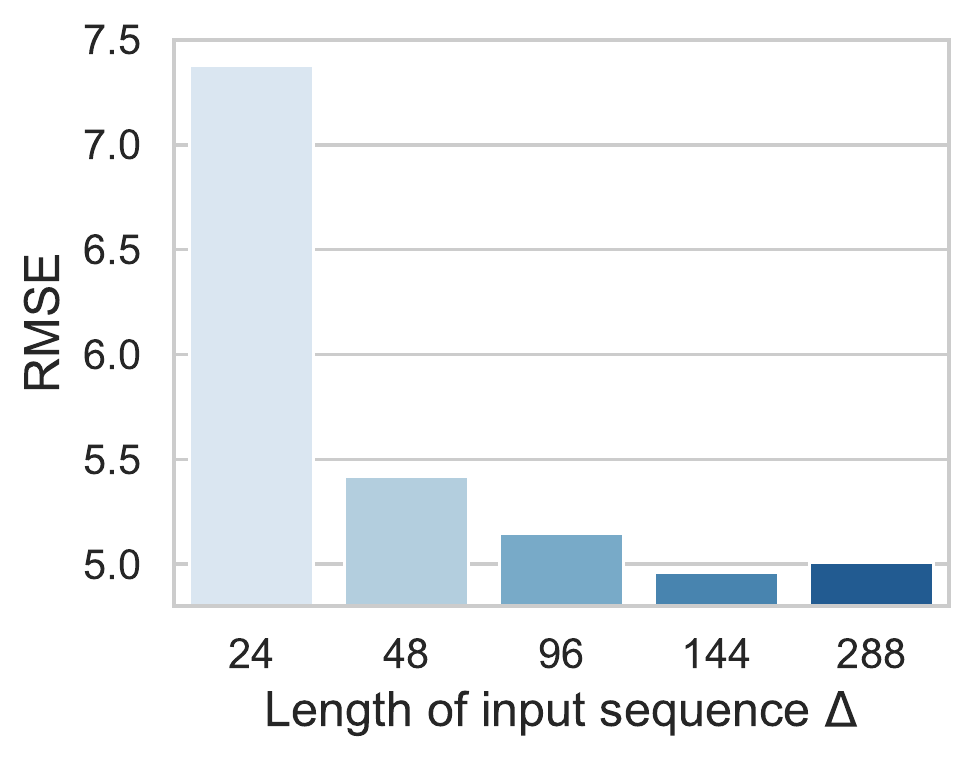}
\label{fig:ninput}
}
\subfloat[]{
\includegraphics[width=0.45\linewidth]{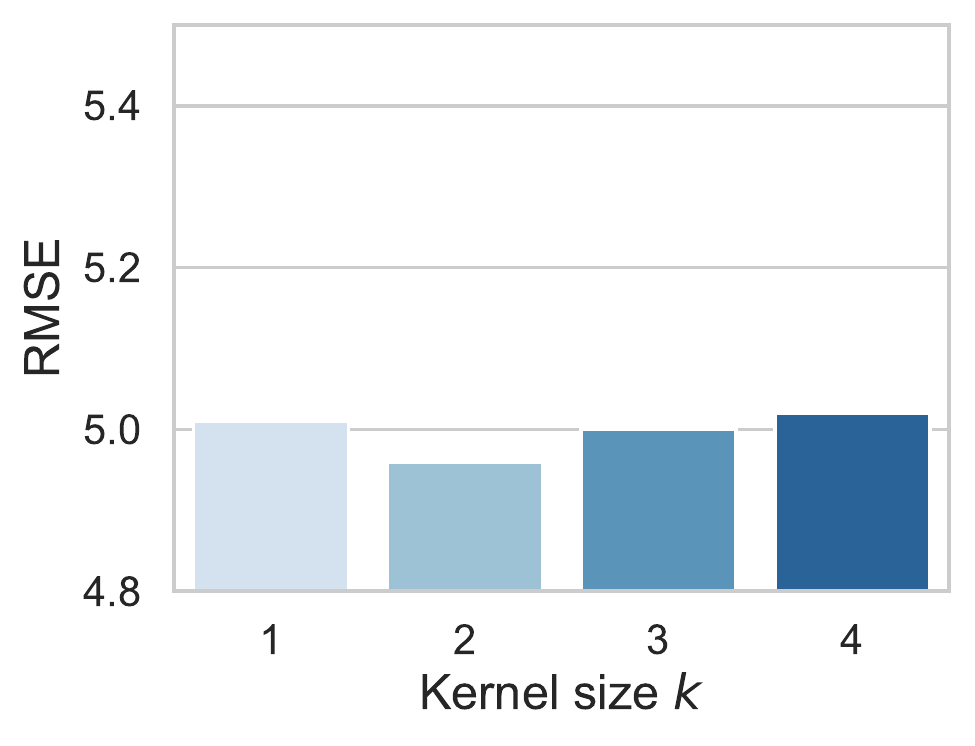}
\label{fig:kernel}
}

\subfloat[]{
\includegraphics[width=0.45\linewidth]{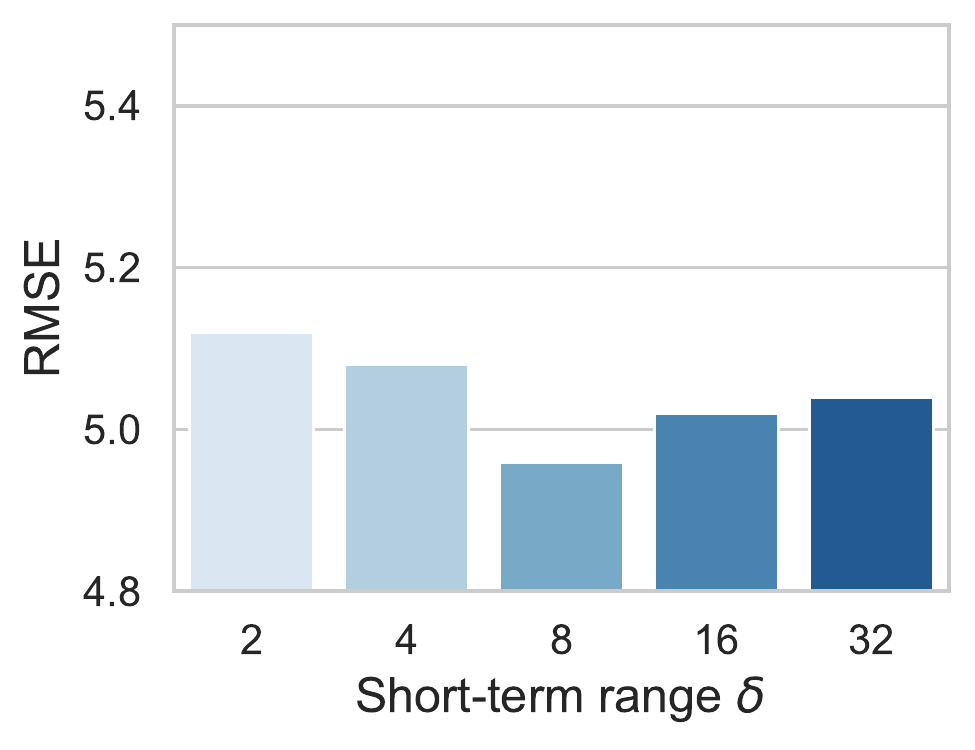}
\label{fig:nshort}
}
\subfloat[]{
\includegraphics[width=0.45\linewidth]{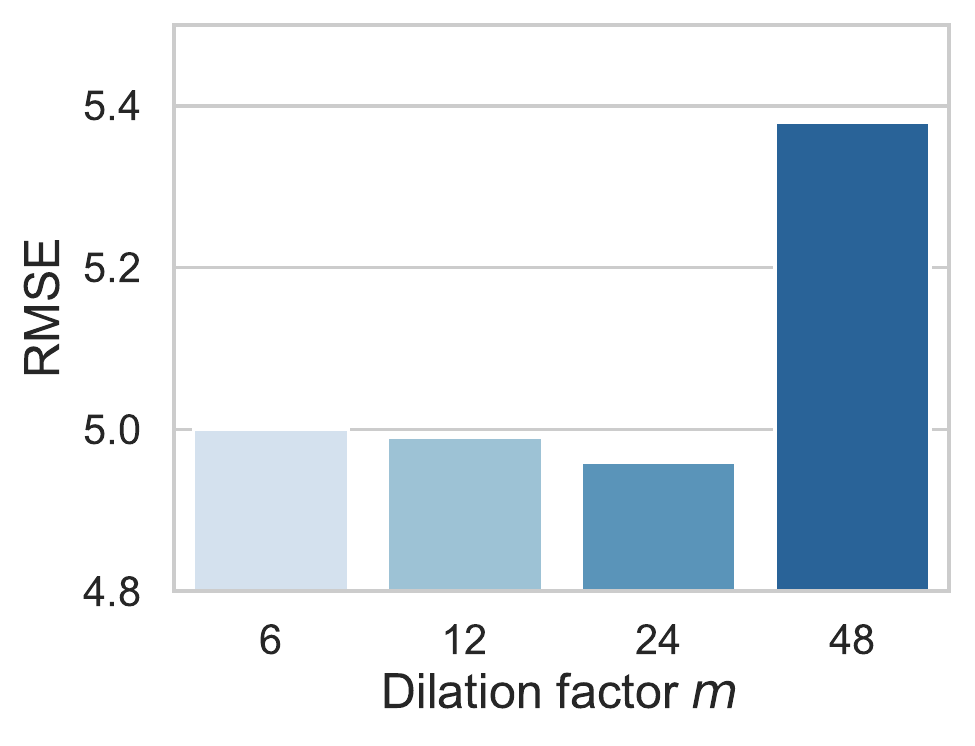}
\label{fig:nseason}
}
\caption{Hyper-parameter analysis on BikeNYC data.}
\label{fig:hyper}
\end{figure}

\begin{figure}
    \centering
    \includegraphics[width=\linewidth]{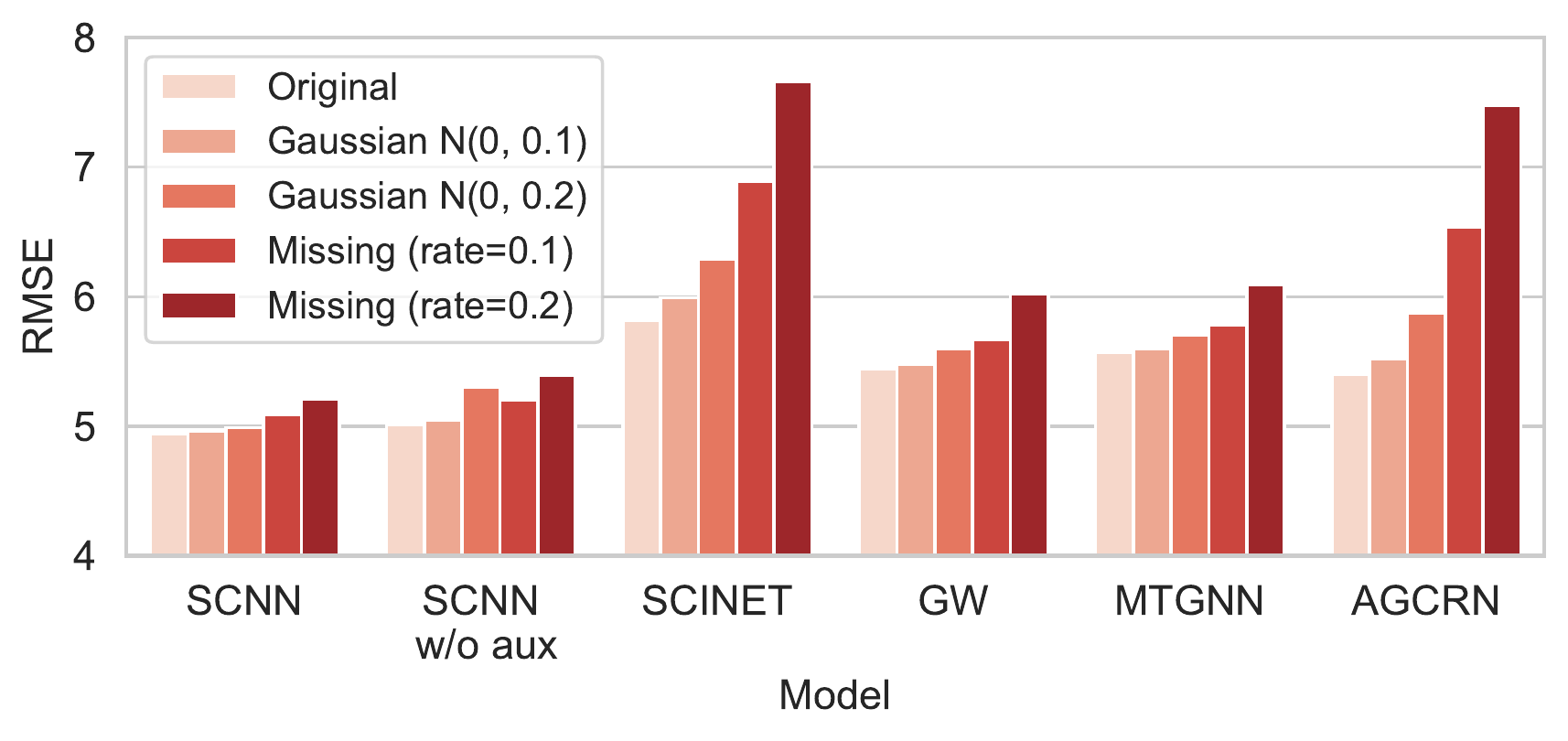}
    \caption{Comparison of robustness.}
    \label{fig:robust}
\end{figure}

\begin{figure}
    \centering
\subfloat[Efficiency]{
\includegraphics[width=0.48\linewidth]{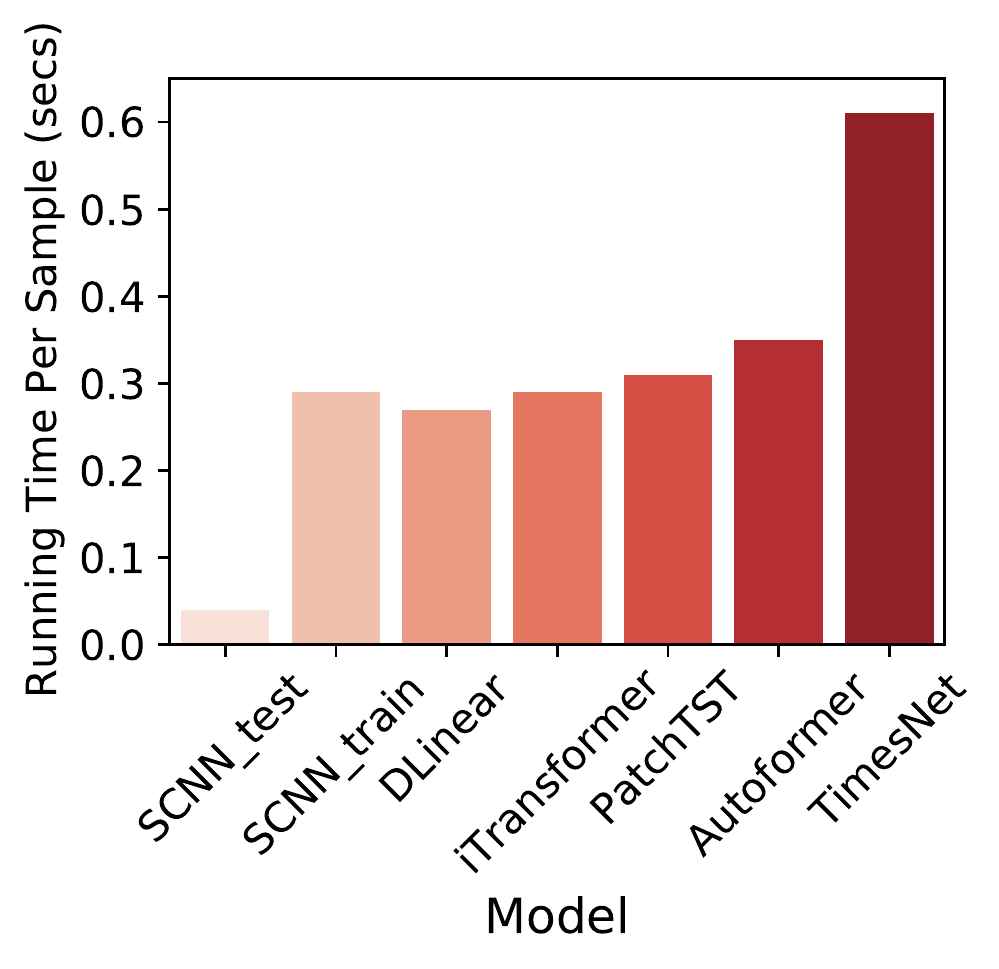}
\label{fig:efficiency}
}
\subfloat[Number of Parameters]{
\includegraphics[width=0.48\linewidth]{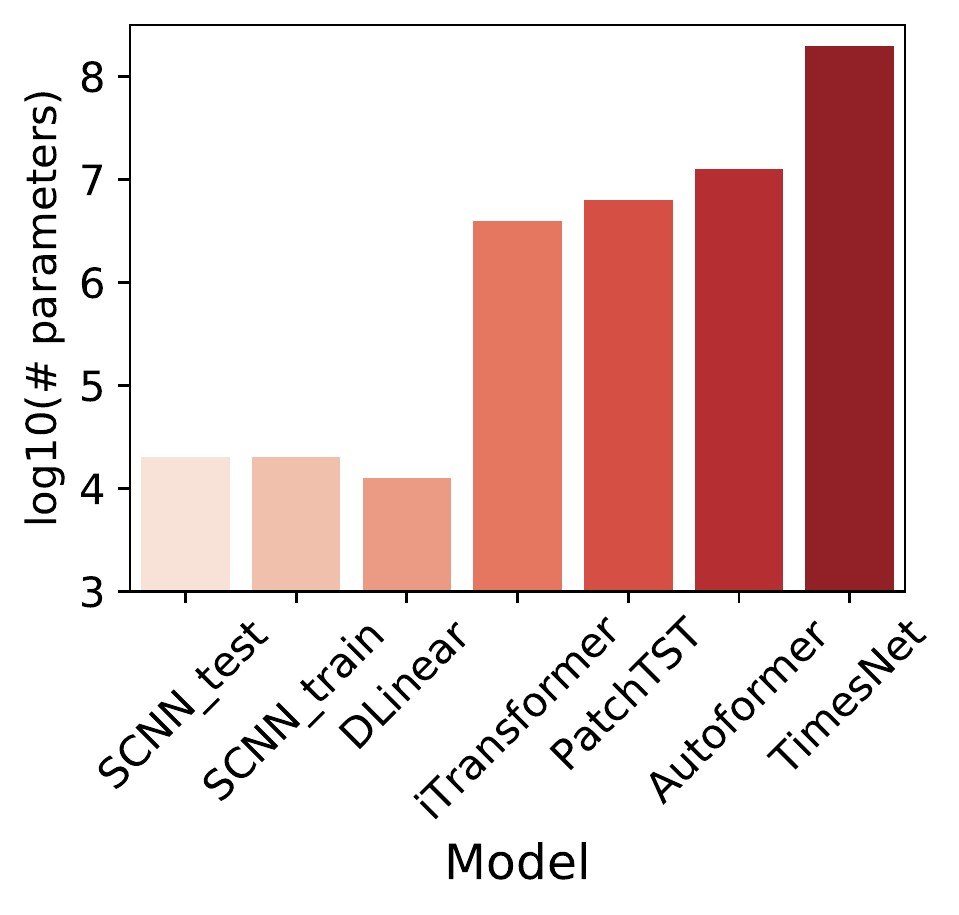}
\label{fig:param}
}
    \caption{Comparison of scalability on ELC dataset.}
    \label{fig:complexity}
\end{figure}

\begin{figure}[tb]
\centering
\subfloat[Short-term Component]{
\includegraphics[width=0.9\linewidth]{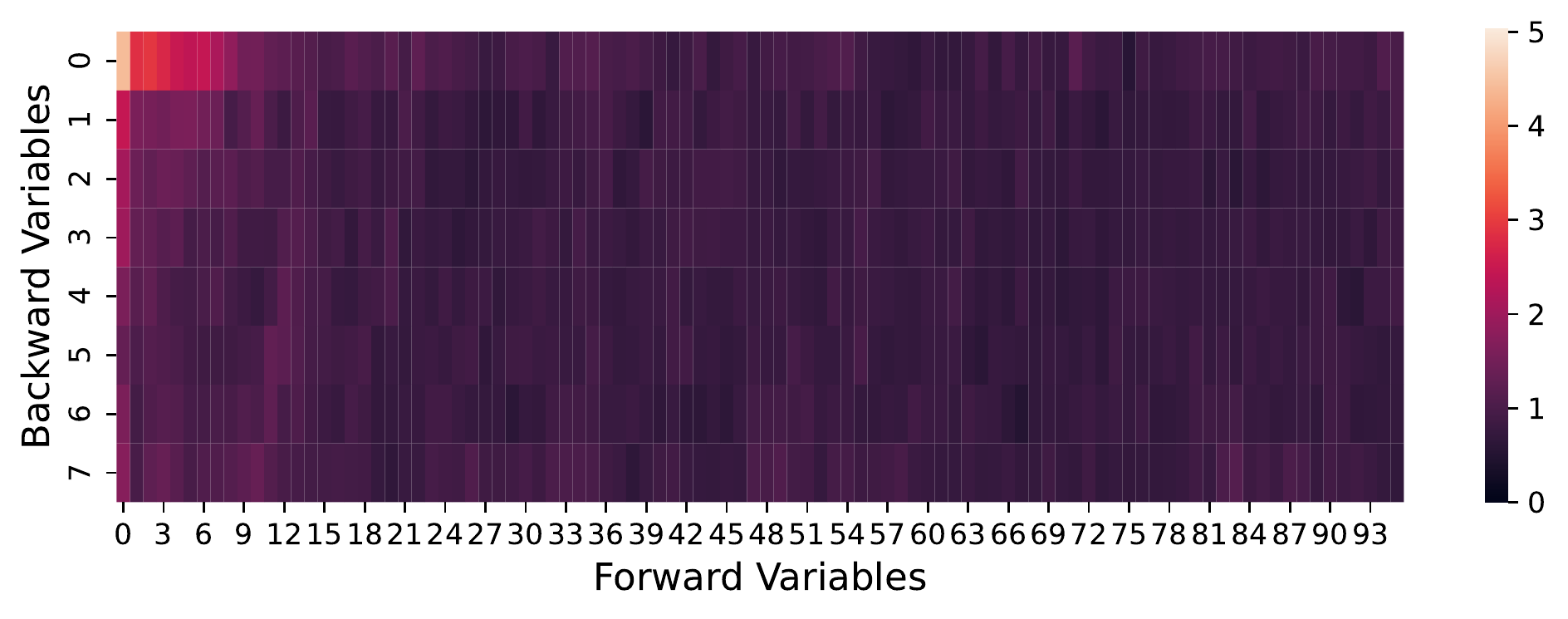}
}

\subfloat[Co-evolving Component]{
\includegraphics[width=0.9\linewidth]{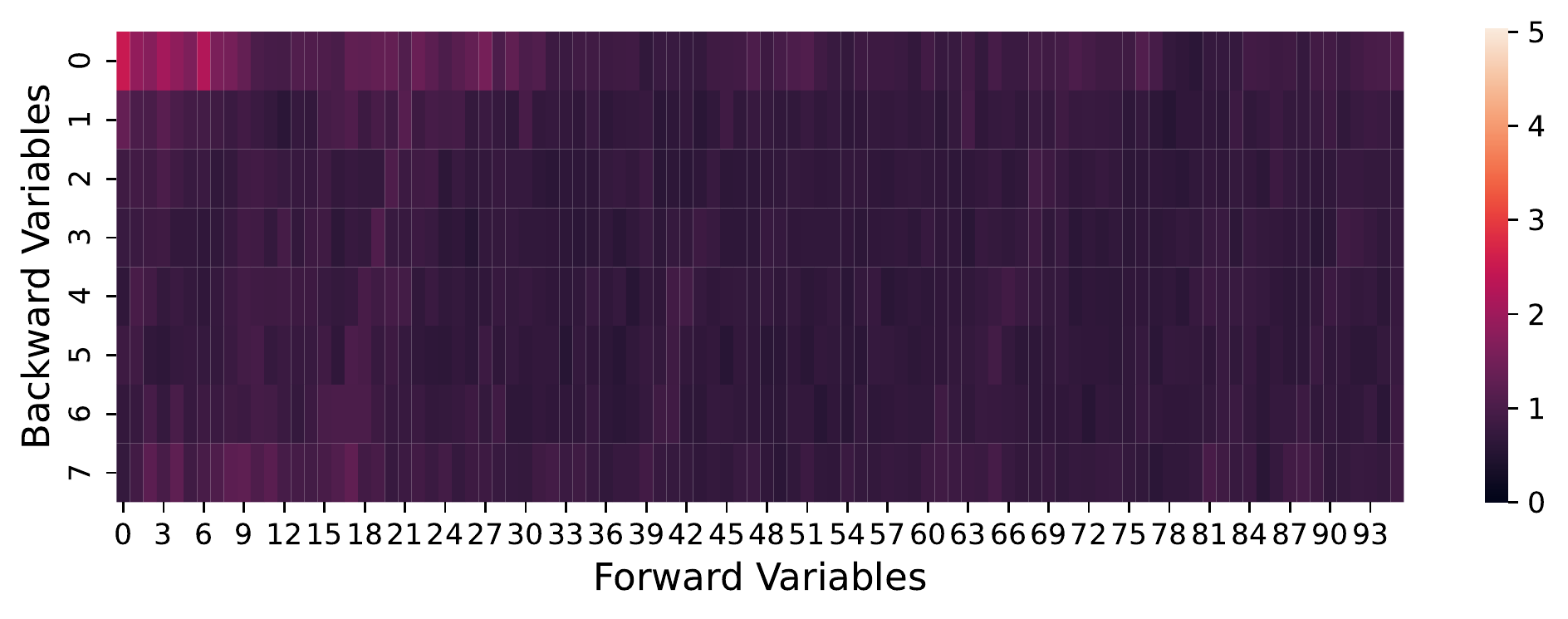}
}

\subfloat[Residual Component]{
\includegraphics[width=0.9\linewidth]{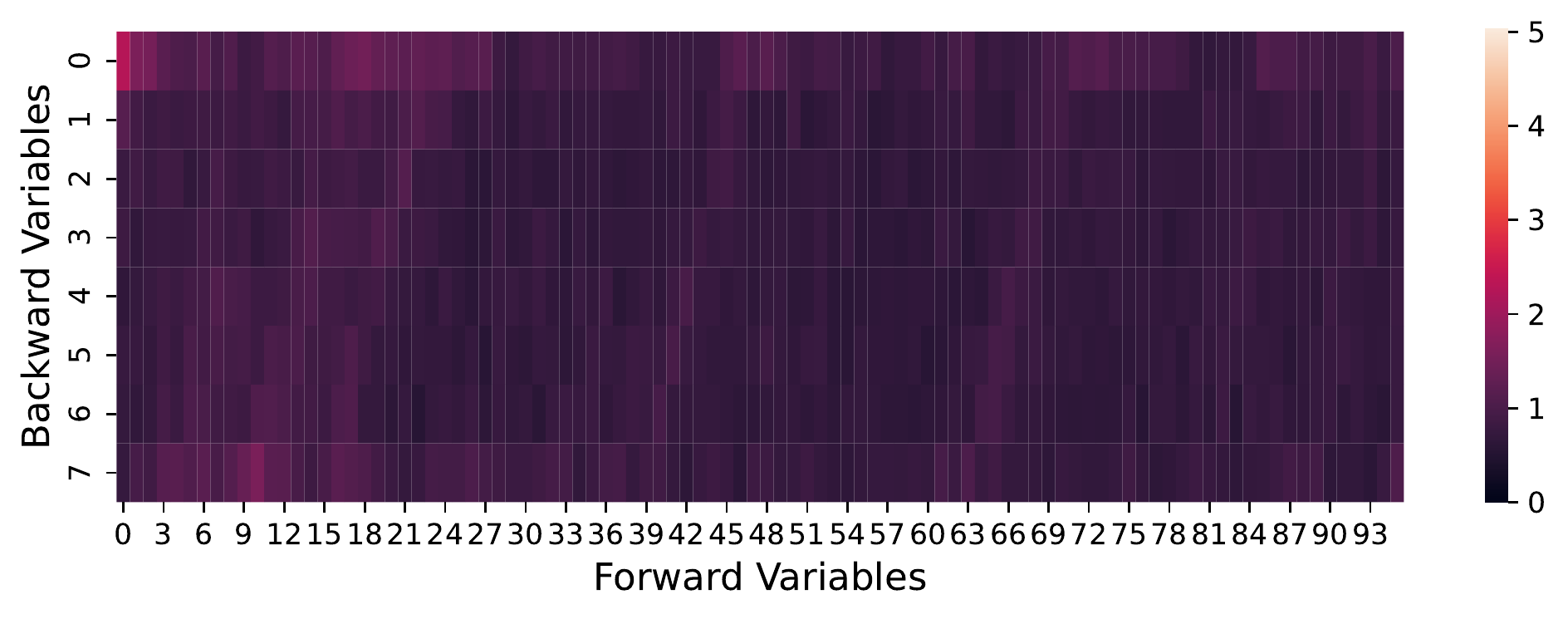}
}
\caption{Evaluation of the interpretability of SCNN on the ELC dataset}
\label{fig:interpretability}
\end{figure}

\begin{figure*}[thb]
\centering
\subfloat[]{
\includegraphics[width=0.45\linewidth]{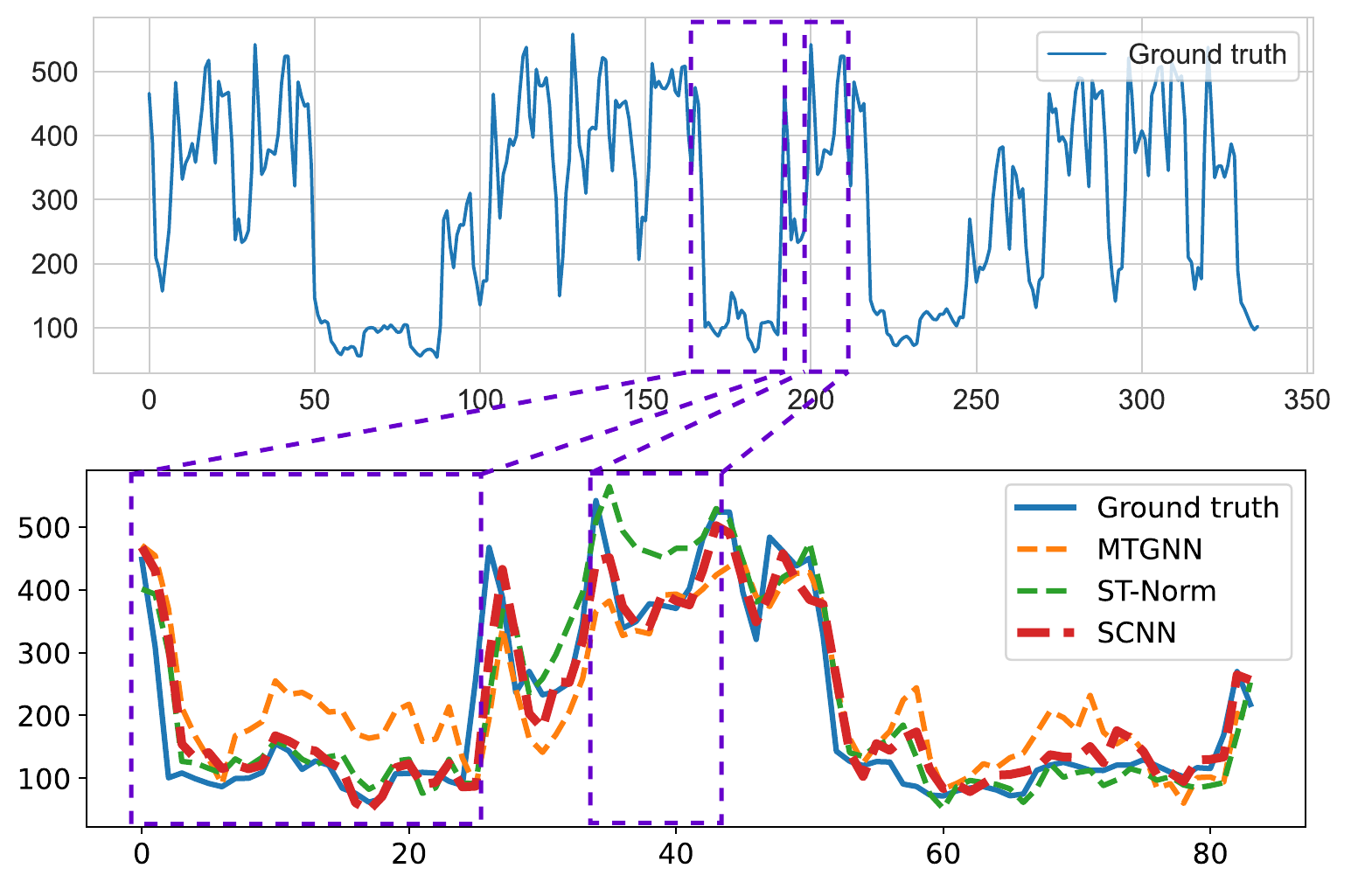}
\label{fig:case1}
}
\subfloat[]{
\includegraphics[width=0.45\linewidth]{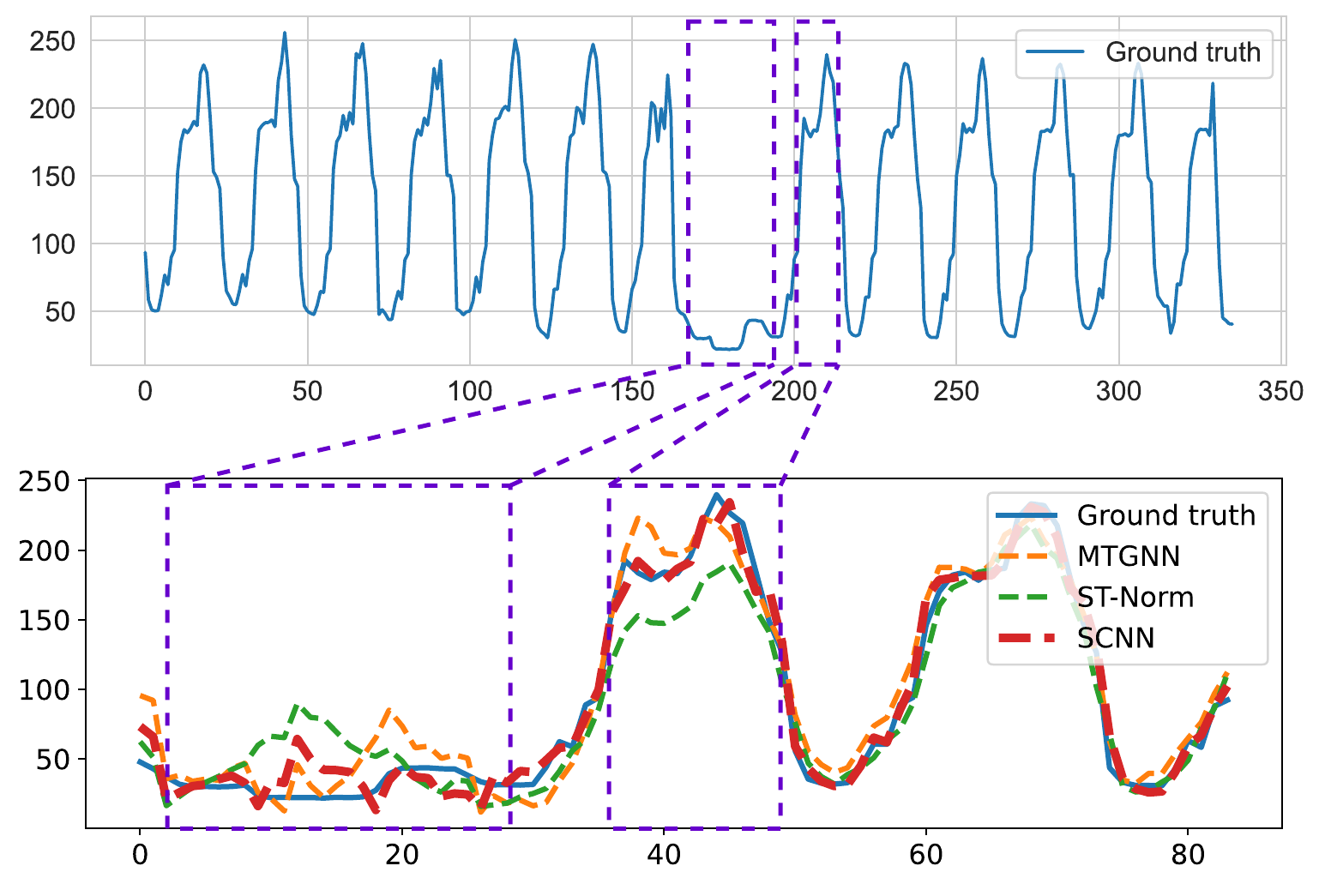}
\label{fig:case2}
}

\caption{Anomalous cases performance evaluation. The results demonstrate that SCNN consistently achieves the lowest prediction error among the three models in diverse and challenging scenarios of distribution shifts and anomalies.}
\label{fig:case}
\end{figure*}

\begin{figure*}[h]
    \centering
    \includegraphics[width=\linewidth]{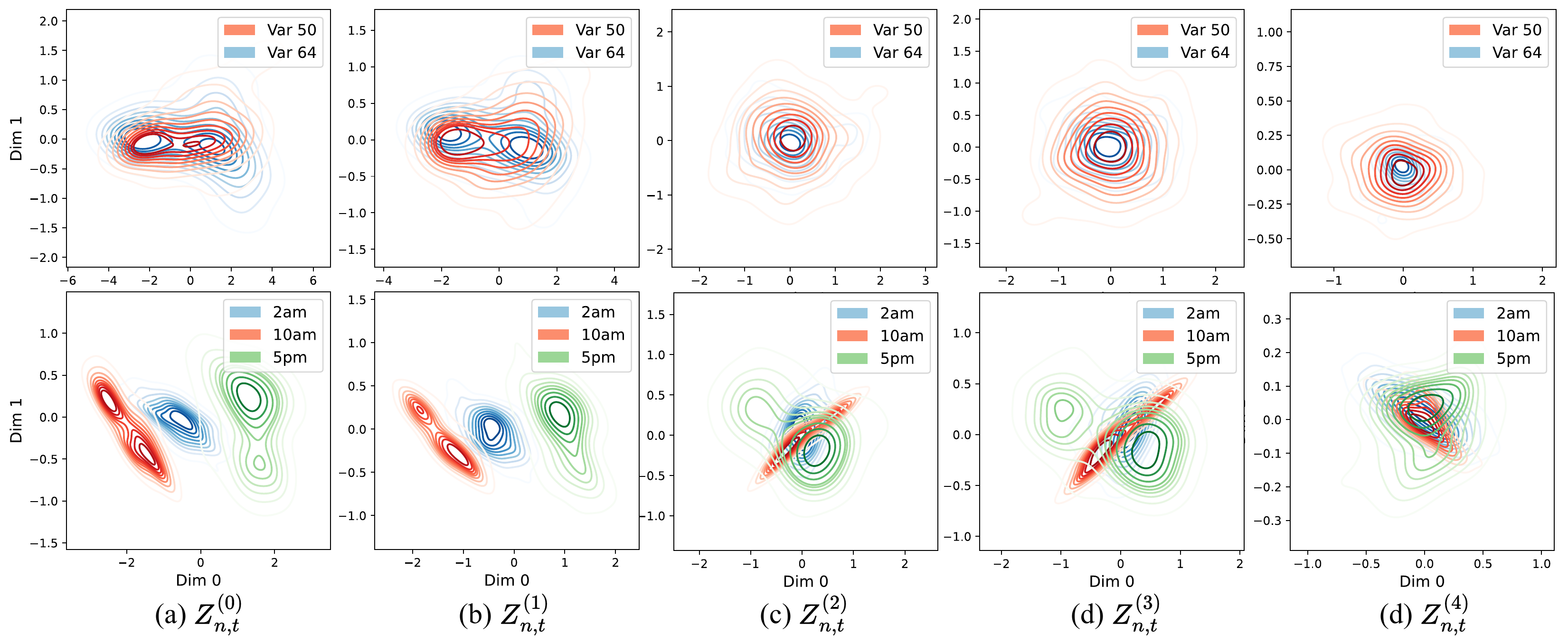}
    \caption{Visualization of residual representations.}
    \label{fig:qualitative}
\end{figure*}

\subsection{Hyper-Parameter Analysis}

As shown in Fig. \ref{fig:nlayer}, it is surprising that a 2-layer SCNN achieves fairly good performance, and more layers only result in incremental improvements. This demonstrates that shallow layers work on coarse-grained prediction, and deep layers perform fine-grained calibration by capturing the detailed changes presented in the MTS data. Fig. \ref{fig:nchannel} shows that the prediction error of SCNN firstly decreases and then increases as the number of hidden channels increases. The number of input steps can affect the estimation of the long-term component and the seasonal component, thereby leading to differences in the accuracy of the forecast, as illustrated in Fig. \ref{fig:ninput}. It is appealing to find from Fig. \ref{fig:kernel} that SCNN behaves competitively with the kernel of size 1, which means that the correlations across the local observations vanish once conditioned on the set of structured components. Fig. \ref{fig:nshort} and Fig. \ref{fig:nseason} demonstrate the effectiveness of the setup of the other two hyper-parameters.

\begin{figure*}[h!]
\centering
\subfloat[$\mu^\text{lt}_{n, t}$]{
\includegraphics[width=0.24\linewidth]{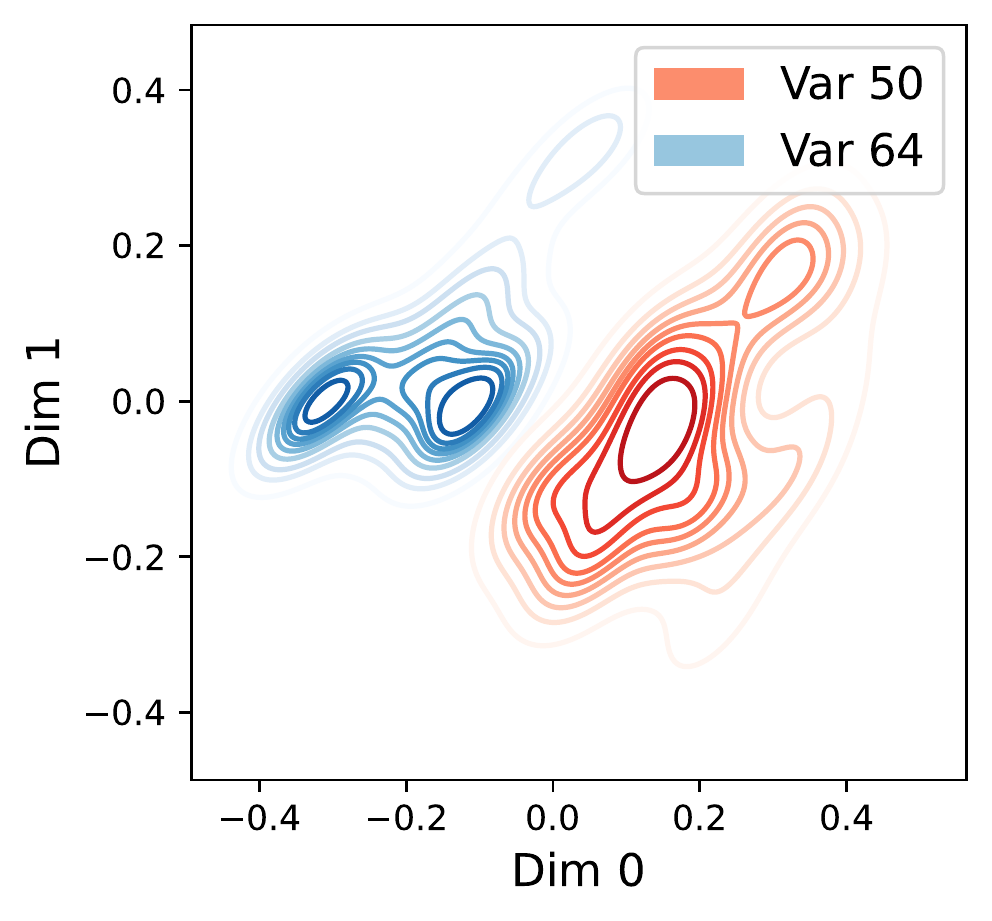}
}
\subfloat[$\mu^\text{se}_{n, t}$]{
\includegraphics[width=0.24\linewidth]{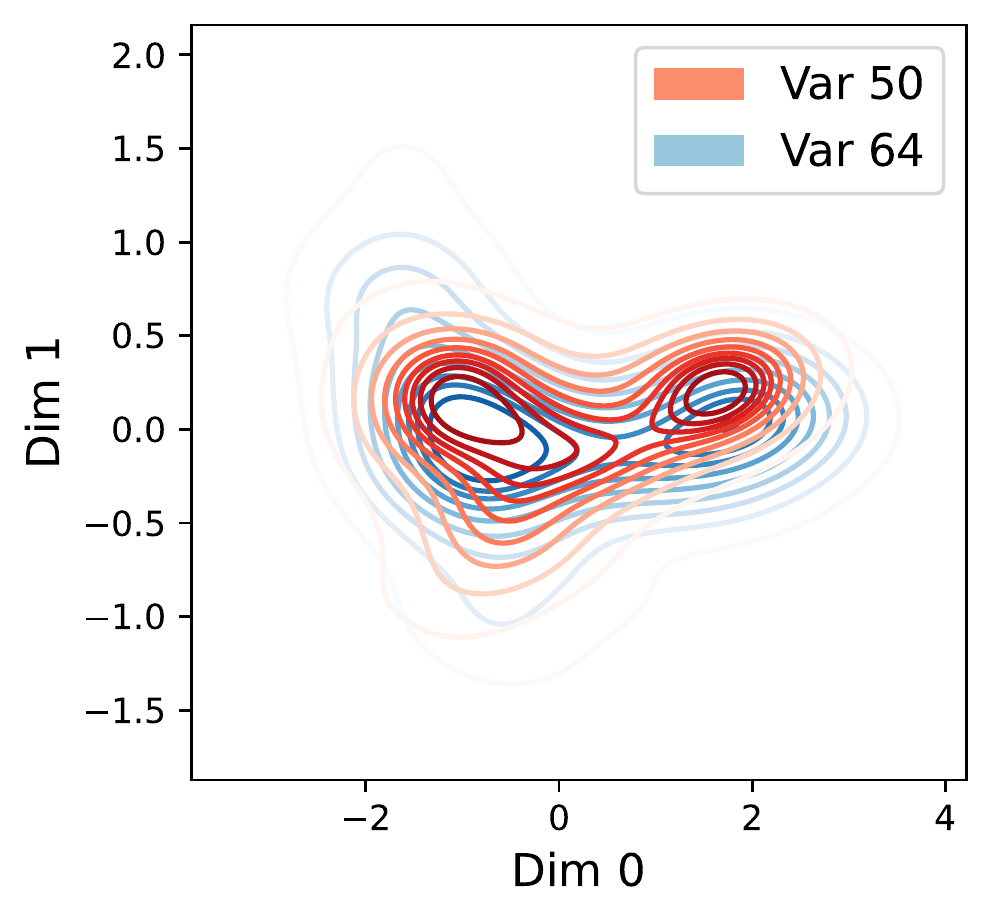}
}
\subfloat[$\mu^\text{st}_{n, t}$]{
\includegraphics[width=0.24\linewidth]{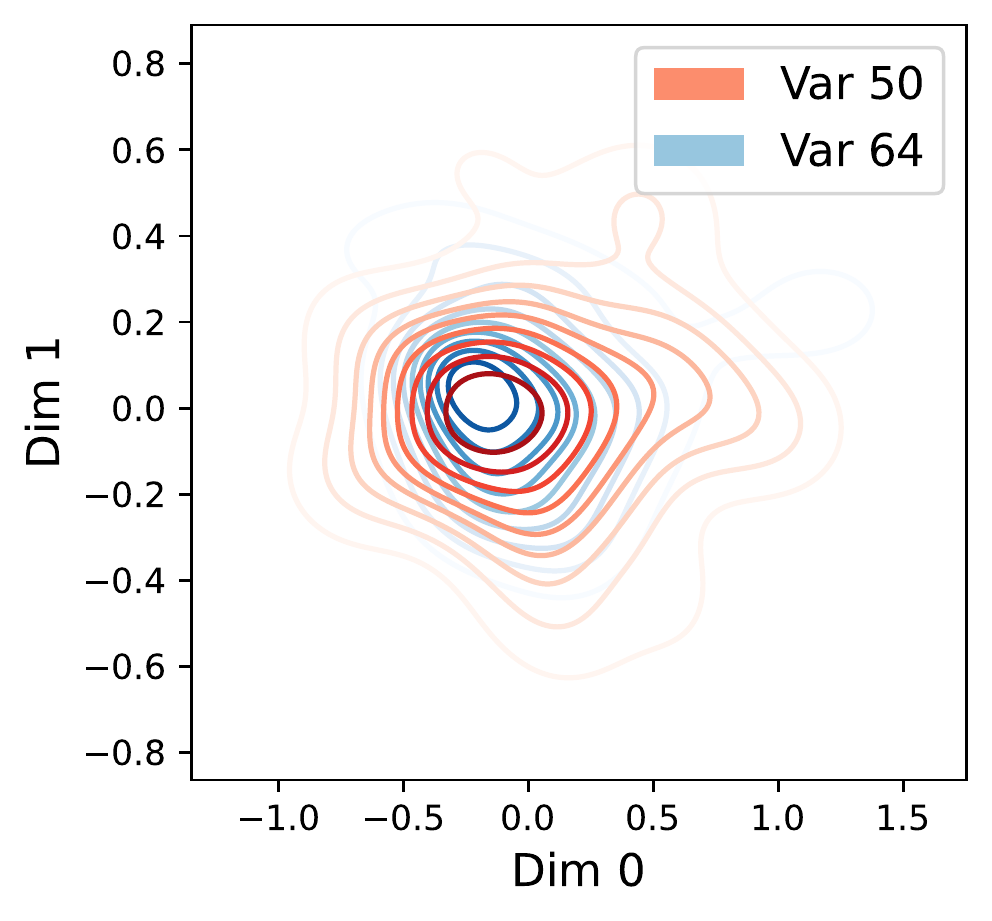}
}
\subfloat[$\mu^\text{ce}_{n, t}$]{
\includegraphics[width=0.24\linewidth]{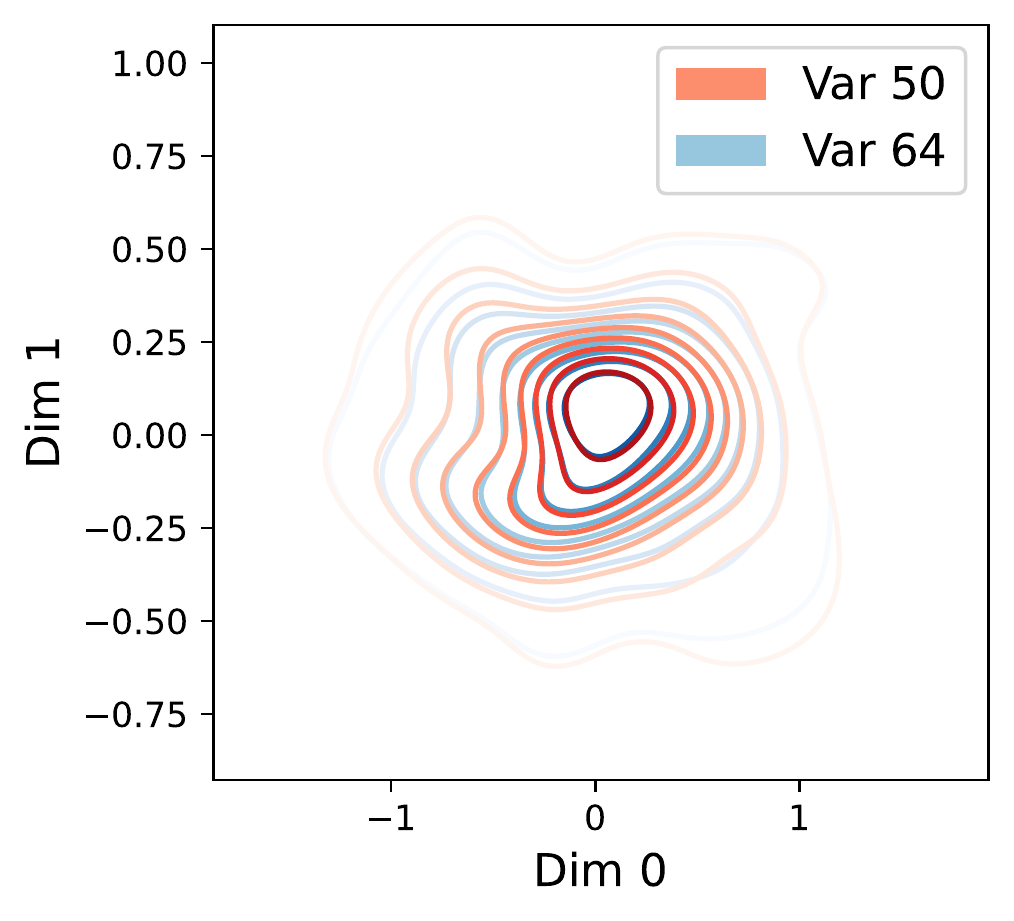}
}

\subfloat[$\sigma^\text{lt}_{n, t}$]{
\includegraphics[width=0.24\linewidth]{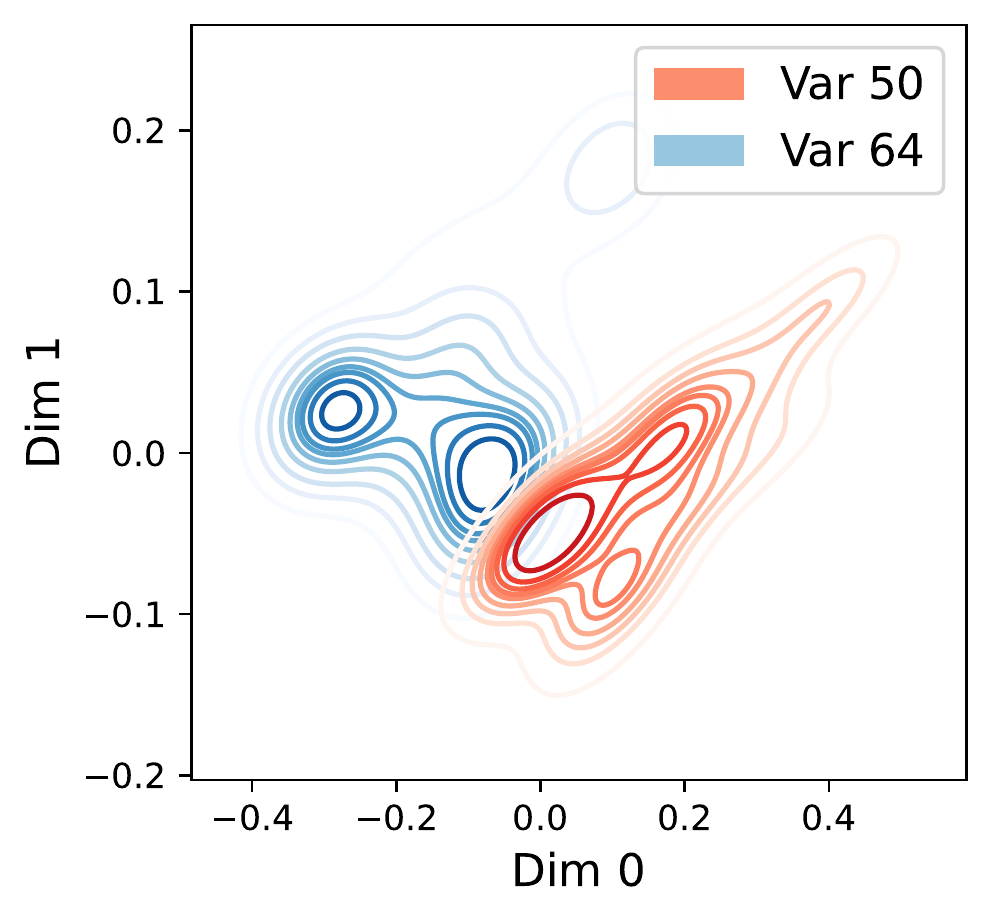}
}
\subfloat[$\sigma^\text{se}_{n, t}$]{
\includegraphics[width=0.24\linewidth]{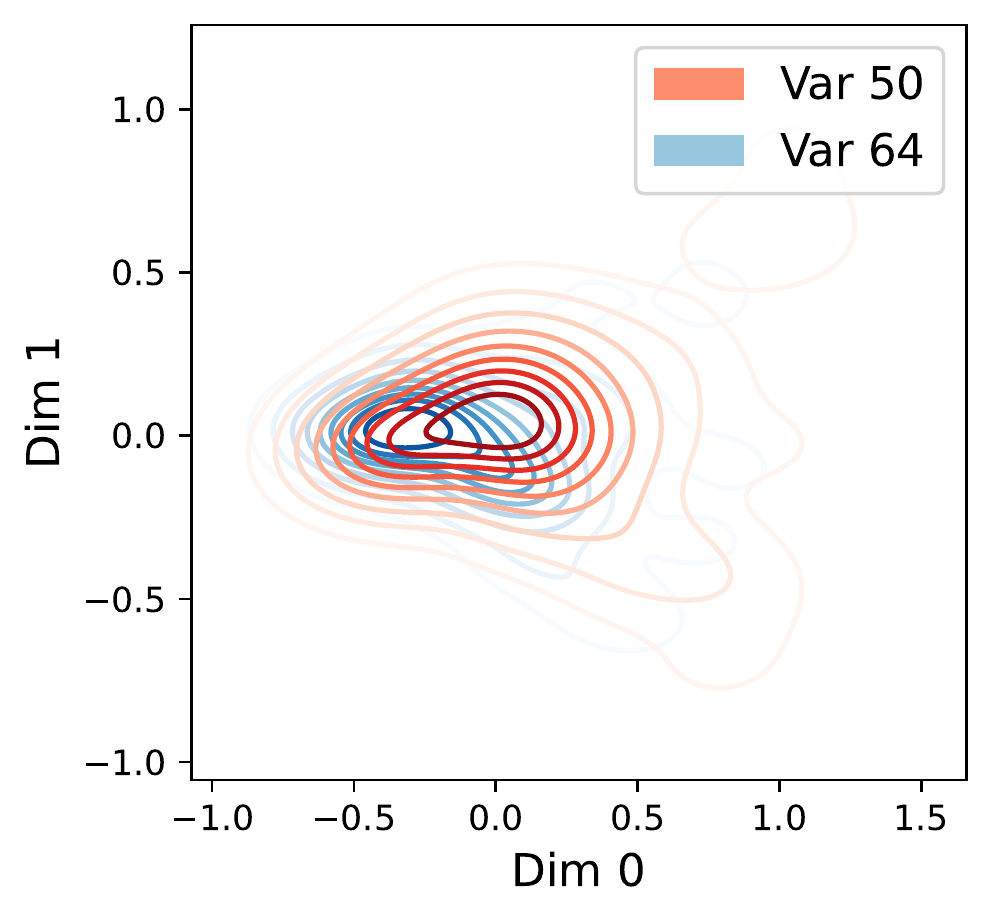}
}
\subfloat[$\sigma^\text{st}_{n, t}$]{
\includegraphics[width=0.24\linewidth]{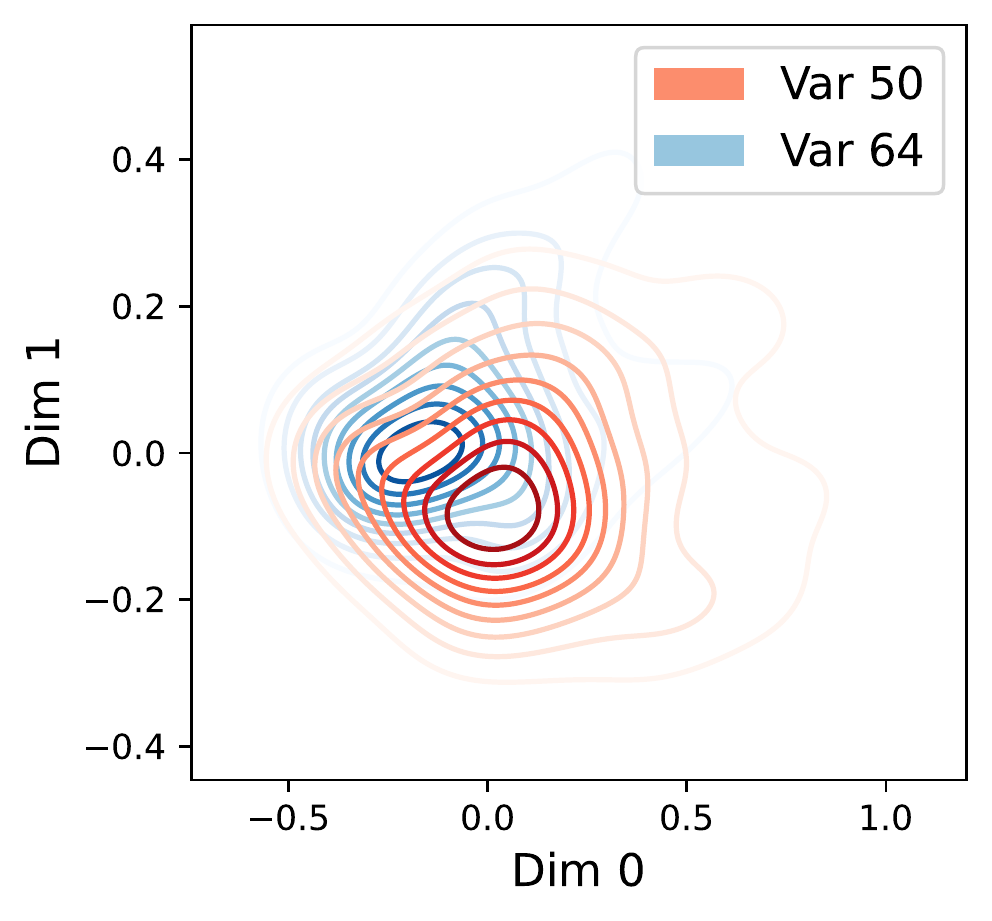}
}
\subfloat[$\sigma^\text{ce}_{n, t}$]{
\includegraphics[width=0.24\linewidth]{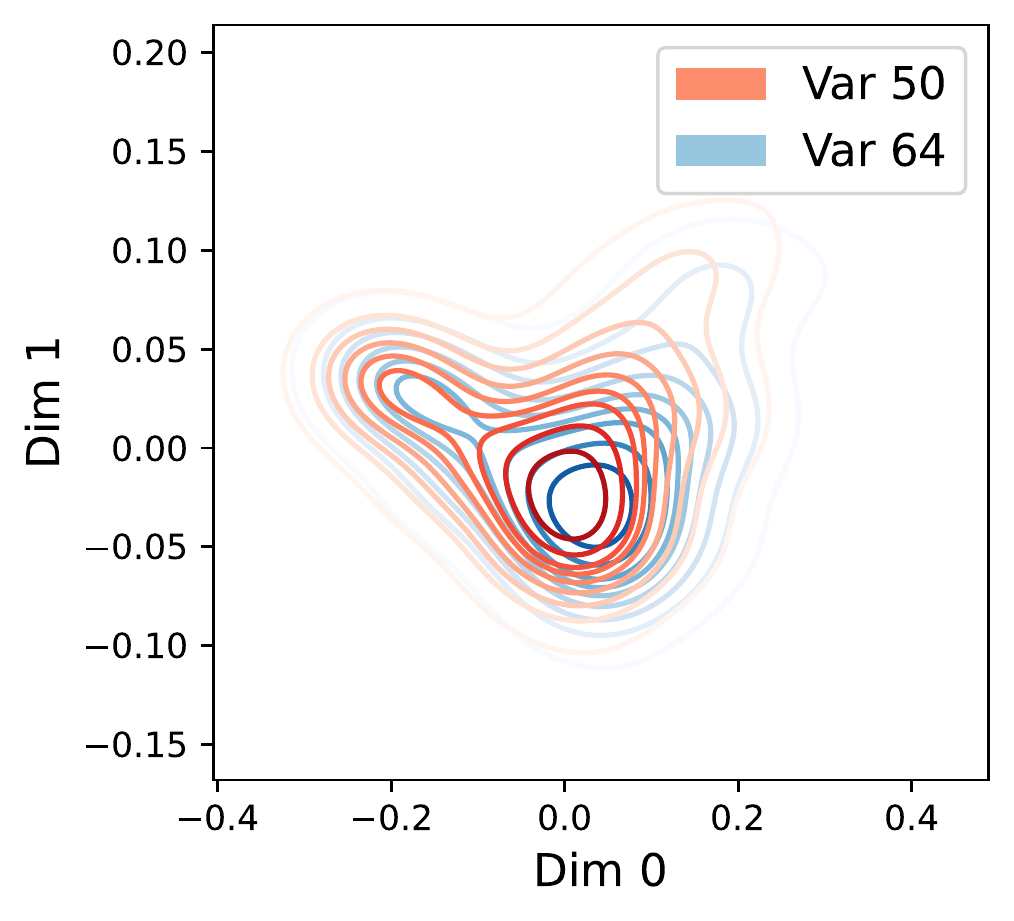}
}

\subfloat[$\mu^\text{lt}_{n, t}$]{
\includegraphics[width=0.24\linewidth]{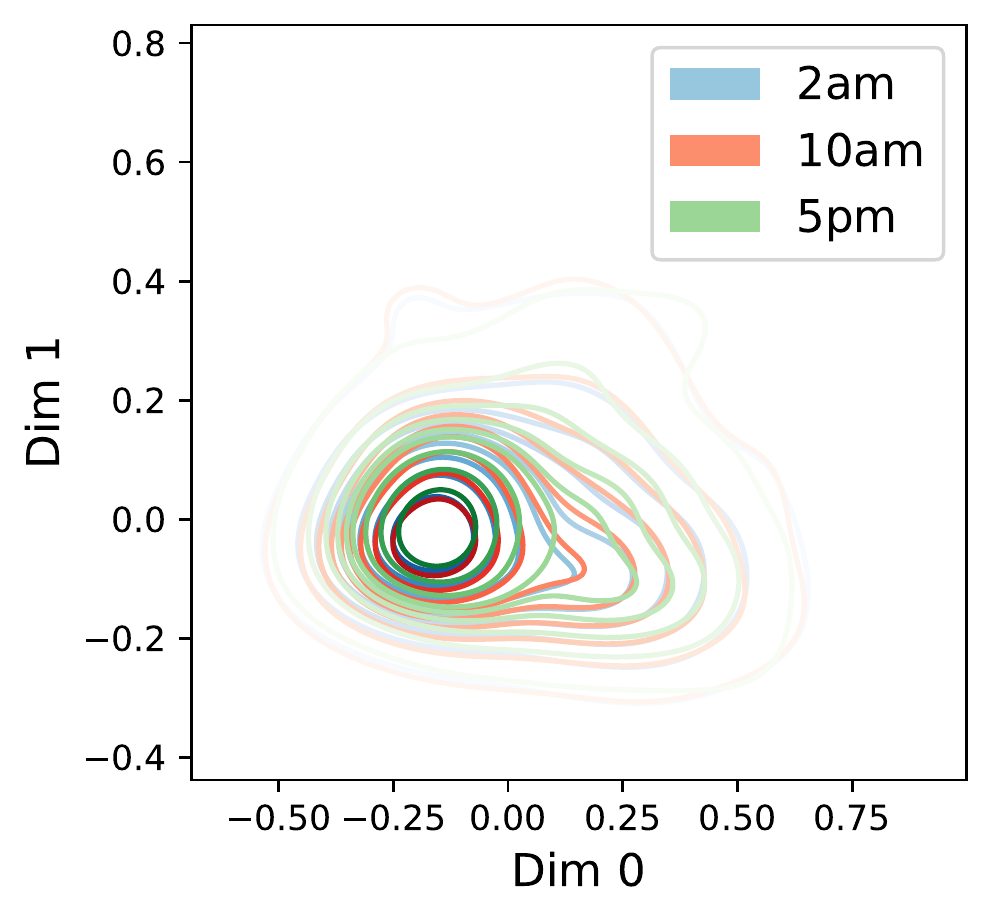}
}
\subfloat[$\mu^\text{se}_{n, t}$]{
\includegraphics[width=0.24\linewidth]{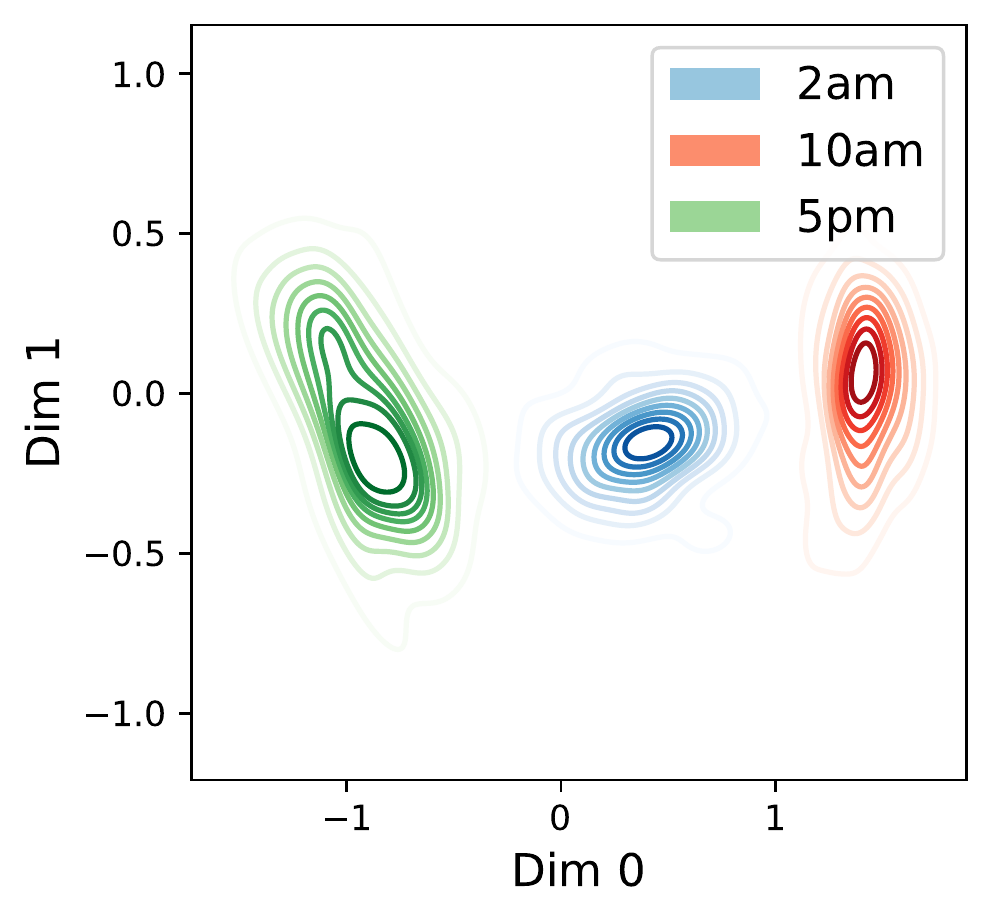}
}
\subfloat[$\mu^\text{st}_{n, t}$]{
\includegraphics[width=0.24\linewidth]{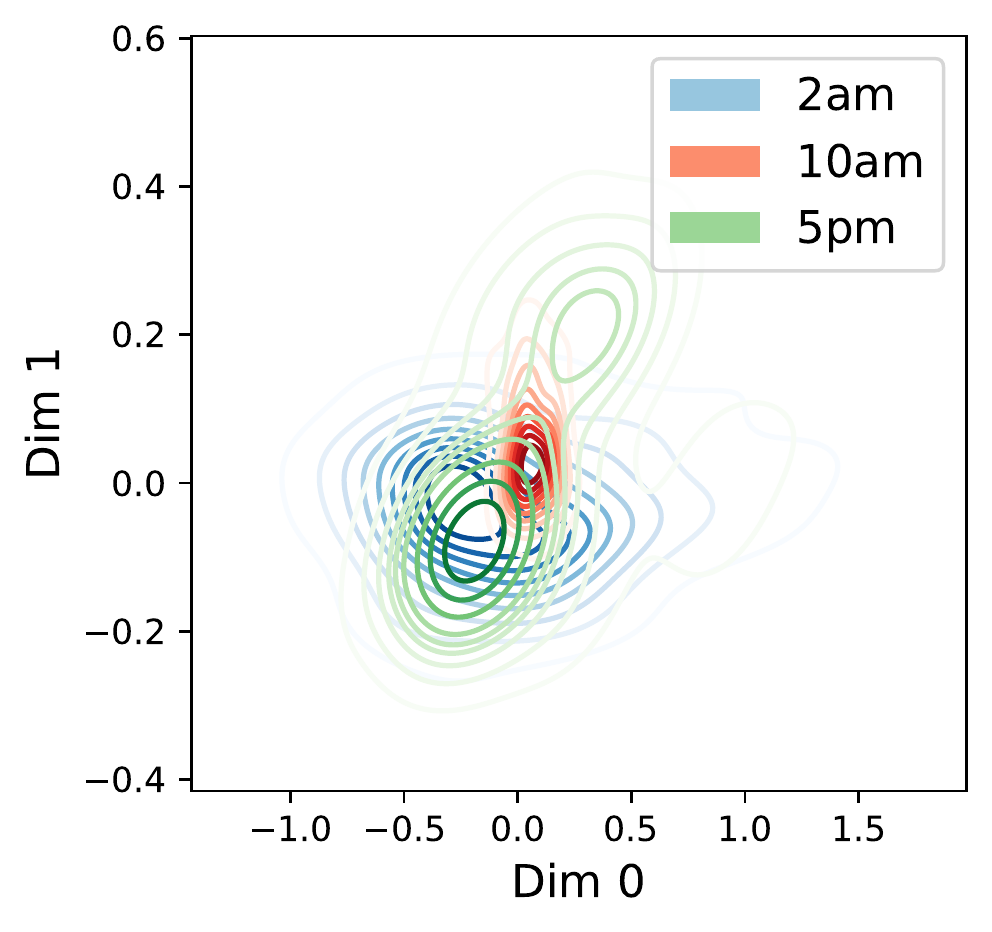}
}
\subfloat[$\mu^\text{ce}_{n, t}$]{
\includegraphics[width=0.24\linewidth]{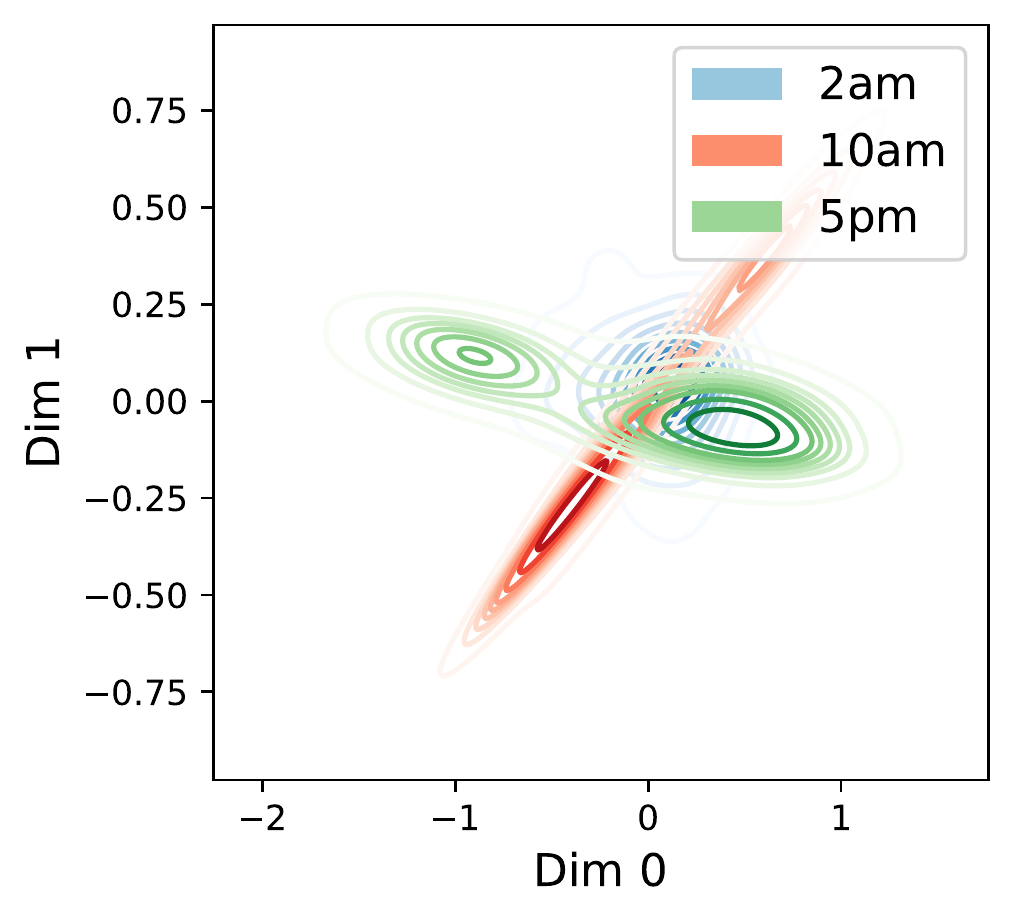}
}

\subfloat[$\sigma^\text{lt}_{n, t}$]{
\includegraphics[width=0.24\linewidth]{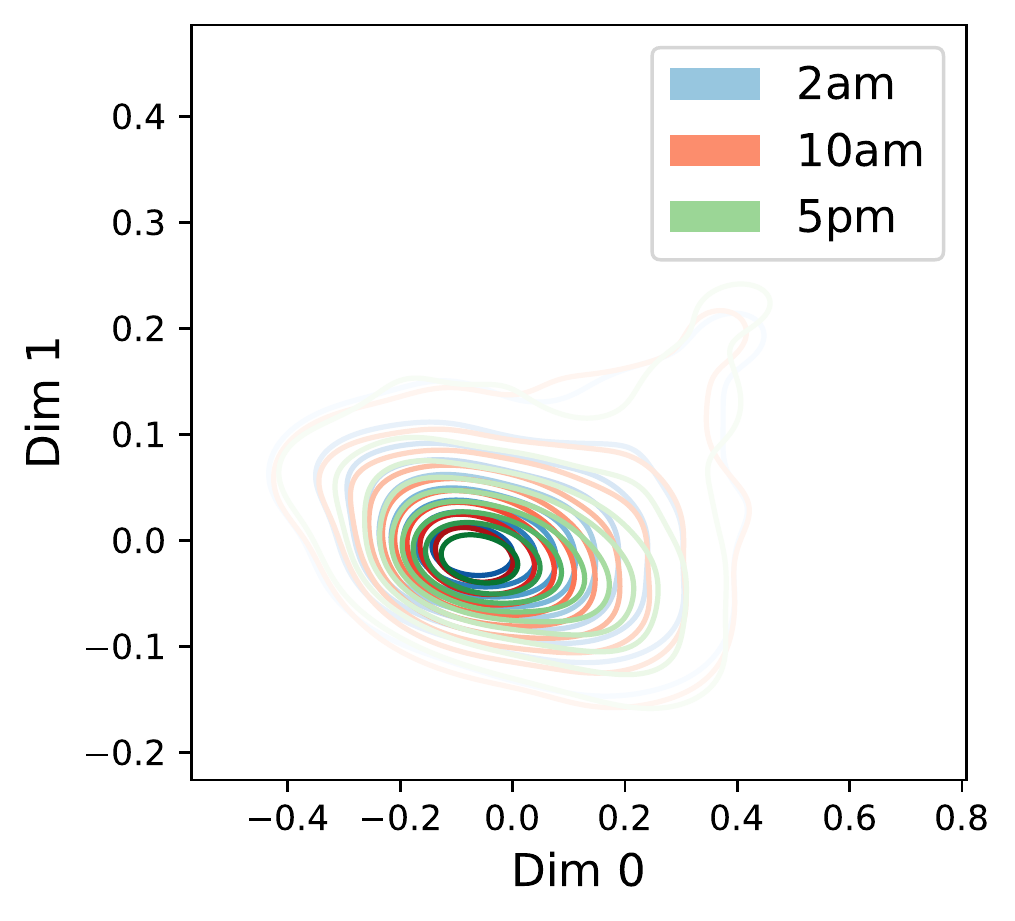}
}
\subfloat[$\sigma^\text{se}_{n, t}$]{
\includegraphics[width=0.24\linewidth]{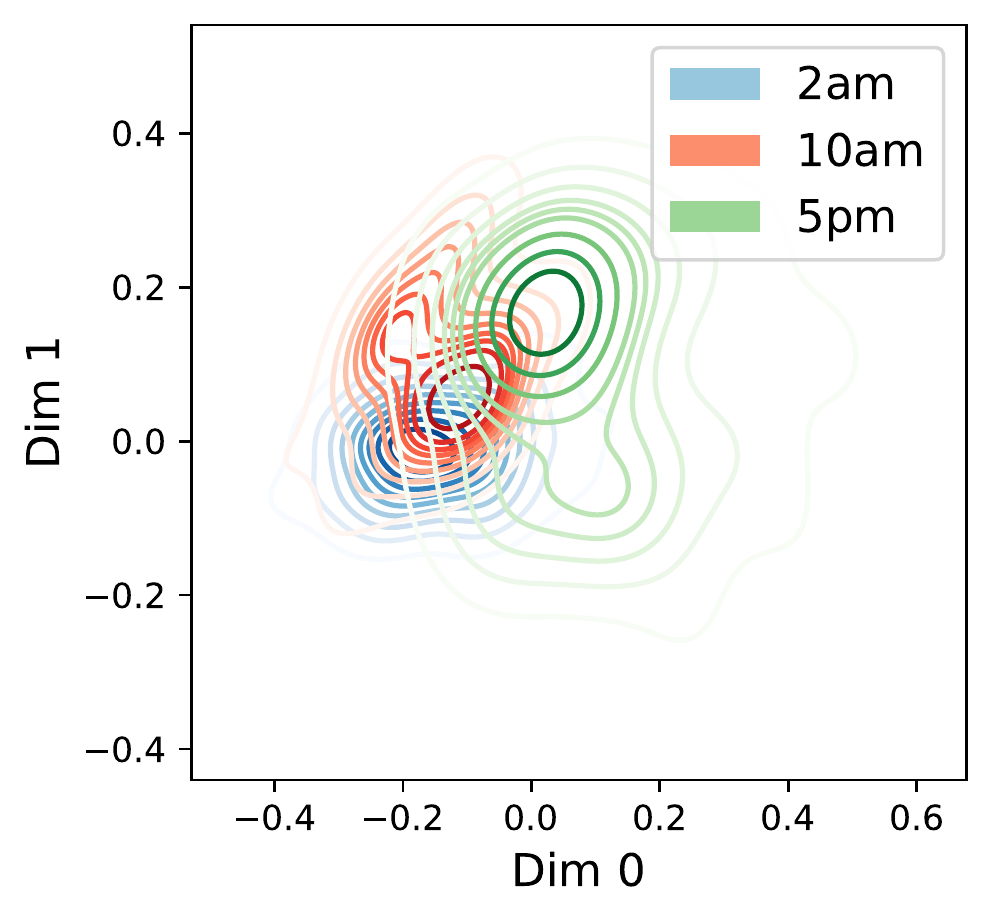}
}
\subfloat[$\sigma^\text{st}_{n, t}$]{
\includegraphics[width=0.24\linewidth]{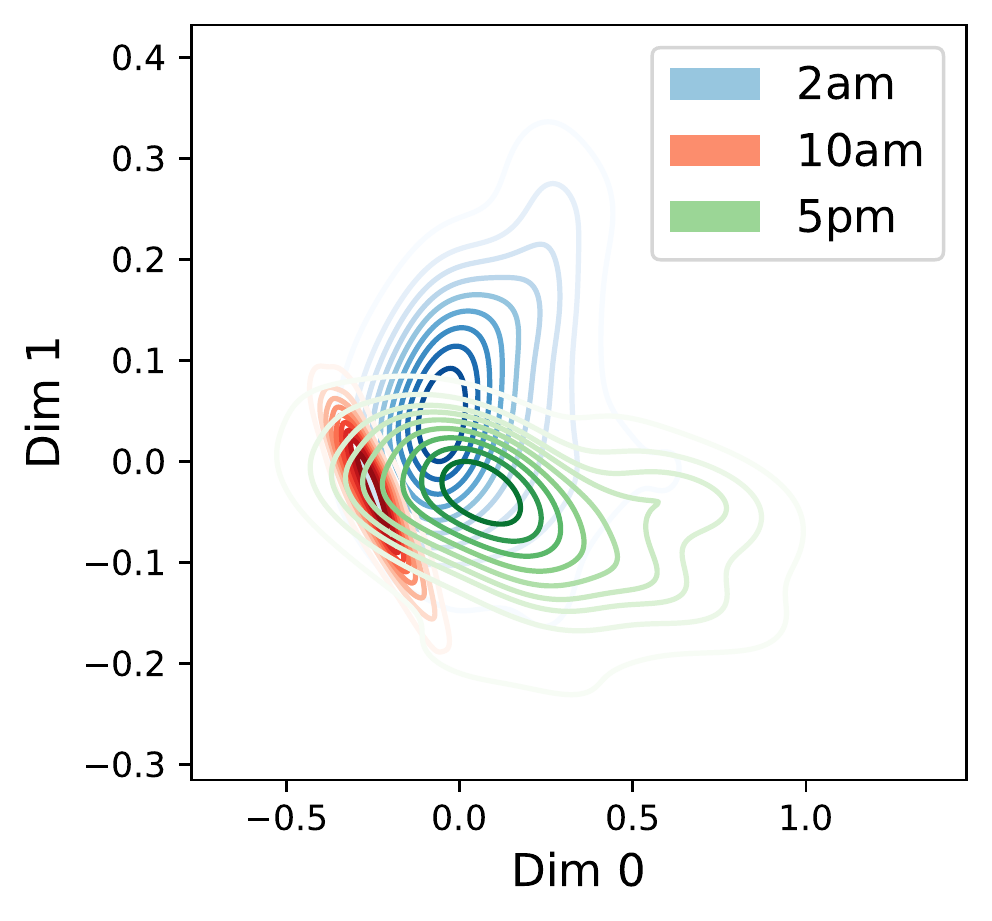}
}
\subfloat[$\sigma^\text{ce}_{n, t}$]{
\includegraphics[width=0.24\linewidth]{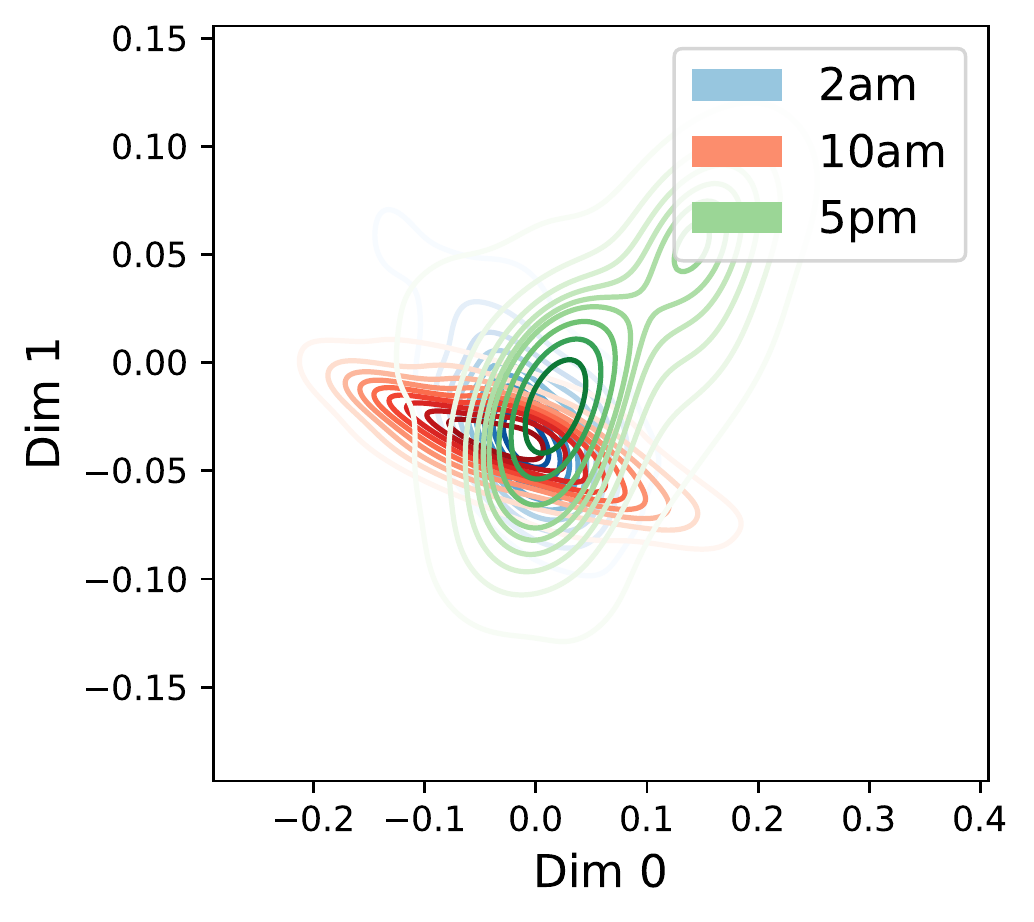}
}

\caption{Visualization of structured components.}
\label{fig:mean_qualitative}
\end{figure*}
\subsection{Robustness Analysis}

To evaluate model robustness, we subject each model to two commonly encountered data corruptions: i.i.d. Gaussian noise and missing data. The less a model's performance degrades in the presence of these corruptions, the more robust it can be considered. In our comparison, we include SCNN, SCNN w/o aux, SCINET, GW, MTGNN, and AGCRN, with 'SCNN w/o aux' denoting the SCNN model without the structural regularization module enabled.

As demonstrated in Fig. \ref{fig:robust}, SCNN consistently exhibits the smallest performance degradation among all models under each type of corruption. This is true even when compared to SCNN w/o aux, which underlines the important role of the structural regularization module in enhancing SCNN's robustness. These results underscore SCNN's superior robustness relative to the other models examined, highlighting its resilience in the face of data corruption. 

\subsection{Scalability Analysis}

In Section \ref{sec:complexity}, we demonstrate through theoretical analysis that the SCNN surpasses SOTA methods in terms of scalability. In this section, we empirically confirm SCNN's enhanced scalability. The comparison of SCNN and conventional methods is visually represented in Fig. \ref{fig:complexity}. SCNN requires significantly fewer parameters compared to NN-based SOTA models, with a parameter count comparable to that of DLinear. Additionally, SCNN, in its test mode, achieves a minimal running time of just 0.04 seconds per sample, making it seven times more efficient than DLinear. In its training mode, SCNN takes 0.3 seconds per sample, which is on par with NN-based SOTA models.

\subsection{Interpretability Analysis}

A widely accepted, non-mathematical definition of interpretability is: "Interpretability is the degree to which a human can understand the cause of a decision" \cite{miller2019explanation}. The greater the interpretability of a machine learning model, the easier it becomes for an individual to comprehend the reasons behind specific decisions or predictions. In the realm of time series forecasting, it's crucial for the model to precisely identify how backward variables influence forward variables, in a manner that aligns with human intuition. Given the demonstration of our study that time series data can be decomposed into heterogeneous components, we evaluate the interpretability of our SCNN by assessing its ability to predict each of these components. This assessment is conducted through an examination of the component extrapolation module.

Addressing long-term and seasonal components is straightforward, thanks to the model's design which replicates estimations from past time points to future horizons, as shown in Eq. \ref{eq:lt_extra} and Eq. \ref{eq:se_extra}. For the remaining three components – short-term, co-evolving, and residual – the influence of a backward variable at time $t-j$ on the prediction at time $t+i$ is captured by the parameter matrix $\hat{W}_{ji}$. This matrix links these time points, as indicated in Eq. \ref{eq:short_term_extra}. We use the Frobenius norm of this parameter matrix to quantify each contribution. The resulting contribution matrix, mapping backward variables to predicted ones, is presented in Fig. \ref{fig:interpretability}. This matrix reveals a trend where the impact of backward variables diminishes over time. This trend is consistent with the intuitive understanding that the predictability of these less regular components is based primarily on recent historical data. Furthermore, our results show that as the regularity of a component decreases, its predictability from historical variables correspondingly drops, aligning well with our expectations.

\subsection{Anomalous Cases Performance Comparison}

We provide evidence through two case studies that the SCNN consistently outperforms two competitive baselines, MTGNN and ST-Norm, particularly when dealing with anomalous patterns. This is illustrated in Fig. \ref{fig:case}. The left figure represents an episode of a time series demonstrating irregular behavior, while the right figure exhibits another episode characterized by a distinct and primarily regular daily cycle.

In examining both regular and irregular episodes, we focus on two specific periods and plot the rolling predictions—predictions made on a rolling basis using a sliding window of data—for the initial forecast horizon as generated by the three models during these periods. The results demonstrate that the SCNN consistently achieves the lowest prediction error among the three models in all four scenarios. This indicates the efficacy of our design in enabling the SCNN to effectively handle anomalies or distribution shifts in a variety of contexts. These results underscore the potential of SCNN to deliver reliable and robust forecasting in diverse and challenging scenarios.

\subsection{Disentanglement Effect Investigation}

We conduct a qualitative study to cast light on how the structure of representation space is progressively reshaped by iteratively disentangling the structured components. The structured components are visualized in Fig. \ref{fig:mean_qualitative}. For the sake of visualization, we apply principal component analysis (PCA) to obtain the two-dimensional embeddings of the residual representations. Then, to convey the characteristics of the structure for any component, we perform two coloring schemes, where the first scheme, as shown in the first row of Fig. \ref{fig:qualitative}, separates the data points according to their spatial identities, and the second one, displayed in the second row of Fig. \ref{fig:qualitative}, respects their temporal identities. For clarity, we plot the kernel density estimate (KDE) for each group of points. It is conspicuous that by progressively removing the structured components from $Z_{n, t}^{(0)}$, the residual representations with different spatial and temporal identities gradually align together, suggesting that the distinct structural information has been held by the structured components.
\section{Conclusion and Future Work}

In this study, we put forth a generative perspective for multivariate time-series (MTS) data and accordingly present the Structured Component Neural Network (SCNN). Comprising modules for component decoupling, extrapolation, and structural regularization, the SCNN refines a variety of structured components from MTS data. Our experimental results affirm the efficacy and efficiency of the SCNN. We also conduct a series of case studies, ablation studies, and hyper-parameter analyses to perform in-depth analyses on SCNN. The model's robustness is tested against common data corruptions, such as Gaussian noise and missing data, and it consistently exhibits the smallest performance degradation among all models under each type of corruption. Furthermore, SCNN is shown to be highly effective in handling diverse and challenging scenarios, including distribution shifts and anomalies, and exhibits superior robustness compared to other models.

Looking forward, our future research will explore the potential for automating the process of identifying the optimal neural architecture, using these fundamental modules and operations as building blocks. This approach promises to alleviate the laborious task of manually testing various combinations in search of the optimal architecture for each new dataset encountered. Moreover, we anticipate that this strategy could aid in uncovering the structures and meta-knowledge inherent in time-series data. For instance, time series with complex dynamics may require high-order interactions among the structured and residual components, necessitating a large-scale neural network comprising numerous modules and complex interconnections. Extending this line of inquiry, we could discern commonalities and differences between various datasets based on the neural architectures trained on them. This represents an exciting direction for future work, potentially unveiling deeper insights into time-series analysis.

\bibliographystyle{IEEEtran}
\bibliography{sample-bibliography-biblatex}

\begin{IEEEbiography}[{\includegraphics[width=1in,height=1.25in,clip,keepaspectratio]{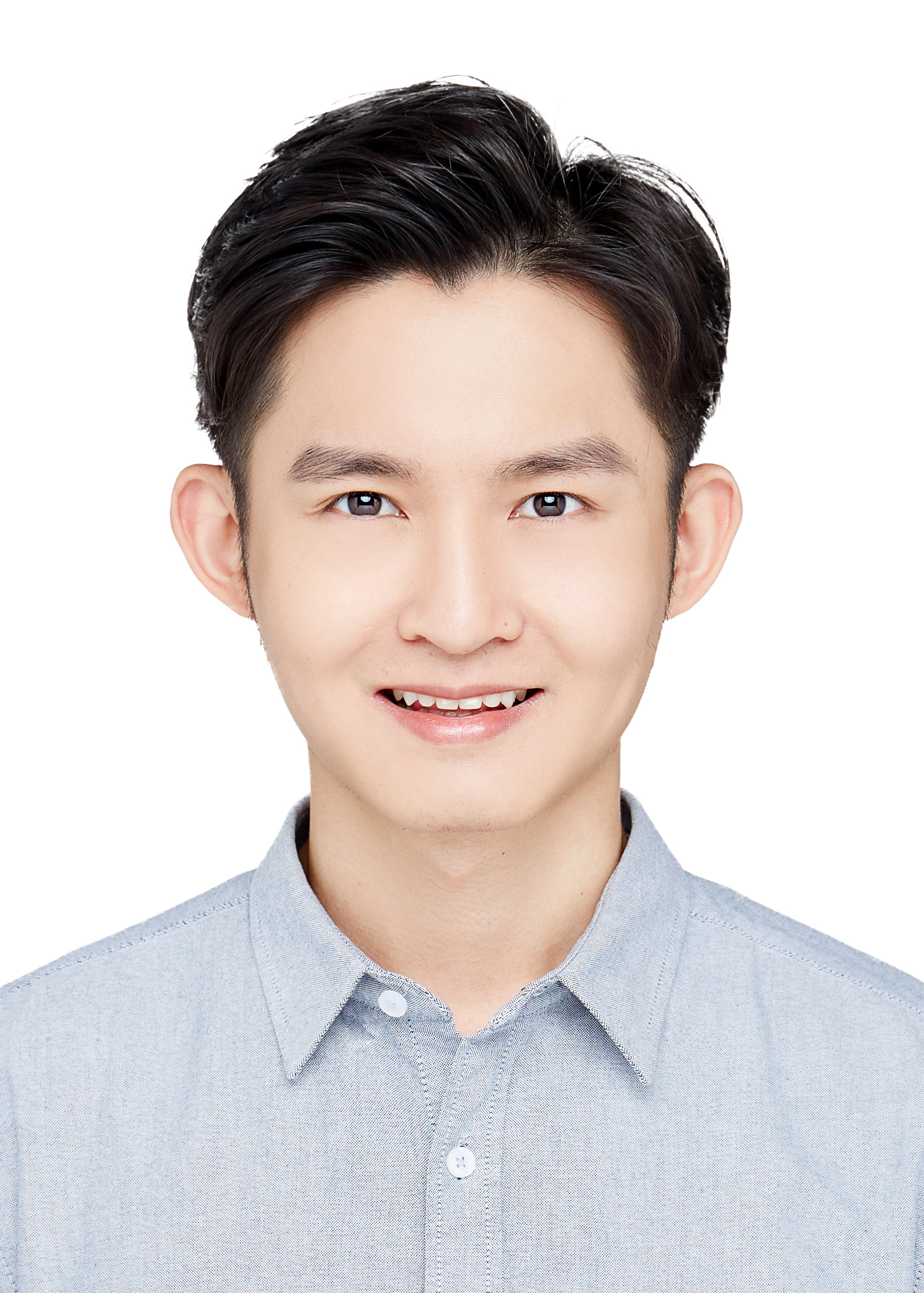}}]{Jinliang Deng} received a B.S.
degree in computer science from Peking University in 2017, and a M.S. degree in computer science from The Hong Kong University of Science and Technology in 2019. He is currently a Ph.D. candidate in the Australian Artificial Intelligence Institute, University of Technology Sydney and the Department of Computer Science and Engineering, Southern University of Science and Technology. His research interests include time series forecasting, urban computing and deep learning.
\end{IEEEbiography}

\begin{IEEEbiography}[{\includegraphics[width=1in,height=1.25in,clip,keepaspectratio]{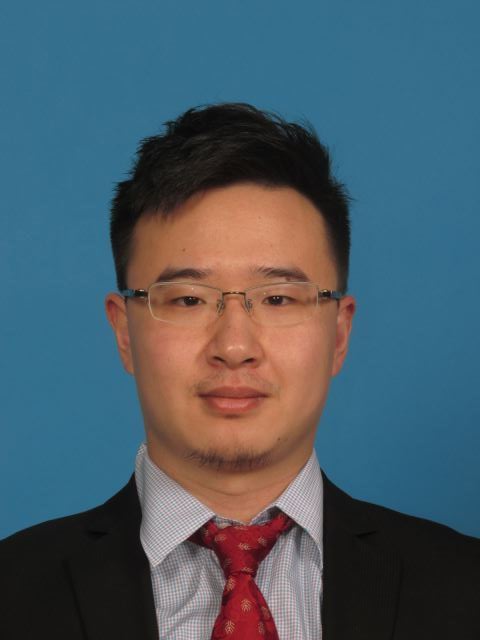}}]{Xiusi Chen} received a B.S.
degree and a M.S. degree in computer science from Peking University, in 2015 and 2018, respectively. He is currently a Ph.D. candidate in the Department of Computer Science, University of California, Los Angeles. His research interests include natural language processing, knowledge graph, neural maching reasoning and reinforcement learning.
\end{IEEEbiography}

\begin{IEEEbiography}[{\includegraphics[width=1in,height=1.25in,clip,keepaspectratio]{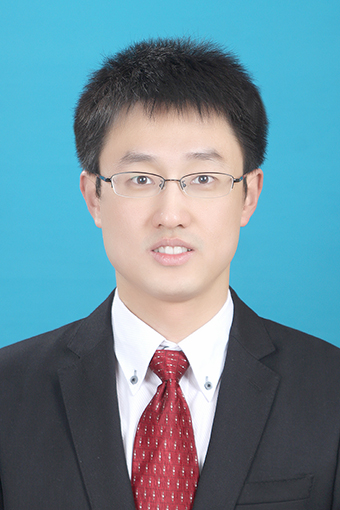}}]{Renhe Jiang} received a B.S. degree in software engineering from the Dalian University of Technology, China, in 2012, a M.S. degree in information science from Nagoya University, Japan, in 2015, and a Ph.D. degree in civil engineering from The University of Tokyo, Japan, in 2019. From 2019, he has been an Assistant Professor at the Information Technology Center, The University of Tokyo. His research interests include ubiquitous computing, deep learning, and spatio-temporal data analysis.
\end{IEEEbiography}

\begin{IEEEbiography}[{\includegraphics[width=1in,height=1.25in,clip,keepaspectratio]{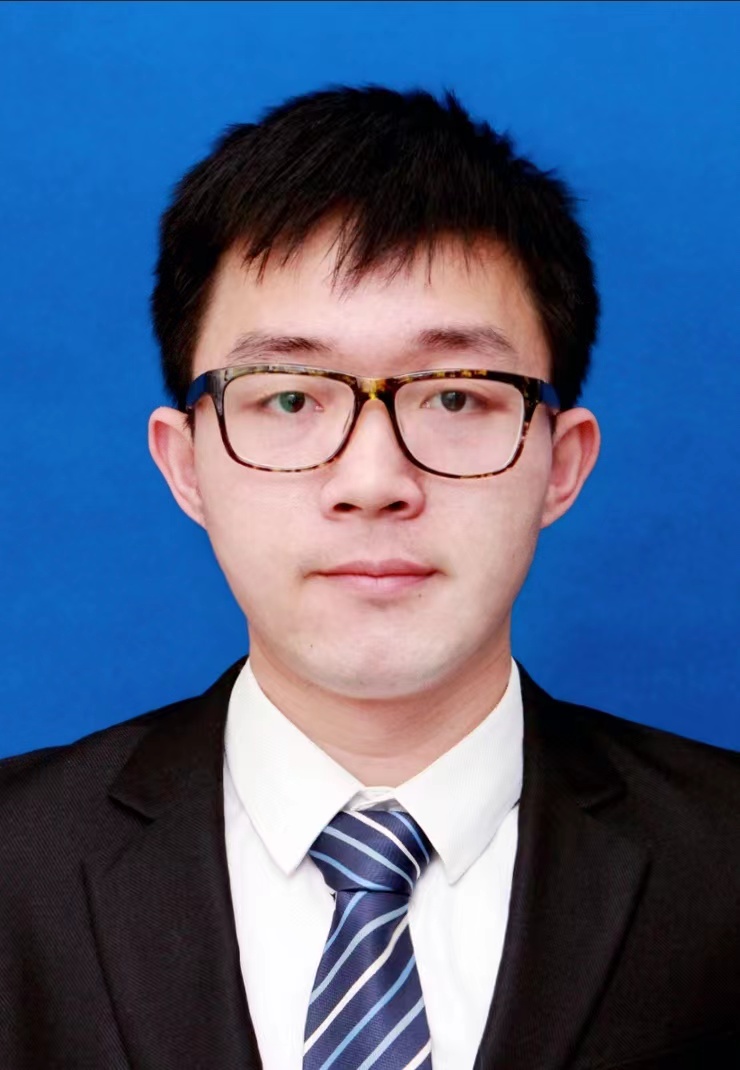}}]{Du Yin} received his B.S. degree in Electronic Information School of Wuhan University, Wuhan, China, in 2017, M.S. degree in the Department of Computer Science and Engineering from Southern University of Science and Technology, Shenzhen, China, in 2022. He is currently persuing a Ph.D. degree with the Department of Computer Science and Engineering, UNSW, Sydney. His research interests include deep learning, spatio-temporal traffic data mining, urban computing and big data.
\end{IEEEbiography}

\begin{IEEEbiography}[{\includegraphics[width=1in,height=1.25in,clip,keepaspectratio]{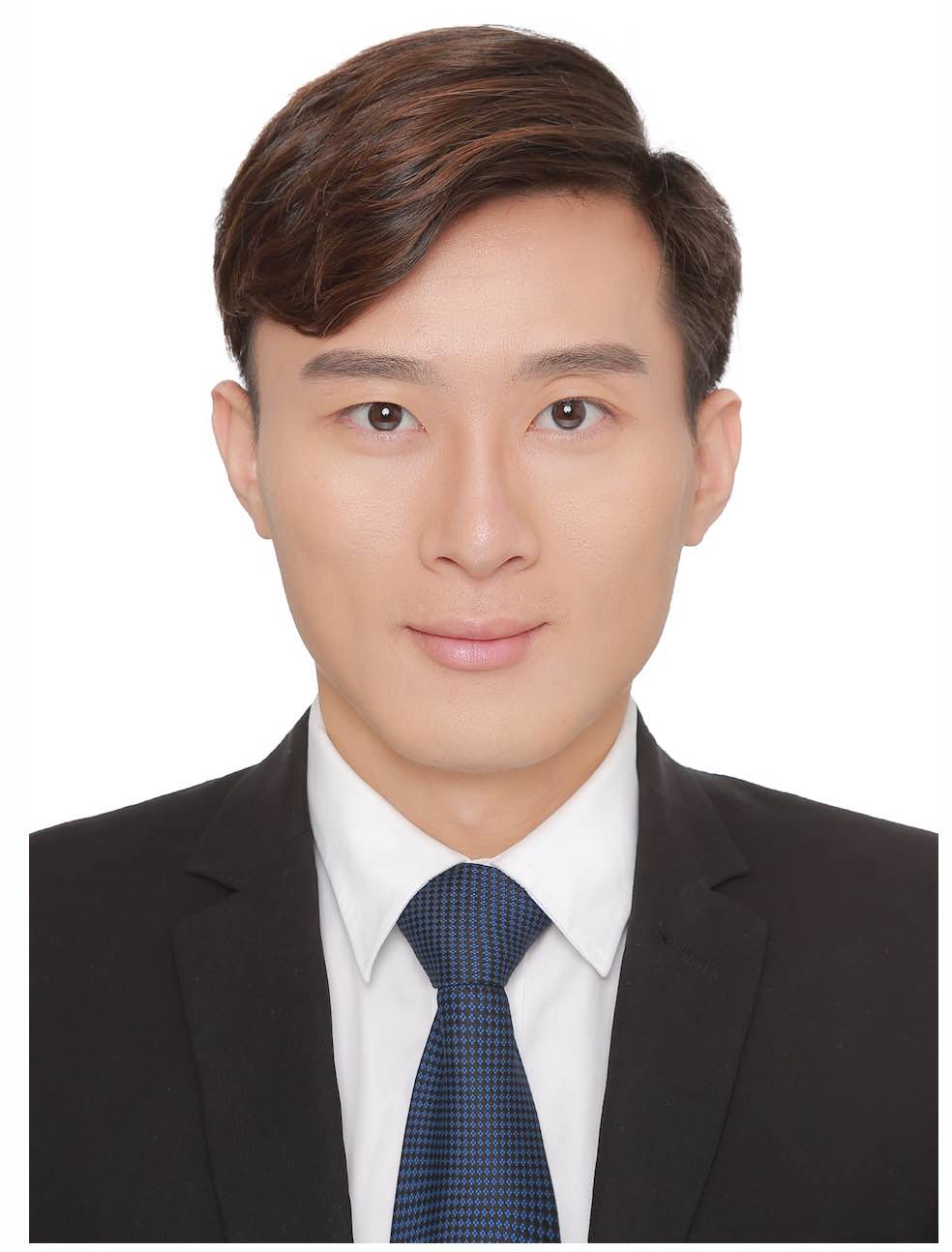}}]{Yi Yang} received a B.S. degree in computer science from Beijing University of Post and communication in 2017, and a M.S. degree in computer science from The Hong Kong University of Science and Technology in 2018. He is currently a DataScience engineer Tencent Technology. His research interests include time series forecasting,casucal inferece, computing and deep learning.
\end{IEEEbiography}

\begin{IEEEbiography}[{\includegraphics[width=1in,height=1.25in,clip,keepaspectratio]{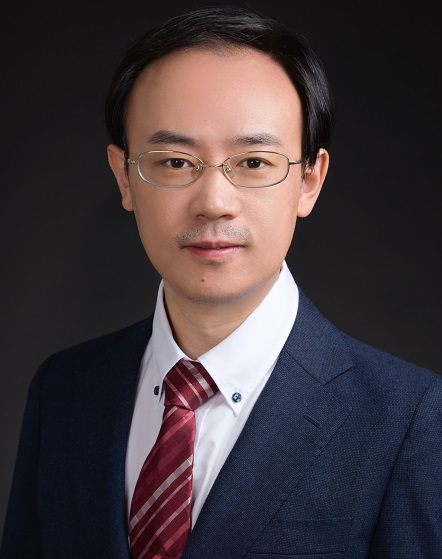}}]{Prof. Xuan Song} received a Ph.D. degree in signal and information processing from Peking University in 2010. In 2017, he was selected as Excellent Young Researcher of Japan MEXT. He has served as Associate Editor, Guest Editor, Area Chair, Senior Program Committee Member for many prestigious journals and top-tier conferences, such as IMWUT, IEEE Transactions on Multimedia, WWW Journal, ACM TIST, IEEE TKDE, Big Data Journal, UbiComp, IJCAI, AAAI, ICCV, CVPR etc. His main research interests are AI and its related research areas, such as data mining and urban computing. To date, he has published more than 100 technical publications in journals, book chapters, and international conference proceedings, including more than 60 high-impact papers in top-tier publications for computer science. His research has been featured in many Chinese, Japanese and international venues, including the United Nations, the Discovery Channel, and Fast Company Magazine. He received the Honorable Mention Award at UbiComp 2015.
\end{IEEEbiography}

\begin{IEEEbiography}[{\includegraphics[width=1in,height=1.25in,clip,keepaspectratio]{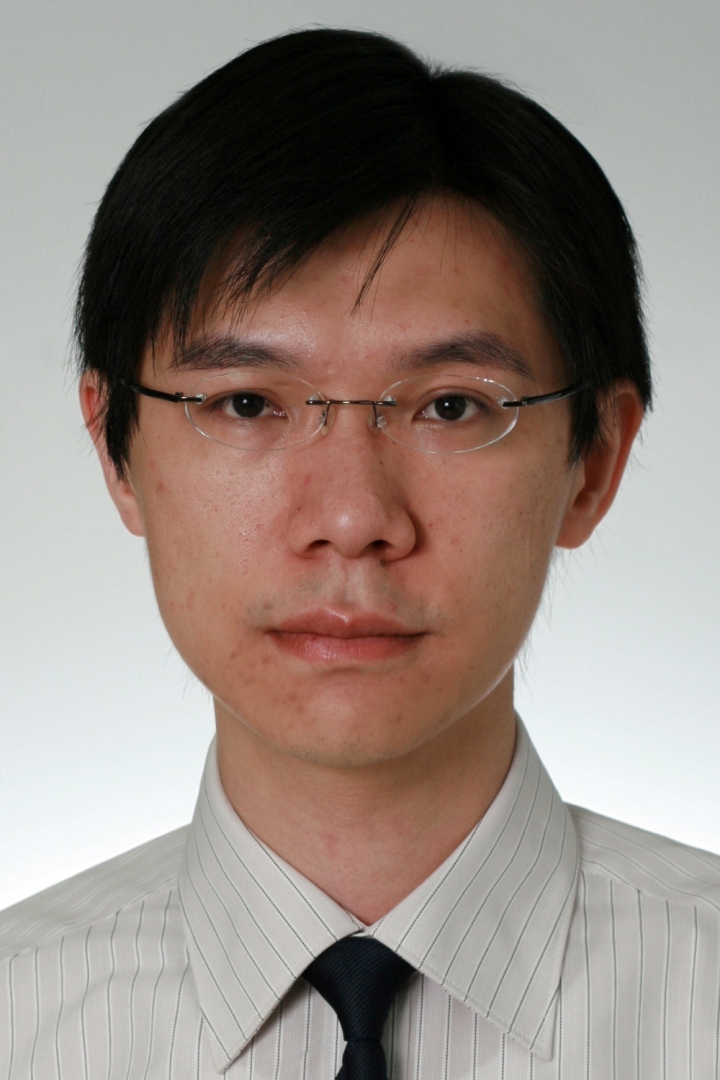}}]{Ivor W. Tsang}
is the Director of A*STAR Centre
for Frontier AI Research. He is a Professor of
artificial intelligence with the University of Technology Sydney, Ultimo, NSW, Australia, and the
Research Director of the Australian Artificial Intelligence Institute. His research interests include
transfer learning, deep generative models, learning with weakly supervision, Big Data analytics for
data with extremely high dimensions in features,
samples and labels. In 2013, he was the recipient
of the ARC Future Fellowship for his outstanding
research on Big Data analytics and large-scale machine learning. In
2019, his JMLR paper Towards ultrahigh dimensional feature selection
for Big Data was the recipient of the International Consortium of Chinese
Mathematicians Best Paper Award. In 2020, he was recognized as the AI
2000 AAAI/IJCAI Most Influential Scholar in Australia for his outstanding
contributions to the field between 2009 and 2019. His research on transfer
learning granted him the Best Student Paper Award at CVPR 2010 and
the 2014 IEEE TMM Prize Paper Award. Recently, he was conferred
the IEEE Fellow for his outstanding contributions to large-scale machine
learning and transfer learning. He serves as the Editorial Board for the
JMLR, MLJ, JAIR, IEEE TPAMI, IEEE TAI, IEEE TBD, and IEEE TETCI.
He serves/served as a AC or Senior AC for NeurIPS, ICML, AAAI and
IJCAI, and the steering committee of ACML.

\end{IEEEbiography}

\end{document}